# ROSS User's Guide and Reference Manual

## (Version 1.0)

Author: Glenn R. Hofford

Date of Publication: November 8, 2014

# Table of Contents

# 1. Introduction

The ROSS method (Hofford 2014) is a new approach in the area of representation that is useful for many artificial intelligence and natural language understanding (NLU) tasks. (ROSS stands for "Representation", "Ontology", "Structure'", "Star" language). ROSS is a physical symbol-based representational scheme. ROSS provides a complex model for the declarative representation of physical structure and for the representation of processes and causality. From the metaphysical perspective, the ROSS view of external reality involves a 4D model, wherein discrete single-time-point unit-sized locations with states are the basis for all objects, processes and aspects that can be modeled.

The ROSS method is also capable of application to the representation of abstract things. The ROSS approach models abstract things by grounding them in a 4D space-time model. Abstract entities that are modeled include the entities that are involved in the representation of representation ("meta-representation"), including representation of intelligent agent mental representations, cognition and communication.

This document describes the general aspects of ROSS but it ties ROSS to the NLU area for examples and in order to illustrate some of the general concepts. ROSS is used in two ways within NLU systems: 1) the Star language is used for the specification of object classes and rule-like constructs referred to as behavior classes in an ontology/knowledge base, and 2) a formal scheme of ROSS called the "instance model" is used for the specification of meaning representations that represent the semantics of a particular situation.

A ROSS repository that houses an ontology and a knowledge base is referred to as an "Infopedia". A ROSS Infopedia stores supporting definitions, object frame classes, and behavior classes that are representations of conceptual, or world knowledge about processes and causality. Infopedias are interchangeable – there is no one Infopedia. An Infopedia is organized into several tiers: an upper tier contains supporting definitions and high-level abstract classes, a middle tier containing classes whose primary purpose is functional (middle tier classes are used in many behavior classes), and a lower tier of object classes containing a large number of classes that are distinguishable from other similar classes by a few features. Examples of lower tier classes include "house cat", "trophy",  and "father-person".

An internal ROSS instance model is used during processing within an NLU system that uses ROSS: the instance model represents factual information about particular situations (past, present or hypothetical situations). An instance model is a type of ROSS *fact repository*.

This document describes the Star language definition features, ROSS fact repositories, and the ROSS instance model. In addition, the use of ROSS for inference is introduced and described at an overview level. The ROSS method has been applied within the NLU area to support the task of anaphora resolution[1]. The appendix lists a set of classes that are used in order to perform anaphora resolution and commonsense reasoning for the original Winograd Schema Challenge schema (schema #1: "councilmen and demonstrators").

---

[1] This NLU system is called "Comprehendor".

## 2. Lexicon of ROSS Terms

ROSS and the Star language employ an extensive set of terms, some of which are unique to ROSS and Star, and some of which are terms that are used in the AI and NLU fields but which have specific meanings in the context of ROSS and Star. A list of important ROSS terms and their definitions as used in this paper is as follows. (Terms that are specific to the NLU area are marked as such).

- **analog** (alt. "**analogue"**): a representation, such as a map, that contains representative elements that have a one-to-one correspondence with that which is represented. (Not to be confused with "analog" as in "analog" versus "digital"). The ROSS method enables the creation of definitions that are actually *pseudo-analogical*.
- **attachment:** the process of positioning an object frame instance in relation to a parent object frame instance so it that exists within the extent dimensions of the structural parent. I.e. attachment establishes an object instance as a *part* of some *whole*. The object instance may be only structurally attached, i.e. added to the list of components of the parent object instance, in which case it exists within the extent dimensions of the structural parent. If more detail is known, the *part* object instance may be positioned via the use of a set of locational and size attributes that are contained within its *RelationshipToParent* section.
- **attribute**: (also: *attribute expression*) a two-part expression that specifies a single feature for an object frame instance. The attribute consists of a predefined attribute type name and a value expression.
- **attribute type**: a definitional expression that declares an attribute type name with an associated attribute value set.
- **attribute value set**: a definitional expression that defines a set of numeric or string values that can be used in other expressions to specify the location or the quality of something. An attribute value set that is composed of a set of string values is an enumerated value set type.
- **behavior:** an instance of a behavior class (sometimes used synonymously with "behavior class").
- **behavior class**: in its basic form, a definitional construct that includes multiple references to constituent populated objects, each of which is a definition that associates an object frame class with a qualitative state; within a behavior class all populated objects are related to one another using a mechanism called the "binder". An extended form of the behavior class allows for elements that are nested behavior references.
- **binder:** a definitional construct that establishes a base spatial and temporal location within a behavior class: this allows the spatial and temporal locations of all other object frame classes of all populated objects of the behavior class to be specified in relation to the object frame class for which the binder is defined.
- **class**: the Star language contains several class types: the *object frame class,* the *behavior class,* the *populated object class* and the *template class.* A Star class is not the equivalent of a *set:* ROSS implements sets using object instance *collections*.

- **collection:** the concept of "collection" in ROSS is abstract; the term indicates multiplicity of a set of things that share some or multiple properties. The collection concept is used in several places and is indicated by a "Multiple" flag: in object frame classes, in populated object classes, and for object instances.
- **communication unit**: an item within a natural language text fragment or document. The most common type of communication unit is the sentence. Other communication unit types include email addresses, web addresses (URLs), and news headlines. *(NLU-specific)*.
- **dimension set expression**: the specification expression that specifies the spatial and/or temporal location of an object frame instance.
- **dimension system**: (also *dimension system type*) a Star language definition that consists of one or more related attribute types, which must be used together in order to specify the spatial and/or temporal location of an object frame instance.
- **existential instantiation:** *(loosely used to convey the following concept)* A ROSS fact-containing repository (e.g. an NLU instance model) must declare a structural parent object frame instance; this must contain or house a dimension system (e.g. a Cartesian coordinate system). This can be visualized as a rectangular shaped region (a cube or rectangular right prism (a cuboid)). It is a collection, or aggregation of unit-sized location entity instances. Because this frame of reference exists, there is not a need for the use of *existential quantifiers* (as with first-order logic) for propositional ROSS expressions. Once this frame of reference has been created, the main subsequent representational task is that of infusing or populating the individual cells (like the cells in a matrix) with values. (cf. *pseudo-analogical representation*).
- **fact**: (also: *simple assertion*) The term *fact* refers to a family of *fact-like constructs*. A fact is a representational construct that represents "where" (location) and "what" (qualitative value) for an object instance; however, the location may be specified using an attribute value range (a disjunction of specific locations), or it may be unspecified, in cases where the object instance has been attached to a structural parent instance as a structural component (in this case it exists within the extent dimensions of the structural parent). Facts can also be negated. Fact-like constructs may pertain to real past situations, or they may describe hypothetical facts in a hypothetical world. A distinction is made between simple assertions that represent completed states (or events) (whether real or hypothetical) and those that represent predictions[2] (predicted states) or goals (e.g. within the context of AI planning). (cf. *two-part attribute cluster*).
- **fact repository:** a general term that is used to describe any representational artifact that stores facts. Examples include prediction/goal *specification transcripts* from the area of AI planning and design, and NLU *instance models*. (cf. *instance model*, *transcript*).
- **fuzzy class:** an object frame class that contains features that are associated with the class using a probability field. E.g this allows for the specification of an animal such as a house cat that has a front left leg with a probability of .99 (since some cats may be missing a front left leg).

---

[2] "Prediction" here refers to a predicted state, a sort of "future fact". This definition differs from the use of "prediction" in the context of machine learning.

- **inference:** (also referred to as *automated reasoning*) 1) a computational process that uses ROSS behavior class-based rules and one or more known facts from an instance model in order to derive new facts (for rules and existing facts that include a probability field, this involves the generation of new facts that are assigned a probability value), 2) a process that involves the instantiation of an object instance based on an object frame class from which it derives a set of features (this corresponds to the term *syllogistic deduction*). (Some features of such (fuzzy) class may be probabilistic (similar to *fuzzy sets*)), 3) other forms of inference such as geometrical/spatial reasoning *(not addressed in this version)*.
- **Infopedia:** a ROSS repository comprising an ontology and knowledge base. "Infopedia" may refer to a collection of text files consisting of Star language definitions, or it may refer to the internal in-memory repository, e.g. within an NLU system.
- **infusion:** the processing task (e.g. by an NLU semantic engine) that involves setting one or more qualitative attribute values for an object frame instance. Infusion uses the ROSS *template class*. (cf. *population*).
- **instance model**: in the context of NLU, a fact-containing representational artifact that is a meaning representation that represents the subject matter (semantic content) of natural language text. An instance model may exist in memory or in serialized form in a text file. The serialized form may use XML or it may use the Star language. *(NLU-specific)*
- **knowledge base**: a repository that contains supporting Star language definitions (including object frame class definitions) and behavior class definitions. (Cf. *ontology*). The term "knowledge base" usually refers to the behavior classes of the repository, especially those behavior classes that are used as rules. A ROSS knowledge base is strictly definitional; it does not store factual knowledge, e.g. historical facts about the past.
- **locational attribute**: an attribute that specifies a location for an object frame class or instance, either fully or in part, when used with a set of related locational attributes.
- **meaning representation instance**: a representational artifact that represents the subject matter of natural language text. An example that uses logic would consist of a collection of logical expressions. An example from the semantic web area that uses RDF/OWL would consist of RDF. A ROSS meaning representation instance is implemented using a ROSS instance model. *(NLU-specific)*
- **meaning unit**: in the context of NLU, a tree-like representational construct that contains a subject and a predicate and a list of adverbial phrases. *(NLU-specific)*
- **object frame class**: an object frame class is a construct that represents a time-independent cuboid region in 3D space. An object frame class is not the equivalent of a movable object. Multiple time-sequential object frame classes are needed in order to represent a movable object such as a bouncing ball. An example object frame class would be a particular "PersonObjectFrameClass".
- **object frame instance**: (or, *object instance*) an (instantiated) instance of an object frame class that exists at a single time point along a timeline.

- **ontology**: a ROSS ontology is a repository of information that consists of supporting definitions and object frame classes. This information exists in any of several forms: 1) the collection of text files that contains Star language code, and 2) the in-memory representation of the Star definitions that gets created by a system that compiles the Star language text files and creates an in-memory repository consisting of the same information. A ROSS ontology is closely related to a ROSS knowledge base. There are two main uses for a ROSS ontology: 1) to support queries about its classes, and 2) to support the creation of fact repositories or transcripts (e.g. instance/situation models). ROSS ontologies are bottom-up, not top-down: a given ROSS ontology need not have a root object frame class (although this is possible).
- **physical symbol system**: a representational system that is based on a representational scheme wherein all information is represented using symbols in such a way that it is both fully human-readable and capable of automation with regard to processes that generate and use the information.
- **populated object class**: an abstract class that supports the task of setting qualitative attributes for an object instance. The populated object class is mainly used within behavior classes.
- **population:** the processing task (e.g. by an NLU semantic engine) that assigns one or multiple qualitative attribute values to an object frame instance. (somewhat similar to *infusion*). Population uses the ROSS *populated object class*.
- **pseudo-analogical representation**: a representational approach that provides a representational construct or mechanism that implicitly represents a set of locations (i.e. a spatial area and temporal set of intervals or timeline time points). In contrast with an analogical representation, a pseudo-analogical representation need not explicitly represent every unit location with a symbol or symbol construct; rather, this is accomplished via representational mechanisms that specify a region; they are accompanied by default assumptions about the values of each location (e.g. that all locations are *space* by default). ROSS implements a pseudo-analogical representation primarily by use of the dimension system, the structural parent object frame class and the structural parent instance.
- **qualitative attribute** (also *value attribute*): an attribute that specifies a static value for a location. Examples include "material composition", with values such as "plastic", "metal", "wood", and color, with values such as "red", "green", "blue".
- **RelationshipToParent section**: an information field for an object frame class (and for object instances that are instantiated from the class) that represents the location, spatial orientation and size (extent) of the object frame (class or instance) in relation to the parent (or "whole") of which it is a part.
- **rule:** a declarative representational construct that represents causality or correlation in the represented world. A ROSS rule is implemented using the behavior class; a rule is structurally more complex than a fact; it contains references to object classes and states within each of an antecedent section and a consequent section. Rules usually exist to support inference-related processing tasks. A ROSS rule may contain items that include a probability field, to allow for probability-based inference.

- **shape template**, **expandable shape template**: a template class that refers to a 3D drawing routine or a 3d bitmap which can be expanded when it is used to infuse values for a specific object instance.
- **situation model:** an NLU instance model; this term usually conveys the idea that the instance model represents a situation that occurred in the physical world at some time in the past.
- **specification system**: (also *dimension system type*) a definition of a type that includes a set of locational attribute types and an *inner content* section. The inner content section is either a set of qualitative attribute types or a flag that indicates that a Structure section is used in order to represent the qualitative features.
- **structural parent class**: a structural parent class is an object frame class that is used for placement of other smaller embedded objects frame classes. The embedded object frame classes are usually not actual structural parts. The structural parent serves as a frame of reference.
- **structural parent instance**: a structural parent object frame instance is an instance of a structural parent class. It is a "large" object frame that holds other object frame instances. (The analogy of a *diorama* can be useful for understanding the structural parent instance).
- **template**, **template object class**: a class that is used in order to set one or multiple values of qualitative attributes for an object frame class or instance (cf. *infusion*).
- **transcript:** (also *specification transcript)* an artifact that contains a collection of related facts that exist for some storage or computational purpose. An NLU instance model is one type of transcript. In the area of AI automated reasoning, transcript types would include: a goal statement transcript for computer software specifications (similar to a computer program), and a transcript that specifies facts for a diagnostic expert system. (cf. *fact repository*).
- **two-part attribute cluster**: any representational construct that represents both of the following: all locational attributes that are defined by a dimension system  (e.g. x, y, z coordinates), and at least one qualitative attribute. (cf. *fact*).
- **value, value expression**: an expression that is used within an attribute to specify a value.
- **value attribute**: (cf. qualitative attribute)

## 3.  The Rationale Behind ROSS

The ROSS method was created as a representational scheme and tool that can be used as a platform and foundation for representation that is complete, expressive, and useful. ROSS is an implementation and embodiment of a unique set of ontological commitments involving a particular viewpoint/world view that is based on a set of naïve representation/modeling premises ("naïve" is used somewhat in the tradition of *naïve physics*). These naïve premises involve a segmentation of all problem domains into discrete space-time units referred to as "unit-sized location objects". A second set of premises is that movable objects do not "exist" and that motion does not "exist". (The premises are not assertions about the physical world but are operational assumptions for the purpose of creating consistent representations or models).

ROSS addresses the need for symbol grounding – the ROSS approach does not ground symbols in sensor-based data, but it uses a sophisticated and elaborate scheme that is both human and machine generatable and readable. It is a physical symbol-based scheme that provides features that allow for the capturing of a sufficient level of 4D detail to enable inference and query. ROSS has been successfully used as a representational platform and infrastructure that supports many natural language understanding tasks; these include anaphora resolution using a method that uses the ROSS instance model, and a further extension of the method that uses commonsense reasoning[3]. (The overall method involves a complex set of inference processes that perform word-sense disambiguation and resolution of difficult pronouns).

*(For more information about the background and rationale for ROSS, refer to "Introduction to ROSS: A New Representational Scheme" (see Reference for details).*

## 4. Conceptual Architecture

ROSS includes the Star language, which is a language for ontology and knowledge base creation. ROSS also includes a formal syntax (schema) for XML-based instance model specifications. An instance model is a fact-oriented transcript of a situation. The two main functional components of ROSS are implemented as follows:

- **Infopedia: Ontology and Knowledge Base**: Externally, this consists of a set of text files that contain Star language code. The Star language code consists of a set of definitions for objects and behaviors such as "everyday object", "common object", "container object", "enclosable object", "person", "car", "food item", the "walking" behavior, the "hitting person" behavior, "intelligent agent", "cognitive explanation abstract entity", the "communicating" behavior, "communicated information", etc. (There are also supporting definitions such as *attribute value set definitions* and *attribute type definitions*). When an Infopedia is used by an NLU system, an in-memory knowledge base is constructed using the external Star code definitions: these are read in and processed by a Star language compiler.

- **Instance Model**: An instance model models physical structure, processes and causality for a particular situation. For instance, an instance model may represent the objects and processes for a story such as a news story. An example from the Winograd schema challenge for the "trophy and suitcase" schema involves a situation where an intelligent agent communicates something: i.e.. that a particular trophy does not fit in a particular suitcase. When created and used by an NLU system, an internal instance model is a set of data structures that is created and maintained by the semantic engine at run time. An internal instance model may be used by an NLU semantic engine for a variety of purposes: these include the generation of summaries, topic modeling and relationship extraction. An external instance model is an artifact that uses XML to represent the same information that is contained in an internal instance model.

---

[3] The anaphora resolution method has been successfully used to create a fully general solution that resolves pronouns for several Winograd Schema Challenge schemas, including the original schema involving the councilmen/demonstrators.

## 5.  Why and How Is the ROSS Method Analogical?

*(This section is presented as a self-contained overview that outlines the use of ROSS within an NLU story comprehension system).*

Star allows for the definition of classes that have more than just the traditional "PartOf" (whole-to-part) relationships: for instance the classes also contain special *relationship-to-parent* attributes that specify the location of a part in relation to the whole (the parent), as well as the size of the part. Any class that can function as a parent must contain a construct called a *dimension system*, which is a definitional type consisting of multiple attribute types that collectively represent a fixed location in space and time. The attribute types of a dimension system are used in order to position, or locate, the parts of a physical structural component within its parent.

ROSS is a hybrid method that integrates an analogical approach with a physical symbol system approach. The use of an analogue representation scheme within a symbol-based infrastructure allows for representations that are more natural - they are similar to human cognitive representations.

Here is an overview of how the ROSS instance model is analogical:

**Background:**

This describes an NLU system that does story comprehension. This system consists of several sub-systems that include a parser and a semantic engine.

The system has two main inputs:

  1) Sample NL text story fragment:

"The two boys crept up to the house. They broke a window at the rear of the house and climbed through the opening. Suddenly they were startled by the flash of a bright light."

2) The internal ontology/knowledge base that contains compiled Star language definitions (these definitions may include classes such as HouseClass, GroundClass, WindowClass, PersonClass and ChildClass, GroundClass, PersonCrawlsBehaviorClass, PersonClimbsBehaviorClass, WindowGetsBrokenBehaviorClass).

The system's outputs are as follows:

1) (intermediate) a list of parser-generated syntax trees (one for each input sentence)

2) an *instance model* – a semantic/conceptual representation of the story. The instance model exists entirely apart from the syntax trees. It represents the objects, entities, events, etc. of the input story using a timeline approach, as follows.

**Explanation:**

The structured instance model *is* the hybrid analogue/symbol system representation. It contains:

(1) A master 4D frame of reference that covers the entire story situation. This consists of symbol-based representations for: a) an area of 3D space (e.g. the house and its immediate neighborhood) and b) a timeline, which makes it 4D. This is called the "structural parent" instance. (the structural parent instance is analogous to a diorama (with the added dimension of time)).

(2) The structural parent instance has a dimension system (i.e. a spatial coordinate system + time) – for this example a basic Cartesian coordinate system plus a simple timeline (with enumerated time points) attribute type is used.

(3) Spatial "compartments" for the objects of the story get placed into the frame of reference system (the structural parent instance). This is analogous to putting several empty containers (or rectangular wire frames) into a diorama. This is performed by the NLU system engine as it creates the instance model, and is called "attachment". Attachment occurs at some point along the timeline. Each individual container that gets attached is situated somewhere specific within the diorama.

(4) Continuing the analogy ... now we are at time t = t1, with a diorama that contains several empty containers. They need to be filled in with something. The next step is called "infusion" (or "population") – it is like "paint by numbers", or analogous to putting tiny colored tiles each into their place in a mosaic. Infusion uses templates. A template can be a 3D bitmap or it can be a set of drawing instructions. Once each container has been populated using the templates, the representation is complete – at a single time point. At this point the instance model represents two boys (frozen as it were in the initial stage of "creeping"), and a house, an implied yard, etc.

(5) Motion (e.g. action) is modeled by extending the concepts along the timeline. (This is a complex process that makes use of ROSS behavior classes that allow for the modeling of events in terms of object states).

**Summary**

The instance model is like a 3D movie, but one where every individual 3D "pixel" is described propositionally – either directly, or indirectly. An advantage of this approach is that any individual unit-sized location in the entire 4D frame of reference can be queried for its value, and used as a basis for automated inference.

## 6. Level of Structural Detail: Several Modes

Although it has an extensive set of features for the modeling of the structural aspects of objects, behaviors and situations, the ROSS method can actually be used in any of several modes, depending on the level of structural detail that needs to be represented. Some applications of ROSS do not require the same level of structural detail as others: for instance the anaphora resolution tasks and the inferences that are performed in support of the NLU system developed by the author that solves several of the Winograd Schema Challenge schemas are dependent only on shallow representations of the objects involved: common objects such as suitcases, people, and cognition and communication entities. Many inference applications depend only on a coordinated representation of attributes and behaviors, and many NLU problems can be adequately handled using shallow structure.

### 6.1. Shallow Structure Mode

The *shallow structure mode* relies on an ontology/knowledge base model of the *attributes* and/or *behaviors* of the relevant objects and processes of a situation. This mode involves the following:

- Structural parent dimension system: this system only needs a simple integer-based "one-dimensional" spatial approach, in order that objects (e.g. a static snapshot of a person or a collection of persons) may be distinguished from other objects; the specifics of their respective locations are not specified.
- Size (extent) attributes are not specified. Within a behavior class, the constituent objects are treated as atomic units: the inner structure (whole-to-part structural features) of an object need not be specified.
- Template-based infusion is absent; e.g. a person instance is represented with a symbol (an instance identifier), however the objects of the generated instance model are transparent or "ghost-like". (This will be referred to as *transparent mode*).
- The features of object instances are described using the following qualitative attributes:

- o Basic qualitative attributes: for instance, a behavior class may define a rule for light-reflecting objects (such as a mirror) using an attribute that describes the material composition of the mirror surface, i.e. that it is metallic.
  - o Qualitative *state* attribute types can also be devised that represent useful abstractions; e.g. the ROSS ontology that is used for the Winograd "trophy and suitcase" schema uses a state attribute called "FunctionalRelativeSizeAttributeType", having values of "NotTooBig" and "TooBig". Such abstractions must be evaluated with respect to their general utility within the ontology/knowledge base.
- Rule-like behavior classes may also use nested behaviors: for instance the rule-like behavior class "PersonHitsPersonCausingHarm" contains a nested behavior class such as "PersonSuffersInjury".

Inference that uses the ROSS shallow structure mode in many ways resembles logical deduction, insofar as it is dependent on an *abstract model* of behavior and of situations. However, even in shallow structure mode the ROSS behavior class provides the following benefits: 1) object *types* are specified: populated object classes within a behavior class contain references to object frame classes (the object frame class name is the "type"), 2) active and passive roles can be specified for objects, and 3) time is handled in a natural way, since by default, all states that are represented in the antecedent section are time-prior to those of the consequent section.

## 6.2. Intermediate Levels

There are intermediate modes where the following types of detail are incorporated into the model. Where all elements are present this can be referred to as deep structure mode.

- Dimension systems use specific units of measure (e.g. the millimeter, the second) rather than just using integer-based value sets.
- Structural parent classes use dimension systems that are capable of 2D or 3D specification (e.g. a dimension system for Cartesian coordinates, e.g. a dimension system for latitude/longitude).
- Transforms between dimension systems exist to support conversions from behavior class structural parent dimension systems to the dimension systems of the object frame classes.
- RelationshipToParent attributes:
  - o AtLocations: specifications of the position of object frame classes or instances with respect to the parent object.
  - o Specifications of spatial orientation of object frame classes and instances.
  - o Size (extent) of object frame classes and instances.
- Part-to-Whole Structure for object frame classes and instances is represented using the Structure section of the object frame class.

- Template classes are used during the instantiation of object frame instances in order to infuse them with qualitative attribute values for *each unit-sized location* within the cuboid region that is covered by the object instance. (*Example:* a fixed location, using millimeter coordinates, of an object instance's unit-sized location at (x=5,y=7,z=7) has a qualitative attribute of (MaterialComposition = "Steel"), whereas (x=5,y=7,z=8) has ((MaterialComposition = "Space").

- Behavior classes include specifications of the *distance aspects* of spatial and temporal relationships among constituent objects using the *binder* mechanism.

### 6.3. Deep Structure Mode

The *deep structural mode* is only partly described in this document. Deep structural approaches are needed for situations that involve relevant features of the inner spatial structure of an object, the spatial orientation aspect of spatial relationships, and/or the inner part-to-whole relationships within an object. Deep structure mode uses the features described above. *(Inference involving deep structure mode is not described in this document).*

## 7. Overview of What Is Represented

The Star language contains two main statements – the object frame class statement and the behavior class statement. Before describing the abstract Star language definitional constructs, an illustration of what some of these constructs represents may be helpful.

### 7.1. Object Frame Class

**Figure 1** illustrates a number of things that could exist in a typical represented world that contains a single person. The diagram also includes several representational abstractions. *(Note that this illustration of concepts presents classes and instances somewhat interchangeably).* These are:

- A structural parent class: the structural parent class "houses" a dimension system which happens to be a 4D coordinate system.
- The 4D dimension system: a representational abstraction, shown within the structural parent object frame class.
- An embedded "person" object frame class.
- Two object frame classes that are components of person: "person head" and "person body".
- The word "RelationshipToParent", as a header for a set of attributes: e.g. for the person's body these attributes are the specification of the location of the body in relation to the overall person, the orientation of the object frame that holds the body in relation to the overall person, and the extent, or size of the body.
- An arrow that represents a timeline. An object frame instance only exists for one instant (with duration = 1 according to some temporal grain size – e.g. 1 second); the presence of the timeline reinforces the concept that there are actually *n* instances of the person object frame instance that exist through time.

Abstractions that are not shown include:

- The multiple dimension systems of the Person object frame class; these typically include:
    - A special "component holder" dimension system.
    - A 4D dimension system (a Cartesian coordinate + time system).
- Transforms between the above two dimension systems.

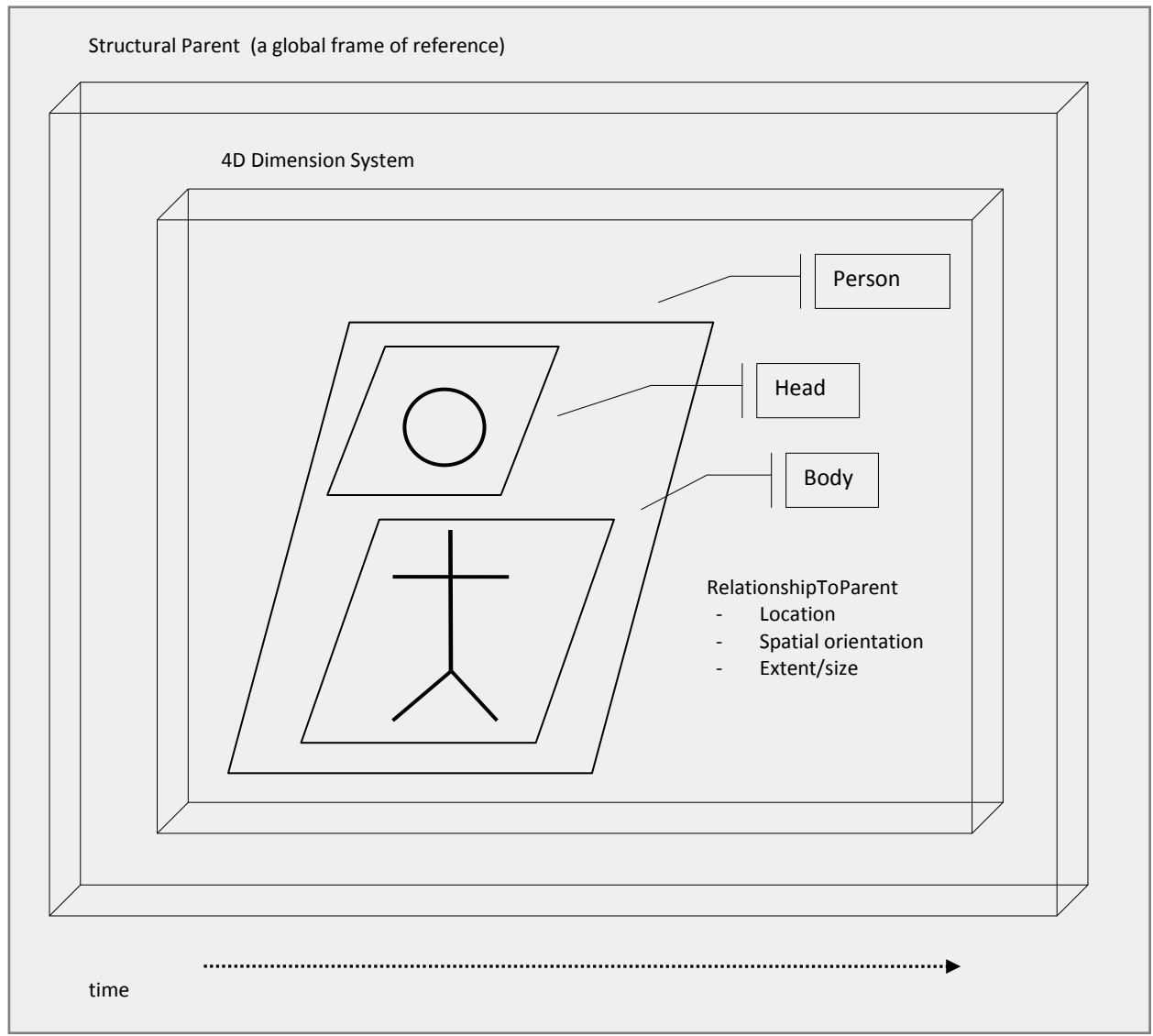

*Figure 1: Structural Parent and Embedded Object Frame Class*

Some of the descriptions of definitions in the following sections will refer to this diagram.

## 7.2. Behavior Class

The behavior class is typically used to represent a situation or process that has multiple time-sequential states. **Figure 2** represents a process wherein a person hits another person.

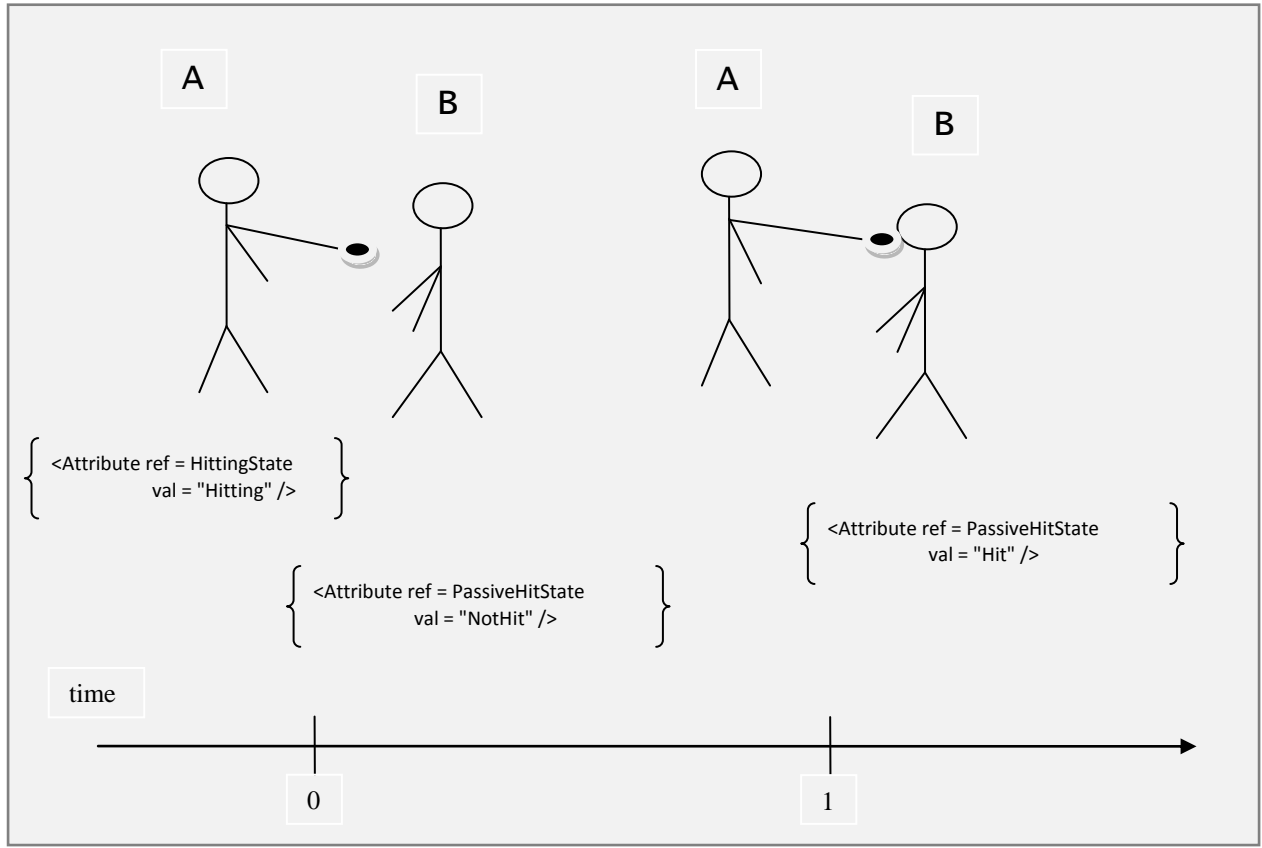

*Figure 1: Visualization for a PersonHitsPerson Behavior Class*

The diagram represents only one approach out of many possible approaches that could be used to represent the process where a "person-A" hits "person-B". The text below the person stick figures shows the attribute expressions that are contained within the populated objects within the behavior class. There are two relevant points on the timeline:

- **(t = 0)** belongs to the behavior class **antecedent** section: at t=0 the following states are true:
  - Person-A is in the "hitting" state
  - Person-B is in the passive "not hit" state
- **(t = 1)** belongs to the behavior class **consequent** section, at t=1 Person-B is in the passive "hit" state, indicating that Person-B has been hit.

## 8. Overview of the Star Language Reference Sections

The Star language contains both built-in language features and syntactic features that are used by users for creating Star language definitions. The following reference sections illustrate some of the main representation constructs using examples; other constructs are only briefly described. The Star language borrows some features from object-oriented programming languages such as C++; Star also uses an XML-style construct (described below). Note that within Star code listings, comments are preceded by "//". Within syntax sections, optional elements of an expression or statement are enclosed by '[' and ']'.

### 8.1. The XML Element Construct

Many Star language expressions and statements use "XML elements", which are expressions that resemble an XML element. An example is the attribute base expression shown here:

```
<Attribute ref = PersonWeight range = {10 .. 800} />
```

XML elements are also used within some statements in order to represent a variety of boolean flags.

### 8.2. Built-in Features

The built-in features include:

- Constant set name keywords
- Attribute super types
- Attribute value super types
- Attribute value types

### 8.3. Expressions

There are several expression types that are used by more than one Star language statement. They include the following:

- DictionaryExpression
- DictionaryPriorWordExpression
- AttributeBaseExpression
- DimensionSet Expression
- TwoPartAttributeCluster Expression
- AtLocation Expression
- OrientationSpecifierSet Expression
- OuterDimensionSystemExtents Expression

### 8.4. Statements

The Star language statements are described in detail in the following reference section. They include the following:

- Integer Declaration Statement
- Floating Point Declaration Statement
- String Declaration Statement
- Routine Statement
- Class Statement
- Value Set Statement
- Mapping Statement
- Transform Statement
- Attribute Type Statement
- Attribute Statement
- Relationship Type Statement *(detail not included)*
- Relationship Statement *(detail not included)*
- Dimension System (Type) Statement
- Specification System (Type) Statement
- Object Frame Class Statement
- Template Statement
- Populated Object Class Statement
- Behavior Class Statement

## 9. Reference: Built-in/Intrinsic Star Language Elements

The following items are predefined.

### 9.1. Constants and String Literal

Star contains two constant types and a string literal type:

- integerconstant – e.g. 0, 99
- floatingpointconstant – e.g. 0.8705
- stringliteral – e.g. "Person-5"

### 9.2. Constant Set Name Keywords

Star contains the following keywords that are names for built-in sets of constants: these names may appear in Star programs:

- IntegerConstant
- FloatingPointConstant  (although ROSS value sets are integer-based this is included for completeness)
- StringLiteral

### 9.3. Built-in Attribute Super Types

There are two pre-defined attribute super types, they are higher-level attribute type categories:
- Locational attribute types
- Qualitative attribute types

### 9.4. Built-in Attribute Value Set Super Types

There are also two pre-defined attribute value set "usage" super types, they correspond to the attribute type super types, and are:
- Locational attribute value set
- Qualitative attribute value set

### 9.5. Built-in Attribute Value Types

These are not attribute value set types, but are special categories for specific attribute values. They are:
- SpaceValue
- NonSpaceValue

These categories play a special role in instance models  and  in inference.

# 10. Reference: Star Language Expressions That Are Used by Multiple Statements

## 10.1. Dictionary and DictionaryPriorWord Expressions

### 10.1.1. Overview and Basic Form

The version of Star that is used for NLU applications contains an element called "Dictionary" that can be used in a variety of contexts. A dictionary associates a word or a set of words with a single concept. A dictionary is just a word list for the concept that it is associated with – there is no specification of a definition since the semantics of the word or words of a dictionary are described as part of the concept. The dictionary has the capacity for multiple language support. In the following example, a Dictionary construct is used within an AttributeType statement in order to create a set of English words for each vehicle exterior color value.

```
AttributeType "VehicleExteriorColor"
(
<SuperType val = "QualityAttributeType"/>

"Values"
(
 { "Black": Dictionary
    ( English
     ( { "black", "charcoal" } ); ); ,
   "Blue": Dictionary
    ( English
     ( { "blue" } ); ); ,
   "Silver": Dictionary
    ( English
     ( { "silver", "grey" } ); ); ,
   "White": Dictionary
    ( English
     ( { "white", "opal" } ); ); ,
  }
 );
);
```

Dictionaries are used by an NLU system in order to create an internal lexicon. A word that is defined within a dictionary element is not limited to use in that element: for instance, the word "opal" in the example here may exist in any of a number of other places within other Dictionary elements. Another example would involve the use of the verb forms for "to walk" – a particular ontology may associate the "walk" verb with one concept, e.g. "MotileLivingOrganismWalks" whereas another ontology might associate "walk" as a verb with each of several behavior classes, e.g. "PersonWalks", "AnimalWalks", "RobotWalks" and "PitcherWalksBatter". It is up to the NLU system to perform word-sense disambiguation and select the class that is appropriate for the natural language input and context.

### 10.1.2. Syntax

The syntax of the dictionary-related expressions is as follows:

```
DictionaryExpression  ->  DictionaryKeyword DictionaryInnerExpression ;

DictionaryPriorWordExpression  ->  DictionaryPriorWordKeyword DictionaryInnerExpression ;

DictionaryKeyword  ->  'Dictionary' ;

DictionaryPriorWordKeyword  ->  'DictionaryPriorWord' ;

DictionaryInnerExpression  ->

    '('
    NaturalLanguageNameKeyword
    '('
    WordList
    ')' ';' ;
    ')' ';' ;

NaturalLanguageNameKeyword  ->

    'English' |
    'German' |
    'Spanish' ;

    // others here

WordList  ->  UnrestrictedWordList | NounWordList | VerbWordList ;

UnrestrictedWordList  ->

    '{'
    InnerWordList
    '}'

InnerWordList  ->  Word
           |  Word ',' InnerWordList ;

Word  ->  '"' <string literal> '"' ;

NounWordList  ->

    '{'
    SingleNounFormsList
    '}'

SingleNounFormsList  ->  SingleNounFormsDepiction
            |  SingleNounFormsDepiction ',' SingleNounFormsList ;

SingleNounFormsDepiction  ->

    SingularNounForm ','
    PluralNounForm ;

VerbWordList  ->

    '{'
    SingleVerbFormsList
    '}'
```

```
SingleVerbFormsList  ->  SingleVerbFormsDepiction
              |  SingleVerbFormsDepiction ',' SingleVerbFormsList ;

SingleVerbFormsDepiction  ->

    InfinitiveVerbForm ','
    SimplePastVerbForm ','
    PastParticipleVerbForm ','
    SimplePresentThirdPersonSingularVerbForm ','
    PresentParticipleVerbForm ;
```

An example SingularNounForm is "person"; an example InfinitiveVerbForm is "walk".

### 10.1.3. Additional Examples

Additional examples are shown here:

```
(1) // (Unrestricted word list for a value within a value set called "ColorValueSet")

    "ColorBlue" :  Dictionary ( English ( { "blue", "turquoise" } ); );

(2) // (for the PersonObjectFrameClass)

    Dictionary
    (
       English
       (
          {
             "person",
             "persons",
             "human",
             "humans"
          }
       );
    );

(3) // (for the PersonHitsBehaviorClass)

    Dictionary
    (
       English
       (
          { "hit",    // (infinitive/base)
            "hit",    // (simple past)
            "hit",    // (past participle)
            "hits",   // (simple present, 3rd p.s.)
            "hitting", // (present participle)
            "punch",
            "punched",
            "punched",
            "punches",
            "punching"
          }
       );
    );

(4)  // (for the StereoSystemObjectFrameClass)
```

```
DictionaryPriorWord ( English
(
  {
    "stereo",
    "stereos" // (plural since it is a noun)
  }
););

Dictionary ( English
(
  {
    "system",
    "systems"
  }
););
```

An NLU system can make use of the ordinal position of noun and verb form words; e.g. for the PersonHitsBehaviorClass every $5^{th}$ word is the present participle form of the verb.

### 10.1.4. DictionaryPriorWord and DictionaryAntePriorWord

The DictionaryPriorWord construct is useful for building lexicons that contain two-word terms such as "stereo system". The DictionaryAntePriorWord construct allows for the specification of a third word (used as the first word in an expression).

## 10.2. AttributeBaseExpression

### 10.2.1. Overview and Basic Form

The *attribute base expression* is an expression that closely corresponds to an FOL atomic sentence that contains a predicate and a term that is a single constant. A ROSS attribute is a strongly typed two-part construct. It consists of a pre-defined attribute type name and an attribute value that is a member of a pre-defined attribute value set. The ROSS notion of attribute type and attribute value is roughly the same as that which has been in widespread use in software applications for many years, for instance, the attribute from the field of logical data modeling for databases.

An example attribute is:

<Attribute ref = VehicleExteriorColor val = "Silver" />

The attribute is composed of the "Attribute" keyword, then the "ref" keyword ("reference") followed by an equal sign and a defined attribute type name, and the "val" keyword ("value") followed by a value that had been defined as a member of an attribute value set (in this example, the attribute value set that was defined within an *attribute type* called "VehicleExteriorColor").

The use of attributes rather than predicates (as with logic) provides for a set of criteria for indexing. For instance, the logic assertion "E(x): Blue(x)" (which can be read "there exists an object such that the object is blue"), makes use of a predicate that actually corresponds to a Star attribute *value*, not the attribute type name. Attribute values do not provide a good basis for indexability since they may be members of very large sets. In contrast attribute types are more appropriate as criteria for indexing as it is needed to support queries of structured information. ROSS attributes are not limited to containing constant values: an attribute value may consist of a value *range*, a math expression, a relational expression, or a *routine* name that refers to a function that has been defined within the ROSS knowledge base. An attribute value may also be a reference to a bitmap file.

Attributes can exist within object frame classes and they can exist within object frame instances within fact transcripts and instance models. An example of an attribute that belongs with a class would exist within a class for gold coins: all instances of this class can be said to have the attribute of compositionality of gold material.

*Attribute value ranges* have an important use within definitions of object frame classes – they allow for a component to be located approximately within the parent structure. For instance, a class called "FrontEngineAutomobile" would specify that the engine compartment is situated within a certain section of the parent class (the Automobile). The class definition does not specify the exact location – it is specified using a range of values (e.g. within the range of 5 cm to 100 cm from the front end of the car). Instances that are based on the class can specify the exact location if this information is relevant.

### 10.2.2. Syntax

```
AttributeBaseExpression  ->

    '<' 'Attribute' 'ref' '=' AttributeTypeName AttributeBaseValueExpression '/' '>' ;

AttributeBaseValueExpression  ->

    SimpleValueExpression
    | ValueRangeExpression
    | VariableDeclarationExpression
    | MathExpression
    | RelationalExpression
    | RoutineNameExpression
    | BitmapNameExpression ;

SimpleValueExpression  -> 'val' '=' '"' integerconstant '"' ;

ValueRangeExpression  -> 'range' '=' '{' integerconstant '..' integerconstant '}' ;

VariableDeclarationExpression  -> 'var' '=' identifier ;

MathExpression  -> 'expr' '=' MathInnerExpression ;

MathInnerExpression  -> // cf. examples

RelationalExpression  -> 'expr' '=' RelationalInnerExpression ;

RelationalInnerExpression  -> // cf. examples
```

```
RoutineNameExpression  ->  'routine' '=' '"' RoutineName '"' ;

RoutineName  ->  identifier ;

BitmapNameExpression  ->  'bitmap' '=' '"' BitmapFileName '"' ;

BitmapFileName  ->  identifier ;
```

### 10.2.3. Examples

(1) Using "val" with value:

    `<Attribute ref = X-Coordinate val = "450" />`

(2) Using "range" with value range:

    `<Attribute ref = PersonWeight range = {10 .. 800} />`

(3) Using "var" with an operand that contains a variable:

    `<Attribute ref = AttributeTypeX var = x$ />`

(4) Using "expr" with a math expression:

    `<Attribute ref = RelativePositionX expr = (x$ + 1) />`

(5) Using "expr" with a relational expression:

    `<Attribute ref = AttributeTypeX expr = (x1$ < (x$ - 36)) />`

(6) Using "routine" with named rendering routine:

    `<Attribute ref = EssentialValueAttributeType routine = "RenderAnimalHead" />`

(7) Using "bitmap" with bitmap file name:

    `<Attribute ref = EssentialValueAttributeType bitmap = "AnimalHead3D.dat" />`

## 10.3.    Dimension Set Expression

### 10.3.1. Overview

The dimension set expression consists of a set of attribute types that are used in a coordinated way to specify the location of a unit-sized object frame class or object instance. A dimension set expression is also used to specify an anchor point location for aggregate object frame classes or instances. This expression type is used in several places. First, it is used within a RelationshipToParent section of an object frame class – in expressions within an AtLocations section, and in expressions within an OuterDimensionSystemExtents section. Second, it is used by the RelationshipType statement. Finally it can be used within a TemplateObjectClass statement.

### 10.3.2. Syntax

```
DimensionSetExpression  ->

    DimensionSystemNameExpression
    AttributeBaseExpressionList ;

DimensionSystemNameExpression  ->

    '<' 'DimensionSystem' 'ref' '=' DimensionSystemName '/' '>' ;

DimensionSystemName  ->  identifier ;

AttributeBaseExpressionList  ->  AttributeBaseExpression
                              |  AttributeBaseExpression AttributeBaseExpressionList ;
```

### 10.3.3. Examples

The following dimension set expression is a specification of a specific location using millimeter x,y,z coordinates:

```
<DimensionSystem ref = PhysicalObjectMillimeterCoordinates />
<Attribute ref = X-Coordinate val = "20" />
<Attribute ref = Y-Coordinate val = "20" />
<Attribute ref = Z--Coordinate val = "35" />
```

## 10.4.   Two Part Attribute Cluster Expression

### 10.4.1. Overview and Basic Form

The ROSS *two-part attribute cluster* is a conceptual feature that can take any of several forms. The two-part attribute cluster satisfies the intuitive concept of a *fully specified fact*: it represents both the *location* and the *value* of an entity that exists in a 4D represented world. A two-part attribute cluster can exist within a class definition in a knowledge base or it may exist within a fact repository artifact (e.g. an instance model). The Star language implementation of the two-part attribute cluster is a representational construct that consists of at least one attribute from the set of locational attribute super-types, and at least one attribute from the set of qualitative attribute super-types. The rationale behind this requirement is that it produces ROSS expressions that *fully* describe entities from the represented world.

The two-part attribute cluster is the equivalent of a set of propositions or assertions in logic; where these assertions would include one or more propositions that represent the *location* of an entity and one or more propositions that represent the *value* of the same entity.

The following is an example of a two-part attribute cluster. (The Star language fragment also shows several preliminary definitions, followed by instance model pseudo-code that includes an attachment statement wherein an object frame instance is instantiated). The assert statement contains the two-part attribute cluster expression. This is a very simple example as might be used for children's stories; for the sake of brevity it does not show the structural parent and "RelationshipToParent" infrastructure for the vehicle class or car instance.

```
// Star Definitions

ObjectFrameClass VehicleObjectClass
{
  AttributeTypes
  (
    AttributeType "SpatialLocation"
    (
      <SuperType val = "Locational"/>

      "Values"
      (
        "Garage",
        "Driveway",
        "Roadway",
      );
    );

    AttributeType "Color"
    (
      <SuperType val = "Qualitative"/>

      "Values"
      (
        "Red",
        "Green",
        "Blue"
      );
    );
  );
};

//-------------------------------------------------------------------------
//  (THIS SECTION IS PART OF AN INSTANCE MODEL)
//
// Attachments (Object Instantiations)

attach VehicleObjectClass Car1;

// Assertion:

assert Car1::
  ( <Attribute ref = SpatialLocation
          val = "Driveway" />,
    <Attribute ref = Color
          val = "Blue" />
  );
```

The "assert" statement contains an expression that is the two-part attribute cluster: it can be interpreted as "the entity at the location called "Driveway" has a color value of "Blue". The essential features of a two-part attribute cluster are illustrated here: it contains at least one *locational* attribute that specifies the location of the object frame instance (the entity), and at least one *value* attribute that specifies the *infused* or *populated* value of the object frame instance.

### 10.4.2. Syntax

```
TwoPartAttributeClusterExpression  ->

    TwoPartAttributeClusterUsingDimensionSystem |
    TwoPartAttributeClusterUsingSpecificationSystem ;

TwoPartAttributeClusterUsingDimensionSystem  ->

    DimensionSetExpression
    InnerContentExpression ;

InnerContentExpression  ->  QualityAttributeBaseExpressionList ;

QualityAttributeBaseExpressionList  ->  QualityAttributeBaseExpression
                    |  QualityAttributeBaseExpression QualityAttributeBaseExpressionList ;

QualityAttributeBaseExpression  ->  // an attribute base expression consisting of quality attributes

TwoPartAttributeClusterUsingSpecificationSystem  ->

    SpecificationSystemNameExpression
    LocationAttributeBaseExpressionList
    InnerContentExpression ;

SpecificationSystemNameExpression  ->

    '<' 'SpecificationSystem' 'ref' '=' SpecificationSystemName '/' '>' ;

SpecificationSystemName  ->  identifier ;

LocationAttributeBaseExpressionList  ->  LocationAttributeBaseExpression
                    |  LocationAttributeBaseExpression LocationAttributeBaseExpressionList ;
```

### 10.4.3. Examples

Two-part attribute cluster expressions are used within PopulatedObjectClasses and within TemplateClasses. The first example is a two-part attribute expression as it would exist in a PopulatedObjectClass that describes attributes of a house cat. The second example is a two-part attribute expression that might exist in a TemplateObjectClass.

(1) Using a dimension system. The main constituent elements are the dimension set expression and the inner content expression, which is a list of qualitative attributes.

```
// DimensionSetExpression:
<DimensionSystem ref = PhysicalObjectMillimeterCoordinates />
<Attribute ref = X-Coordinate var = x$ />
<Attribute ref = Y-Coordinate var = y$ />
<Attribute ref = Z-Coordinate var = z$ />
// Qualitative attributes:
<Attribute ref = ExteriorColor val = "Brown" />
<Attribute ref = StandingState val = "Sitting" />
```

(2) Using a specification system. The main constituent elements are the specification system reference and the list of attributes.

```
<SpecificationSystem ref = AnimalPhysicalComposition />
// Location attributes:
 <Attribute ref = X-Coordinate var = x$ />
 <Attribute ref = Y-Coordinate var = y$ />
 <Attribute ref = Z-Coordinate var = z$ />
 // Qualitative attribute:
 <Attribute ref = EssentialValueAttributeType bitmap = "HouseCat3D-01.dat" />
```

### 10.5.    AtLocationSet Expression

The AtLocationSet expression is the first of three expressions that are usually used in combination in order to specify the location, size and spatial orientation of an object frame class (the others are the OrientationSpecifierSet and the OuterDimensionSystemExtentSet). An AtLocationSet is an instance of a dimension set expression. The attributes contained within the expression refer to a "frame", not to the object that is contained in the frame. For instance, a PersonObjectFrameClass is a rectangular (if it has only 2D dimensions) or cuboid (3D) frame: it can *contain* a person (with a specific shape and set of internal attributes), but the frame itself is only a sort of 3D outline.

When used to specify the "at locations" of a unit-sized object frame class or instance, it refers to the location of that unit location. When used to specify the location of an aggregation of unit-sized classes or instances, the at location attribute set is a specification of a particular point within the cuboid region, referred to as the "anchor point".

### 10.5.1. Basic Form

This example illustrates the AtLocationSet expression in its basic form, as would be used to specify the location of a unit object frame class or instance:

```
AtLocationSet
(
    <DimensionSystem ref = PhysicalObjectMillimeterCoordinates />
    <Attribute ref = X-Coordinate val = "20" />
    <Attribute ref = Y-Coordinate val = "20" />
    <Attribute ref = Z-Coordinate val = "0" />
);
```

### 10.5.2. Examples

(1) Basic: this shows an AtLocationSet within an AtLocations section that would exist within a RelationshipToParent section in an ObjectFrameClass statement.

```
AtLocations
(
    AtLocationSet  // a DimensionSetExpression
    (
        <DimensionSystem ref = PersonObjectHolder />
        <Attribute ref = RelativePlace val = "PersonHeadReceptacle" />
    );
);
```

(2) Advanced, as would be used for an aggregate object frame, as it specifies an anchorpoint.

```
AtLocationSet
(
    <DimensionSystemType val = "CartesianCoordinates" />
    <AnchorPoint type = "Numeric" val = "[0,0,0]" />
    <DimensionSystem ref = PhysicalObjectMillimeterCoordinates />
    <Attribute ref = X-Coordinate val = "20" />
    <Attribute ref = Y-Coordinate val = "20" />
    <Attribute ref = Z-Coordinate val = "0" />
);
```

(3) Advanced, showing two at location sets in an AtLocations section as would appear within an ObjectFrameClass statement. A transform would be needed to derive and transform a specific set of values from those specified within either at location set to the other.

```
AtLocations
(
    AtLocationSet
    (
        <DimensionSystem ref = PersonObjectHolder />
        <Attribute ref = RelativePlace val = "PersonHeadReceptacle" />
    );
    AtLocationSet  // (the Origin)  // [0,0,0] situated @ [20,20,0]
    (
        <DimensionSystemType val = "CartesianCoordinates" />
        <AnchorPoint type = "Numeric" val = "[0,0,0]" />
        <DimensionSystem ref = PhysicalObjectMillimeterCoordinates />
        <Attribute ref = X-Coordinate val = "20" />
        <Attribute ref = Y-Coordinate val = "20" />
        <Attribute ref = Z-Coordinate val = "0" />
    );
);
```

## 10.6.    OrientationSpecifierSet Expression

*(Details about the OrientationSpecifierSet expression are not contained in this version).*

## 10.7.    OuterDimensionSystemExtentSet Expression

The OuterDimensionSystemExtentSet expression is used in order to specify the size ("extent") of an object frame class or instance. Like the AtLocationSet, an OuterDimensionSystemExtentSet is an instance of a dimension set expression. The name of this expression is intended to describe the fact that it uses the dimension system of the "outer" (i.e. parent) object frame class.

### 10.7.1. Basic Form

This example shows an outer dimension system extent set as would appear within an ObjectFrameClass statement.

```
OuterDimensionSystemExtents
```

```
(
    OuterDimensionSystemExtentSet
    (
        <DimensionSystem ref = AnimalComponentMillimeterCoordinates />
        <Attribute ref = X-Coordinate val = "nil" />
        <Attribute ref = Y-Coordinate val = "nil" />
        <Attribute ref = Z-Coordinate val = "nil" />
    );
);
```

# 11. Reference: Star Language Statements

## 11.1.    Integer Declaration Statement

The integer declaration statement declares an integer constant.

### 11.1.1. Syntax

DeclarationInteger  ->  IntegerKeyword IntegerName '=' IntegerSimpleExpression  ;

IntegerKeyword  ->  'Integer' ;

IntegerName  ->  identifier ;

IntegerSimpleExpression  ->  integerconstant

### 11.1.2. Example

Integer lenMaxVehiclePhysicalDimension = 12000;

## 11.2.    Floating Point Declaration Statement

The floating point number declaration statement declares a floating point number constant.

### 11.2.1. Syntax

DeclarationFloatingPoint  ->  FloatingPointKeyword FloatingPointName '=' FloatingPointSimpleExpression  ;

FloatingPointKeyword  ->  'FloatingPoint' ;

FloatingPointName  ->  identifier ;

FloatingPointSimpleExpression  ->  floatingpointconstant

### 11.2.2. Example

FloatingPoint approximateAge = 50.3;

## 11.3.    String Declaration Statement

The string declaration statement declares a string literal.

### 11.3.1. Syntax

DeclarationString  ->  StringKeyword StringName '=' StringLiteral ;

StringKeyword  ->  'String' ;

StringName  ->  identifier ;

StringLiteral  ->  stringliteralconstant

### 11.3.2. Example

```
String countryNameUnitedStates = "United States of America" ;
```

## 11.4.  Routine Statement

*(Details about the Routine statement are not contained in this version: see the Transform statement for an example).*

## 11.5.  Value Set Statement

### 11.5.1. Overview and Basic Form

A value set (or "attribute value set") statement allows for the creation of a set of values that can be used in attribute expressions. An attribute value set is defined using the "ValueSet" keyword, followed by a value set name, and then by a value set expression. The following is a basic example that defines two value sets, "Millimeter" and "VehiclePhysicalDimension", which uses Millimeter. A declaration is also included here for the purpose of defining a constant value (the maximum length of a vehicle dimension). *VehiclePhysicalDimension* is a value set that will be used for *locational* attributes.

```
ValueSet "Millimeter"
(
  IntegerConstant
);

Integer lenMaxVehiclePhysicalDimension = 12000;   // vehicle max size in any dimension

ValueSet "VehiclePhysicalDimension"
(
    <BaseValueSet ref = Millimeter />  // the unit of measure

    <SuperTypeUsage val = "Locational" />

    { 1, .. lenMaxVehiclePhysicalDimension }
);
```

The attribute value sets defined here can subsequently be used in other statements and expressions as needed. The following value set – "VehicleComponentMaterialComposition" – is one that will be used for *value* (also called "quality") attributes rather than locational attributes.

```
ValueSet "VehicleComponentMaterialComposition"
(
    <SuperTypeUsage val = "Qualitative" />

    // The following are quality values:

    { "Space",
      "SolidUnspecified",
      "Metal",
      "Plastic" }
);
```

Other forms exist; a partial BNF grammar is in the syntax section that follows.

## 11.5.2. Restrictions for Numeric Value Sets

Numeric attribute values are either natural numbers or integers. Numeric attribute value sets that are used for locational attribute types must be finite subsets of the set of integers. Numeric attribute value sets that are used for qualitative attribute types must be finite subsets of the set of natural numbers. *(Within the primary information section of a fact repository artifact, where data or natural language text represents real numbers, or where values are computed (e.g. by division) to yield a real number, rounding or truncation of numeric values must take place).*

## 11.5.3. Syntax

The syntax of the value set statement is as follows:

```
ValueSetStatement  -> [ValueSetKeyword] ValueSetName ValueSetExpression;

// (note: ValueSetKeyword is optional when this statement appears within an AttributeType statement)

ValueSetKeyword  -> 'ValueSet' ;

ValueSetName  -> '"' identifier '"' ;

ValueSetExpression  -> ValueSetRenameExpression
         | ValueSetNormalExpression;

ValueSetRenameExpression  -> '(' UserDefinedValueSetBaseName ')' ';' ;

UserDefinedValueSetBaseName  -> identifier;

ValueSetNormalExpression  -> ValueSetNormalExpressionUsingConstantSetName  // BaseTypeName
         | ValueSetNormalExpressionUsingSetDepiction
         | ValueSetNormalExpressionUsingEnumeratedValues;

ValueSetNormalExpressionUsingConstantSetName  -> '(' PreDefinedStarLanguageConstantSetName ')' ';' ;

PreDefinedStarLanguageConstantSetName  -> 'IntegerConstant' | 'FloatingPointConstant' | 'StringLiteral' ;;

ValueSetNormalExpressionUsingSetDepiction  ->

  '('
  [ XMLElementBaseValueSet ]
  [ XMLElementSuperTypeUsage ]
  [ XMLElementOrderedCollection ]
  SetDepictionExpression
  ')' ';' ;

XMLElementBaseValueSet    -> '<' "BaseValueSet" "ref" '=' UserDefinedValueSetBaseName '/' '>' ;

XMLElementSuperTypeUsage    -> '<' "SuperTypeUsage" "val" '='
PreDefinedStarLanguageValueSetSuperTypeName '/' '>' ;
```

```
PreDefinedStarLanguageValueSetSuperTypeName  ->  "LocationalValues"
                           |   "QualitativeValues" ;

XMLElementOrderedCollection  ->  '<' "OrderedCollection" "val" '=' BooleanStringValue '/' '>' ;

BooleanStringValue  ->  "true"
             |   "false" ;

SetDepictionExpression  ->  '{' LowerValueConstant '..' UpperValueConstant '}' ;

LowerValueConstant  ->  integerconstant | floatingpointconstant ;
UpperValueConstant  ->  integerconstant | floatingpointconstant ;

ValueSetNormalExpressionUsingEnumeratedValues  ->

  '{'
  [ XMLElementSuperTypeUsage ]
  [ XMLElementOrderedCollection ]
  EnumeratedValuesExpression
  '}' ';' ;

EnumeratedValuesExpression  ->  '{' EnumeratedValueList '}' ;

EnumeratedValueList  ->  EnumeratedValueExpression
              |   EnumeratedValueExpression ',' EnumeratedValueList ;

EnumeratedValueExpression  // (see examples)
```

## 11.5.4. Additional Examples

(1) Using a "rename" (i.e. a user-defined value set base name)

```
ValueSet "MyPrimeNumberValueSet" (PrimeNumberValueSet);
```

(2) Using a base type (built-in Star language constant set name)

```
ValueSet "MyIntegerValueSet"
(
    IntegerConstant
);
```

(3) Using a set depiction

```
"MyValueSetName"
(
    <BaseValueSet ref = Millimeter />
    <SuperTypeUsage val = "LocationalValues" />
    { 1, .. 9999 }
);
```

(4) Using enumerated values

```
"SimpleTimelineValueSet"
(
    <SuperTypeUsage val = "LocationalValues" />
    <OrderedCollection val = "true" />
    { "T01",
```

```
        "T02" }
);

"ChemicalCompositionValueSet"
(
    <SuperTypeUsage val = "QualitativeValues" />
    { "Organic",
      "InOrganic" }
);
```

## 11.6.    Mapping Statement

Mapping statements are used to allow mapping of members of one value set to members of another value set. The function expression specifies the computation that must be performed to map a value from source set to destination set.

### 11.6.1. Basic Form

(In the example here, the value sets named "Meter" and "Foot" have already been defined).

```
Mapping "MeterToFoot"
(
  <Source ref = Meter />
  <Dest ref = Foot />
  <Function expr = (x$ * 3.2808) />
);
```

This can now be used by system processing components that need to convert meter values to foot values.

### 11.6.2. Syntax

```
MappingStatement  ->  MappingKeyword MappingName MappingExpression;

MappingKeyword  ->  'Mapping' ;

MappingName  ->  '"' identifier '"' ;

MappingExpression  ->

    '('
    [ XMLElementSourceValueSetReference ]
    [ XMLElementDestValueSetReference ]
    [ XMLElementSourceFunctionExpression ]
    ')' ';' ;

XMLElementSourceValueSetReference  ->  '<' "Source" "ref" '=' ValueSetName '/' '>' ;

ValueSetName  ->  identifier;

XMLElementSourceValueSetReference  ->  '<' "Dest" "ref" '=' ValueSetName '/' '>' ;

XMLElementSourceFunctionExpression  ->  '<' "Function" "expr" '=' FunctionExpression '/' '>' ;

FunctionExpression  ->  // (see examples)
```

### 11.6.3. Examples

The following is another example:

```
ValueSet "Meter"
(
    IntegerConstant
);

Mapping "MillimeterToMeter"
(
    <Source ref = Millimeter />
    <Dest ref = Meter />
    <Function expr = (x$ / 1000) />
);
```

## 11.7.    Transform Statement

The transform statement allows for the mapping of a set of locational attributes from one dimension system that specify a specific location to a set of attributes of another dimension system that specify the same location.

The capability for translating or mapping between different dimension systems is based on the author's perception of how human memory and cognition work. Humans seem to have a general-purpose three-dimensional frame of reference that underlies perhaps all cognitive representations. The capability for representing the location of things in the physical world often involves smaller "customized" frames of reference. An example would be a mental representation that a particular house is at 1000 State Street, in some city, in a particular state or province, within a particular country, etc.. The custom frame of reference has a dimension system that consists of country, region, city identification (name), street name and street number. However there is a mental capacity for going back and forth between this custom representation and the master mental 3D frame of reference. The transform statement allows for this type of conversion between dimension systems.

### 11.7.1. Syntax

```
TransformStatement  ->  TransformKeyword TransformName TransformExpression;

TransformKeyword  ->  'Transform' ;

TransformName  ->  '"' identifier '"' ;

TransformExpression  ->

    '('
    [ XMLElementSourceDimensionSystemReference ]
    [ XMLElementDestDimensionSystemReference ]
    RoutineStatement
    ')' ';' ;
```

### 11.7.2. Example

*Background:* Many of the object frame classes of the main ROSS ontology use an upper ontology class called *EverydayObjectStructuralParent* as their structural parent class. They in turn are used in behavior classes that use a class called *BehavioralStructuralParentClass* as the behavior class's structural parent; therefore it is necessary to define a routine that performs transformations of a set of coordinates from the dimension system of the one structural parent to the other. The following only shows the Transform statement for the spatial coordinates (the temporal transform is not shown). (Note that this is only one of many transforms that might be defined for conversions between the dimension systems in question).

```
// Transform for:  (source) BehavioralStructuralParentClass.RelativePosition to
   (dest) EverydayObjectStructuralParent.EverydayObjectSpatialCoordinates, which is based upon
   PhysicalObjectMillimeterCoordinates:

        Transform "RelativePositionSpatialToMillimeterBasedCoords-01"
        (
            <Source ref = RelativePosition.SpatialAttributeTypes />
            <Dest ref = PhysicalObjectMillimeterCoordinates.SpatialAttributeTypes />

            bool Routine
            {
              Parameters
              (
                 string Source;   // one of: "IdenticalLocation", "Adjacent", "NotAdjacent"
                 int Dest[3];
              );

              Locals
              (
                 int x = 0;
                 int y = 0;
                 int z = 0;
              );

              if (Source == "IdenticalLocation")
              {
                 Dest [x] = 0;
                 Dest [y] = 0;
                 Dest [z] = 0;
              }
              else if (Source == "Adjacent")
              {
                 Dest [x] = 2;  // arbitrary distance of 2 millimeters
                 Dest [y] = 0;
                 Dest [z] = 0;
              }
              else if (Source == "NotAdjacent")
              {
                 Dest [x] = 1000;  // arbitrary distance of 1000 millimeters
                 Dest [y] = 0;
                 Dest [z] = 0;
              }

              Return true;
            }
```

```
);
```

## 11.8.   Attribute Type Statement

The attribute type statement defines an attribute type. Once an attribute type has been defined, the defined attribute type name can then be used in other statements and expressions as needed. Where the attribute type is used, type checking can be performed for values that derive from the attribute value set.

### 11.8.1. Basic Form

The basic form of the attribute type is shown by this example: in this case the attribute value set is explicitly defined within the attribute type statement. Alternately, the attribute value set may be referred to by name.

```
AttributeType "VehicleExteriorColor"
(
<SuperType val = "QualityAttributeType"/>

"Values"
(
 { "Black": Dictionary
    ( English
      ( { "black", "charcoal" } ); ); ,
   "Blue": Dictionary
    ( English
      ( { "blue" } ); ); ,
   "Silver": Dictionary
    ( English
      ( { "silver", "grey" } ); ); ,
   "White": Dictionary
    ( English
      ( { "white", "opal" } ); );
 }
 );
);
```

### 11.8.2. Additional Features

The attribute type has the following optional features:

- ProbabilityInObjectFrameClass field: this field specifies a value (between 0 and 1) that represents the probability that an encountered instance is an instance of the object frame class that contains the attribute type (see following section: Use of ProbabilityInObjectFrameClass Field).
- AttributeSuperType field: this field is used to distinguish locational attribute types from qualitative attribute types.
- StateAttribute boolean field: if true, the attribute type contains values that are representations of states of an object; such attribute types are used within behavior classes.

- OptionalCausalFeature boolean field: a qualitative attribute type may be an optional causal feature: this has significance only within a behavior class where the attribute type is used (use of this field is described in more detail in the section on the behavior class statement).

### 11.8.3. Syntax

AttributeTypeStatement  ->  AttributeTypeKeyword AttributeTypeName AttributeTypeExpression;

AttributeTypeKeyword  ->  'AttributeType' ;

AttributeTypeName  ->  '"' identifier '"' ;

AttributeTypeExpression  ->

  '('
  [ XMLElementProbabilityInObjectFrameClass ]
  [ XMLElementAttributeSuperType ]
  [ XMLElementStateAttributeTypeBooleanFlag ]
  [ OptionalCausalFeatureBooleanFlag ]
  ValueSetConstruct
  ')' ';' ;

XMLElementProbabilityInObjectFrameClass  ->  '<' "Probability" "expr" '=' floatingpointconstant '/' '>' ;

XMLElementAttributeSuperType  ->  '<' "SuperType" "val" '=' AttributeSuperType '/' '>' ;

AttributeSuperType  ->  "Locational" | "Qualitative" ;

XMLElementStateAttributeTypeBooleanFlag  ->  '<' "StateAttributeType" "val" '=' BooleanStringValue '/' '>' ;

OptionalCausalFeatureBooleanFlag  ->  '<' "OptionalCausalFeature" "val" '=' BooleanStringValue '/' '>' ;

ValueSetConstruct  ->  // (see examples)

### 11.8.4. Examples

Here are additional examples of attribute type definitions: in this case the value sets called "VehiclePhysicalDimension" and "VehicleComponentMaterialComposition" have been previously defined. These definitions would exist to support a VehicleObjectFrameClass.

```
AttributeType "VehiclePhysicalDimensionAttributeTypeX"
(
    <SuperType val = "LocationAttributeType"/>
    <ValueSetName ref = VehiclePhysicalDimension/>
);

AttributeType "VehiclePhysicalDimensionAttributeTypeY"
(
    <SuperType val = "LocationAttributeType"/>
    <ValueSetName ref = VehiclePhysicalDimension/>
);

AttributeType "VehiclePhysicalDimensionAttributeTypeZ"
(
```

```
    <SuperType val = "LocationAttributeType"/>
    <ValueSetName ref = VehiclePhysicalDimension/>
);

AttributeType "VehicleComponentMaterialCompositionAttributeType"
(
    <SuperType val = "QualityAttributeType"/>
    <ValueSetName ref = VehicleComponentMaterialComposition/>
);
```

The following is a qualitative attribute type which includes "Dictionary" items which may be used by an NLU system:

```
AttributeType "PersonAge"
(
    <SuperType val = "Qualitative"/>

    "Values"
    (
      { "Infantile" :  Dictionary ( English ( { "infant" } ); ); ,
        "YoungChild" :  Dictionary ( English ( { "young" } ); ); ,
        "Child" :  Dictionary ( English ( { "young" } ); ); ,
        "Teenager" :  Dictionary ( English ( { "teenage" } ); ); ,
        "Adult" :  Dictionary ( English ( { "adult" } ); ); ,
        "MiddleAgedAdult" :  Dictionary ( English ( { "middle-aged", "adult" } ); ); ,
        "AdvancedAgedAdult" :  Dictionary ( English ( { "elderly", "senior", "older", "old" } ); ); 
      }
    );
);
```

The next example is of a state attribute type that is useful for "person walks" behavior classes: this particular attribute type could belong to a class called PersonObjectFrameClass.

```
AttributeType "WalkingState"
(
    <SuperType val = "Qualitative"/>
    <StateAttributeType val = "true"/>

    "Values"
    (
      { "NotWalking",
        "Walking"
      }
    );
);
```

### 11.8.5. Use of ProbabilityInObjectFrameClass Field

The following code shows two object frame classes: a "hospital resident doctor" class and a "hospital resident patient" class. The "doctor" class has an attribute type called "AttributeTypeOnDutyState". Each of the two classes has an attribute type called "AttributeTypeIllnessState". When used in an NLU system, the probability field within the attribute type indicates the probability that instances of the word "resident" that are encountered have an *association* with the attribute type. This information is useful for class selection (word-sense disambiguation), e.g. given texts such as "The active resident examined the patient", and "The resident is very sick", it is more likely that a "resident" that is described as being sick is an instance of the "hospital resident patient" class than of the "hospital resident doctor" class.

```
ObjectFrameClass "HospitalResidentDoctorObjectFrameClass"  // a resident doctor
(
    <StructureTrait val = "Compound"/>
    Dictionary ( English
    (
      { "resident",
        "residents" }
    ););
    HigherClasses ( { "PersonObjectFrameClass" } );

    AttributeTypes
    (
      AttributeType "AttributeTypeOnDutyState"
      (
          <Probability expr = 0.8 />
          <SuperType val = "Qualitative"/>

          "Values"
          (
            {
              "NotOnDuty",
              "OnDuty" : Dictionary ( English ( { "active" } ); );
            }
          );
      );

      AttributeType "AttributeTypeIllnessState"
      (
          <Probability expr = 0.01 />
          <SuperType val = "Qualitative"/>

          "Values"
          (
            {
              "NotIll",
              "Ill" : Dictionary ( English ( { "sick" } ); );
            }
          );
      );
    );

); // ObjectFrameClass "HospitalResidentDoctorObjectFrameClass"

ObjectFrameClass "HospitalResidentPatientObjectFrameClass"  // a resident patient
(
    <StructureTrait val = "Compound"/>
    Dictionary ( English
    (
      { "resident",
        "residents" }
    ););
    HigherClasses ( { "PersonObjectFrameClass" } );

    AttributeTypes
```

```
(
    AttributeType "AttributeTypeIllnessState"
    (
        <Probability expr = 0.9 />
        <SuperType val = "Qualitative"/>

        "Values"
        (
            {
                "NotIll",
                "Ill" : Dictionary ( English ( { "sick" } ); );
            }
        );
    );
);

); // ObjectFrameClass "HospitalResidentPatientObjectFrameClass"
```

## 11.9. Attribute Statement

The attribute statement allows for the definition of an attribute for an object frame class. The attribute thus specified strictly applies to all members of the class. The Probability field allows for the specification of a probability value that indicates the probability that a given instance of the object frame class has the attribute.

### 11.9.1. Basic Form

The following attribute statement could exist within a person object frame class. (This shows the Attribute statement within an enclosing "Attributes" section). This demonstrates the utility of the attribute range feature: a person class describes instances that have a body weight that within the range of 0 to 800 (pounds).

```
Attributes
(
    Attribute "BodyWeight"
    (
        <Attribute ref = BodyWeightAttributeType range = { 0 .. 800 } />
    );
);
```

### 11.9.2. Syntax

```
AttributeStatement  ->  AttributeKeyword AttributeName AttributeExpression;

AttributeKeyword  ->  'Attribute' ;

AttributeName  ->  '"' identifier '"' ;

AttributeExpression  ->  '(' AttributeBaseExpression ')' ';' ;
```

### 11.9.3. Examples

The following is another variation of the "body weight" attribute (also for a PersonObjectFrameClass); it represents the fact that 96% of the instances of this class are expected to have a bodyweight in the range of 20 to 250 (the unit of measure would be defined via the BodyWeightAttributeType, and here refers to pounds).

```
Attribute "BodyWeight"
(
    <Probability expr = 0.96 />

    <Attribute ref = BodyWeightAttributeType range = { 20 .. 250 } />
);
```

## 11.10.  Dimension System (Type) Statement

The dimension system type (or, just "dimension system") definition statement creates a dimension system (for instance, a coordinate system), that is used by object frame classes and instances. The dimension system is a mechanism for aggregating attribute types that are intended for collective use into a group in order to fully describe the locational attributes of an object frame instance. The expression that uses the attribute types in order to specify a specific set of attributes is referred to as a "dimension set expression", which has already been described.

A dimension system consists of a set of related *location* attribute types. For instance, these might involve three spatial (Cartesian) coordinates: an x-coordinate, a y-coordinate, a z-coordinate, and the added dimension of time. The set of location attributes are used within a dimension set expression that is based on the dimension system; the attributes describe *where* an entity is in space and time. When a dimension system definition is used in generating or creating an instance model that contains dimension set expressions, type checking can be performed (e.g. by an NLU semantic engine) to ensure that each required attribute type is actually used. This ensures that the generated specification expressions conform to the ROSS requirements for the specification of structure.

An example dimension system type for the description of geographical positions would involve attribute type definitions for each of latitude and longitude.

### 11.10.1.        Basic Form

The basic form of a dimension system is illustrated here:

```
Integer lenMaxPhysicalDimension = 1000000000;
// (1 million meters is large enough for the intended uses
//  of this dimension system)

DimensionSystem "PhysicalObjectMillimeterCoordinates"
(
    LocationAttributeTypes
    (
        SpatialAttributeTypes
        (
            "X-Coordinate"
            (
```

```
                    <SuperType val = "Locational"/>
                    "ValueSet"
                    (
                        <BaseValueSet ref = Millimeter />
                        <SuperTypeUsage val = "LocationalValues" />
                        { 1, .. lenMaxPhysicalDimension }
                    );
                );
                "Y-Coordinate"
                (
                    <SuperType val = "Locational"/>
                    "ValueSet"
                    (
                        <BaseValueSet ref = Millimeter />
                        <SuperTypeUsage val = "LocationalValues" />
                        { 1, .. lenMaxPhysicalDimension }
                    );
                );
                "Z-Coordinate"
                (
                    <SuperType val = "Locational"/>
                    "ValueSet"
                    (
                        <BaseValueSet ref = Millimeter />
                        <SuperTypeUsage val = "LocationalValues" />
                        { 1, .. lenMaxPhysicalDimension }
                    );
                );
            );
    );
); // DimensionSystem "PhysicalObjectMillimeterCoordinates"
```

## 11.10.2.    Example

The following example dimension system defines a set of coordinates for components of a vehicle. (The detail of each of the attribute types – "VehiclePhysicalDimensionAttributeTypeX", etc. is not shown because they have been defined with definitions within the same scope).

```
DimensionSystem "VehicleComponentCoordinates"
(
    LocationAttributeTypes
    (
        SpatialAttributeTypes
        (
            VehiclePhysicalDimensionAttributeTypeX;
            VehiclePhysicalDimensionAttributeTypeY;
            VehiclePhysicalDimensionAttributeTypeZ;
        );
        TemporalAttributeTypes
        (
            SecondBasedTimelineAttributeType;
        );
    );
);
```

## 11.11.  Specification System (Type) Statement

The specification system type definition statement incorporates a dimension system and an *inner content* section in order to create a system that can be used for fully specifying the place and *qualitative value* of unit-sized or aggregate object frame instances. The dimension system has already been described. The inner content section either defines a set of attribute types that describes the value of an entity (e.g. the car is blue, the ignition key is made of steel), or it is a specification of component-wise structure. A *specification set expression* can use a specification system similar to how a dimension set expression uses a dimension system.

The basic form of a specification system is illustrated here; this specification system uses the "VehicleComponentCoordinates" dimension system that was defined in the previous section.

```
SpecificationSystem "VehicleComponentPhysicalComposition"
(
    DimensionSystem "VehicleComponentCoordinates" (MillimeterCoordinates);

    InnerContent
    (
        QualityAttributeTypes
        (
            "EssentialValueAttributeType" (VehicleComponentMaterialCompositionAttributeType);
        );
    );
);
```

The upcoming description of the template class statement will illustrate the use of a specification system.

## 11.12.  Object Frame Class Statement

The object frame class definition statement is typically used in order to represent a spatially adjacent aggregation of unit-size objects. An object frame class may also represent a single unit-sized object frame. When it represents an aggregation of such units it has the shape of a 3D cuboid.

The object frame class is the foundation for the representation of the instances that get instantiated and thus exist in a fact transcript or in an NLU instance model. Object frame classes are also used within definitions (within other object frame classes) and are referred to within populated object classes in behavior classes.

The main features of the object frame class are listed here. (This uses an example of an object frame class for a "fire engine"). (This description includes some NLU-specific features).

```
ObjectFrameClass ->

    ObjectFrameClassName
    MassSubstance Boolean flag
    DictionaryPriorWord structure   // (e.g. contains "fire")
    Dictionary structure            // (e.g. contains "engine")
    HigherClasses list              // (e.g. contains "EverydayObjectFrameClass")
    StructuralParentClassesBase     // (e.g. contains "EverydayStructuralParentClass")
    RelationshipToParent structure
```

AttributeTypes list
Attributes list
Templates (used for infusion)
RelationshipTypes list
DimensionSystems list
Structure (list of ObjectFrameClass)
BehaviorClass list

The HigherClasses list represents all higher classes in the optional inheritance hierarchy for an object frame class. For instance, a Car class may get some of its attributes and structure via inheritance from a Vehicle class. The StructuralParentClassesBase item is a list that usually consist of a single item that represents the structural parent class of the object frame class. The RelationshipToParent structure contains attributes that specify how the object frame class is tied to a structural parent class or classes. An example would involve a set of attributes relating an Engine class to a Car class.

The Structure section is where sub-parts, or components of the class are represented. The BehaviorClass list contains references to behavior classes that can be associated with object frame instances that are instantiated from the object frame class.

### 11.12.1.    Basic Form

An example object frame class is shown: this is a class that represents an ignition key for a vehicle (a simple steel ignition key typical of earlier automotive eras). This demonstrates the basic structure of an object frame class that exists mainly to specify the attribute that the class is made of steel.

```
ObjectFrameClass "SteelIgnitionKeyObjectFrameClass"
(
    <StructureTrait val = "Compound"/>

    DictionaryPriorWord
    (
      English
      (
        { " ignition ", "ignition" }
      );
    );
    Dictionary
    (
      English
      (
        { "key", "keys" }
      );
    );

    StructuralParentClassesBase
    (
      { "EverydayObjectStructuralParentClass" }
    );

    RelationshipToParent
    (
```

```
AtLocations  // (location)
(
   AtLocationSet // placeholders:
   (
      <DimensionSystem ref = VehicleComponentCoordinates />
      <Attribute ref = X-Coordinate val = "nil" />
      <Attribute ref = Y-Coordinate val = "nil" />
      <Attribute ref = Z-Coordinate val = "nil" />
   );
);

OuterDimensionSystemExtents  // (size)
(
   OuterDimensionSystemExtentSet // placeholders:
   (
      <DimensionSystem ref = VehicleComponentCoordinates />
      <Attribute ref = X-Coordinate val = "nil" />
      <Attribute ref = Y-Coordinate val = "nil" />
      <Attribute ref = Z-Coordinate val = "nil" />
   );
);

AttributeTypes
(
   AttributeType "MaterialCompositionAttributeType"
   (
      <SuperType val = "QualityAttributeType"/>

      "Values"
      (
         <SuperTypeUsage val = "QualityValues" />

         { "Brass",
           "Steel" }
      );
   );
);

Attributes
(
   //  Every instance of this class has this specific attribute:

   Attribute "MaterialComposition"
   (
      <Attribute ref = MaterialCompositionAttributeType val = "Steel" />
   );
);

// (not needed here)  DimensionSystems ();

// (not needed here)  Structure ();

); // SteelIgnitionKeyObjectFrameClass
```

Each of these sections is described below.

## 11.12.2.        Syntax

The syntax of the object frame class is as follows:

```
ObjectFrameClassStatement  ->  ObjectFrameClassKeyword ObjectFrameClassName
                                      ObjectFrameClassExpression;

ObjectFrameClassKeyword  ->  'ObjectFrameClass' ;

ObjectFrameClassName  ->  '"' identifier '"' ;

ObjectFrameClassExpression  ->

     '('
     [ XMLElementSealedClassBooleanFlag ]
     [ XMLElementProbabilityOfExistenceWithinStructuralParent ]
       XMLElementStructureTraitEnumeratedTypeValue ]
     [ XMLElementMultipleBooleanFlag ]
     [ XMLElementStructuralParentClassBooleanFlag ]
     [ XMLElementMassSubstanceBooleanFlag ]
     [ DictionaryPriorWordExpression ]
     [ DictionaryExpression ]
     [ HigherClassesSection ]
     [ StructuralParentClassesBaseSection ]
     [ RelationshipToParentSection ]
     [ AttributeTypesSection ]
     [ DimensionSystemsSection ]
     [ SpecificationSystemsSection ]
     [ RelationshipTypesSection ]
     [ AttributesSection ]
     [ CompositionSection ]
     [RelationshipsSection ]
     [ StructureSection ]
     [ BehaviorsPotentialSection ]
     [ BehaviorsActualSection ]
     ')' ';' ;

XMLElementSealedClassBooleanFlag  ->  '<' "SealedClass" "val" '=' BooleanStringValue '/' '>' ;

BooleanStringValue  ->  "true"
                           |  "false" ;

XMLElementProbabilityOfExistenceWithinStructuralParent  ->  '<' "ProbabilityInStructuralParent" "expr" '='
                                                floatingpointconstant '/' '>' ;

XMLElementStructureTraitEnumeratedTypeValue  ->  '<' "StructureTrait" "val" '='
                                           StructureTraitEnumeratedTypeStringValue '/' '>' ;

StructureTraitEnumeratedTypeStringValue  ->  "Unit"
                                     |  "Compound"
                                     |  "Range" ;

XMLElementMultipleBooleanFlag  ->  '<' "Multiple" "val" '=' BooleanStringValue '/' '>' ;

XMLElementStructuralParentClassBooleanFlag  ->  '<' "StructuralParentClass" "val" '=' BooleanStringValue '/'">' ;

XMLElementMassSubstanceBooleanFlag  ->  '<' "MassSubstance" "val" '=' BooleanStringValue '/' '>' ;
```

*(See following sections for description and examples of the remaining elements).*

### 11.12.3.　Sealed Class Flag

The SealedClass flag is used by NLU systems that generate Star language code – it serves to indicate that the class is read-only and cannot be modified.

### 11.12.4.　ProbabilityInStructuralParent

The ProbabilityInStructuralParent field allows for the specification of a probability value that indicates the probability that a component class is a member of the parent class in which it appears. This is one of several features that together allow for the definition of fuzzy classes. The following example illustrates this feature. This is the Structure section for a HouseCatObjectFrameClass; this allows for the representation of the fact that "98% of house cats have a front left leg".

```
Structure   // the following are components of a cat body:
(
   ObjectFrameClass "FrontLegLeft"
   (
      <ProbabilityInStructuralParent expr = 0.98 />

      <StructureTrait val = "Compound"/>

      Dictionary ( English
      (
         { "leg",
           "legs" }
      ););

      RelationshipToParent ();
   );

   // (not shown) ObjectFrameClass "FrontLegRight"
   // (not shown) ObjectFrameClass "RearLegLeft"
   // (not shown) ObjectFrameClass "RearLegRight"
);
```

### 11.12.5.　StructureTraitEnumeratedTypeStringValue

This field is an XML element that defines the structure trait for the class. Values are:

- Unit: the object frame class (and instances instantiated from it) consists of only one unit-sized location.
- Compound: the object frame class or instance may contain at least one structural component. (A "compound" class may also be referred to as an "aggregate" class). (Note: "Compound" is often specified for a class that does not yet have structural components, with the expectation that structural components may at some point be added to the class; in this case the class and instances derived from it function like a "Unit" class).
- Range: the object frame class or instance contains multiple contiguous unit-sized locations (1D, 2D, or 3D), but it does not contain named structural components.

### 11.12.6.    Collections (Multiple flag)

The collection is an abstraction that represents a set of object frame classes. When an object instance is instantiate from an object frame class with the Multiple flag set to true, the object instance itself is a collection of object instances. The collection concept addresses the need for an implementation mechanism that corresponds to the universal quantifier of FOL.

The members of a collection are handled differently from individual parts within an object frame class:

- They do not have individual RelationshipToParent information
- Each member must exist within the spatial cuboid region (3D) that is defined by the RelationshipToParent section.
- The features that follow the RelationshipToParent section are features that apply to each set member individually. This is useful for establishing set membership criteria. The sections that can be used to define identifying features include the Attributes section, the Structure section and the Behaviors section. (Examples that would use the Behaviors section include: "the set of all farmers who beat their donkeys", and "the set of all barbers who do not shave themselves" (cf. Russell's Paradox)).

The following is an example of a collection: a set of checkout lanes in a grocery store: in this example, the grocery store is the "whole" and a grocery store has a collection of parts – checkout lanes. Each checkout lane has several features; i.e. a checkout lane is defined as something that has those features, as shown. The identifying features of all checkout lanes include the existence of two structural components: a loading area and a cash register.

```
ObjectFrameClass "GroceryStoreObjectFrameClass"
(
    <StructureTrait val = "Compound"/>

    DictionaryPriorWord
    (
        English
        (
            { " grocery ", "grocery" }
        );
    );
    Dictionary
    (
        English
        (
            { "store", "stores" }
        );
    );

    Structure
    (
        ObjectFrameClass "CheckoutLaneObjectFrameClass"
        (
            < StructureTrait val = "Compound"/>
```

```
            <Multiple val = "true" />
            <Cardinality val = "nil" />  // placeholder

       RelationshipToParent
       (
           // Detail not shown: this establishes a cuboid region within which all
           // set members (i.e. each checkout lane) are located.
       );

       // Attributes: none here, but attributes can also be used to define features for set members

       Structure
       (
           ObjectFrameClass "LoadingAreaObjectFrameClass"
           (
               // (not shown)
           );
           ObjectFrameClass "CashRegisterObjectFrameClass"
           (
               // (not shown)
           );
       );

       // Behavior: none here, but behaviors can also be used to define features for set members

    );  // ObjectFrameClass "CheckoutLaneObjectFrameClass"

  );  // Structure

);  // ObjectFrameClass "GroceryStoreObjectFrameClass"
```

### 11.12.7.      StructuralParentClass Boolean Flag

If true, this flag indicates that the object frame class is a structural parent class. For instance, this flag is set to true within the definition of an upper ontology class called "EverydayObjectStructuralParent".

### 11.12.8.      MassSubstance Boolean Flag

(NLU-specific) If true, this flag indicates that the object frame class holds a "substance" of something: e.g. water, air, gold or silver.

### 11.12.9.      Dictionary and DictionaryPriorWord Sections

These sections allow for the specification of a set of natural language-specific words and phrases that can be used to represent members of this class within natural language text. Although it is optional, an object frame class definition will usually have a Dictionary field (exceptions include structural parent classes and object frame classes that are used abstractly, i.e. as higher-level classes only and not to instantiate object instances). A DictionaryPriorWord field is optional.

### 11.12.10. HigherClasses Section

The HigherClasses list represents all higher classes in the optional inheritance hierarchy for an object frame class. For instance, a Car class may get some of its attributes and structure via inheritance from a Vehicle class. This section is optional because an object frame class may in fact define all of the types that it needs and all of the features (e.g. attributes) that pertain to instances of the class.

### 11.12.11. StructuralParentClassesBase Section

The StructuralParentClassesBase item is a list that usually consist of a single item that represents the structural parent class of the object frame class. A structural parent object frame class is used in order to instantiate a structural parent instance that instances of an object frame class can be attached to. E.g. object instances of the PersonObjectFrameClass can be attached to an object instance of the EverydayObjectStructuralParentClass.

### 11.12.12. RelationshipToParent Section

The RelationshipToParent section contains several groups of attributes that specify how the object frame class is related to a structural parent class or classes. An example would involve a set of attributes relating a PersonBody class to a Person class. The RelationshipToParent section may define a set of placeholder attributes (wherein all actual attribute values are designated "nil"). When object instances are instantiated from this class, if specific location attributes are or become known (e.g. from input natural language text), the attribute values will get filled in. The attribute groups of this section are:

- AtLocations
- OrientationSpecifierSets
- OuterDimensionSystemExtents

### 11.12.13. AttributeTypes Section

This optional section defines a list of attribute type definitions for the class.

### 11.12.14. DimensionSystems Section

This optional section defines dimension systems that are needed by structural components of the object frame class. This section is not needed for all classes: e.g. in the above example for the ignition key class, dimension systems are not needed since the ignition key class does not contain any sub-parts.

### 11.12.15.    SpecificationSystems Section

This optional section defines specification systems that are needed by structural components of the object frame class, specifically for those structural components that themselves use template classes and infusion. For instance a vehicle engine compartment class may have a component called CarburetorObjectFrameClass. The vehicle engine compartment class can define a MaterialCompositionSpecificationSystem that provides locational and qualitative attribute types that are used by template classes that specify specific carburetors.

### 11.12.16.    RelationshipTypes Section

This optional section defines relationship types.

### 11.12.17.    Attributes Section

This optional section specifies specific attributes that pertain to all instances of the class. In the above example for the ignition key class it represents that all ignition key instances are of steel composition.

### 11.12.18.    Composition Section

*(Not included here).*

### 11.12.19.    Relationships Section

*(Not included here).*

### 11.12.20.    Structure Section

The optional Structure section implements "part-to-whole" structure; it is where sub-parts, or components of the class are represented. This section need not be present if the class does not need to define structural parts. (An upcoming example demonstrates the usage of this section).

### 11.12.21.    BehaviorsPotential Section

The optional BehaviorsPotential list contains references to behavior classes that can be associated with object frame instances that are instantiated from the object frame class. During compilation of Star code, when this section is absent and a subsequent behavior class refers to the object frame class, a post-compilation processing step exists that adds the behavior class to the object frame class's internal list of associated behavior classes. This section is useful for defining a probability value for the potential behavior.

```
BehaviorsPotentialSection
(
  BehaviorClassDesignator
  (
    <BehaviorClassName val = "PersonWalksBehaviorClass" />
```

```
    // Note: this represents  the probability that a given person *can* walk:
    //
    <Probability expr = 0.94 />
  );
);
```

## 11.12.22.    BehaviorsActual Section

A BehaviorsActual section designates behaviors that are part of the definition of the class. These behaviors are similar to attributes of the class insofar as they are essential features. The following behavior would be defined for a "FarmerWhoBeatsHisDonkeyObjectFrameClass" in order to specify that this is the class consisting of "every farmer who owns a donkey beats it".

```
BehaviorsActualSection
 (
   BehaviorClassDesignator
   (
       <BehaviorClassName val = "FarmerBeatsDonkeyBehaviorClass" />
       <Probability expr = 1.0 />
   );
);
```

## 11.12.23.    Example

The following example object frame class statement defines a rudimentary person class. This example builds on the concepts that were illustrated by the simple "ignition key" example above by showing additional fields and sections. (This example shows some *structural aspects* of the class; the structural aspects are described in greater detail in the section that follows).

```
ObjectFrameClass "PersonObjectFrameClass"
(
   <SealedClass val = "false" /> // (the default)

   <StructureTrait val = "Compound"/>  // ("Compound" since it has a Structure section, below)

   <StructuralParentClass val = "false"/> // (the default)

   Dictionary
   (
     English
     (
       { "person", "persons", "human", "humans" }
     );
   );

   HigherClasses
   (
     { "EverydayObjectFrameClass",
       "EarthBoundObjectFrameClass" }  // provides orientation specifiers, e.g. "above", "below"
   );

   // (the following is not needed since it is gotten via the inheritance hierarchy)
   // StructuralParentClassesBase
   // (
```

```
//     { "EverydayObjectStructuralParentClass" }
// );

// (when the RelationshipToParent section is absent, instances of the class can be instantiated and
//   attached to a structural parent instance, but the location and size attributes cannot be set)
//
// RelationshipToParent

AttributeTypes
(
    AttributeType "PersonAge"
    (
        <SuperType val = "Qualitative"/>

        "Values"
        (
          { "Infantile" :  Dictionary ( English ( { "infant" } ); ); ,
            "Child" :  Dictionary ( English ( { "young" } ); ); ,
            "Teenager" :  Dictionary ( English ( { "teenage" } ); ); ,
            "Adult" :  Dictionary ( English ( { "adult" } ); ); ,
            "MiddleAgedAdult" :  Dictionary ( English ( { "middle-aged", "adult" } ); ); ,
            "AdvancedAgedAdult" :  Dictionary ( English ( { "elderly", "senior", "older", "old" } ); );
          }
        );
    );
);

// (when this is absent, the components of the person class must use the dimension system(s)
//  of the structural parent classes for the person class).
//
// DimensionSystems

AttributesSection
(
    Attribute "MaterialComposition"
    (
        <Attribute ref = MaterialCompositionAttributeType val = "Organic" />
    );
    Attribute "BodyWeight"
    (
        <Probability expr = 0.90 />
        <Attribute ref = BodyWeightAttributeType range =  { 120 .. 220 } />
    );
);

Structure
(
    ObjectFrameClass "PersonHeadObjectFrameClass"
    (
        <ProbabilityInStructuralParent expr = 0.99999 />
    );
    ObjectFrameClass "PersonLeftArmObjectFrameClass"
    (
        <ProbabilityInStructuralParent expr = 0.989 />
    );
    ObjectFrameClass "PersonRightArmObjectFrameClass"
    (
        <ProbabilityInStructuralParent expr = 0.989 />
    );
```

```
   // (other components here)
);

BehaviorsPotentialSection
(
   BehaviorClassDesignator  // (a person can walk)
   (
      <BehaviorClassName val = "PersonWalksBehaviorClass" />
      <Probability expr = 0.8 />
   );
);

); // "PersonObjectFrameClass"
```

### 11.12.24.        Overview: How an Object Frame Class Implements Structure

There are two ways to represent mereological structure in ROSS. For the first approach, a group of representational constructs enables the representation of structure that involves components (a *whole* with one or multiple *parts*). These constructs include the *structural parent* entity, a set of *relationship to parent* locational attributes and a *structure* section for aggregate entities that models the component-wise ("PartOf") features of the class. The second approach involves the use of the object frame class range, described in the next section.

#### 11.12.24.1.        Part-to-Whole Structure

Part-to-whole physical structure is implemented for the person class as follows. The outline shown here may also be used to define nested structure where needed – e.g. a person body component class may itself have a Structure section that defines components such as Trunk, LeftLeg and RightLeg.

First, the DimensionSystems section defines one or more dimension systems that allow for the specification of the location of the sub-part object frame classes of the person class. This section defines a special "holder" dimension system ("PersonObjectHolder") that allows for the sub-parts of the person class to be enumerated. This is useful for establishing a list of the sub-parts at a level of abstraction that does not include details about size and location. *(Cross-dimension system transforms can be used to derive more detailed specifications where this is needed).*

Second, a person body class is defined in the Structure section of the person class. Since its relative place within the person class is not specified, the person body class can be said to "float" somewhere within the perimeters of the person class.

Third, the RelationshipToParent structure of the person body class specifies either specific location attributes of this component in relation to the parent person class, or it can be used to declare placeholders that are used for such specification in fact repository artifacts (instance models) when such information is available. These location attributes make use of the dimension system of the parent (the person class), in order to specify both spatial place and spatial size ("extent").

Note that the RelationshipToParent section for PersonBodyObjectFrameClass contains a number of attributes with "nil" for the attribute value: these attributes are placeholders. An instance that is instantiated from this class will fill in specific attribute values here as available (e.g. if the input NL text specifies something to represent the value).

```
ObjectFrameClass "PersonObjectFrameClass"
(
    <StructureTrait val = "Compound"/>

    Dictionary
    (
        English
        (
            { "person", "persons", "human", "humans" }
        );
    );

    HigherClasses
    (
        { "EverydayObjectFrameClass",
          "EarthBoundObjectFrameClass" }  // provides orientation specifiers, e.g. "above", "below"
    );

    DimensionSystems
    (
        DimensionSystem "PersonObjectHolder"
        (
            <RoleTrait val = "Holder"/>

            LocationAttributeTypes
            (
                SpatialAttributeTypes
                (
                    "RelativePlace"
                    (
                        <SuperType val = "LocationAttributeType"/>

                        "GeneralLocationValueSet"
                        (
                            <SuperTypeUsage val = "LocationValues" />

                            { "PersonHeadSlot",  // a compartment or receptacle (i.e. a cuboid region) that can
                                                 // be correlated with coordinates of other dimension systems,
                                                 // e.g. the coordinate-based dimension system, below.
                              "PersonBodySlot"
                            }
                        );
                    );
                );
            );
        );

        // (note: this statement declares a local name that is defined to refer to a more-general
        // dimension system called "PhysicalObjectMillimeterCoordinates" that is part of
        // a basic definitions Star code file).
        //
        DimensionSystem "PersonPhysicalCoordinates" (PhysicalObjectMillimeterCoordinates);
    );
```

```
Structure
(
   // other person sub-parts here, e.g. "head", "neck"

   ObjectFrameClass "PersonBodyObjectFrameClass"     // a part/component
   (
      Dictionary ( English
      (
         {
           "body",
           "bodies"
         }
      ););

      // HigherClasses  (not needed here)

      RelationshipToParent
      (
         AtLocations
         (
            AtLocationSet
            (
               <DimensionSystem ref = PersonObjectHolder />
               <Attribute ref = RelativePlace val = "PersonBodySlot" />
            );
            AtLocationSet // placeholders:
            (
               <DimensionSystem ref = PersonPhysicalCoordinates />
               <Attribute ref = X-Coordinate val = "nil" />
               <Attribute ref = Y-Coordinate val = "nil" />
               <Attribute ref = Z-Coordinate val = "nil" />
            );
         );

         // OrientationSpecifiers // (not shown)

         OuterDimensionSystemExtents
         (
            OuterDimensionSystemExtentSet
            (
               <DimensionSystem ref = PersonPhysicalCoordinates />
               <Attribute ref = X-Coordinate val = "nil" />
               <Attribute ref = Y-Coordinate val = "nil" />
               <Attribute ref = Z-Coordinate val = "nil" />
            );
         );
      );

      // DimensionSystems  (not needed here)

      // Structure  // (this would be used for nested structure within person body)

   ); // "PersonBodyObjectFrameClass"

); // Structure

); // "PersonObjectFrameClass"
```

### 11.12.24.2. Structure Using the Object Frame Class Range

The object frame class range is a special type of aggregate object frame class (composed of multiple spatially adjacent unit-sized location entities that span one, two or three dimensions). This construct does not have in internal structure that is composed of structural components; rather, it is like a 3D drawing canvas on which a picture can be drawn. A simple example would involve a cubical object frame class range in which can be drawn a sphere. The representational construct that is used for rendering (specifying qualitative values of each unit-sized location within the range) is called a *template* class.

### 11.12.25. Class Hierarchy

The *higher class* construct allows for the specification of one or more higher classes. Class inheritance is viewed solely as a way of aggregating or consolidating groups of attributes and structural features – it is only a *convenience mechanism*. Higher classes (parent classes) supply additional representational information about a given class. This bottom-up approach to inheritance hierarchies distinguishes ROSS from other ontological approaches. ROSS does not enforce the use of a single "root object" class, (although an Infopedia may inadvertently exhibit this property if all classes point upward to some class that *happens to be* a root class). ROSS allows for multiple inheritance (multiple higher classes per class).

ROSS also allows for the existence of multiple classes that may be inadvertently similar, based on the view that some classes, such as a "person class", may be better handled by the use of multiple classes in order to model a variety of feature collections. Different domains would use different such classes. An example might involve several classes such as *PersonAsCountryCitizenClass*, *PersonAsBiologicalLivingEntityClass*, and *PersonAsTravellerClass*. Each of these would be useful in different domains.

The following are examples of derived classes.

```
ObjectFrameClass "CarObjectFrameClass"
(
    Dictionary ( English
    (
        {
          "car",
          "cars",
          "auto",
          "autos",
          "automobile",
          "automobiles"
        }
    ););
    HigherClasses ( { "VehicleObjectFrameClass" } );
);

ObjectFrameClass "TruckObjectFrameClass"
(
    Dictionary ( English
```

```
(
    {
      "truck",
      "trucks",
      "pickup",
      "pickups"
      }
    ););
    HigherClasses ( { "VehicleObjectFrameClass" } );
);
```

The derived classes ("car" and "truck") automatically get all of the structural, attributive and behavioral information of the parent, or higher class ("vehicle").

## 11.13.  Template Class Statement

*(This section is draft and for review-only).*

A template object class describes a range of locations within an object frame using the RelationshipToParent information, and it refers to either a 3D bitmap, or to a set of "drawing" instructions that specify how to render the compositional properties of the object or component.

The template class can be understood using the metaphor of drawing: a template class describes a method that is used to draw a picture within an object frame range instance.  A simple example would be a template class that contains a function to draw an oval within an object frame range instance that has a rectangular shape. A more complex example would involve a set of drawing instructions that can be used for drawing a face, or for the 3D rendering of a person's head within a cuboid-shaped object frame range instance. The process of drawing/rendering is referred to as "infusion". A template class must contain an attribute value expression that specifies either a drawing routine or a 3D bitmap.

### 11.13.1.        Basic Form

The example here illustrates the basic form: this would be used to infuse the values of all unit-sized object locations within an object frame instance for an animal's head (e.g. a house cat).

```
TemplateObjectClass "AnimalHead_Template001"
(
    <StructuralParentClass ref = AnimalObjectFrameClass />
    <ObjectFrameClass ref = AnimalObjectFrameClass.HeadObjectFrameClass />
    <ShapeTemplate val = "false" />

    // TwoPartAttributeCluster:
    <SpecificationSystem ref = AnimalComponentPhysicalComposition />
    <Attribute ref = X-Coordinate var = x$ />
    <Attribute ref = Y-Coordinate var = y$ />
    <Attribute ref = Z-Coordinate var = z$ />
    <Attribute ref = EssentialValueAttributeType routine = "RenderAnimalHead" />
    //<Attribute ref = EssentialValueAttributeType bitmap = "AnimalHead3D" />

    OuterDimensionSystemExtentSet
    (
        <DimensionSystem ref = AnimalComponentMillimeterCoordinates />
```

```
            <Attribute ref = X-Coordinate val = "700" />
            <Attribute ref = Y-Coordinate val = "700" />
            <Attribute ref = Z-Coordinate val = "700" />
      );
   );
```

## 11.14.  Populated Object Class Statement

The populated object class is a representational construct that allows for the application of a set of qualitative attribute values to an aggregate object frame instance. The process of applying a populated object class to an object frame instance is referred to as "population". Populated object classes are primarily used within behavior classes, described in the following section. They may also be defined standalone (at global scope in a Star code file). Fields that are not described here include the "Negation" field and the BinderSourceFlag. (Several of the examples here make use of a behavior class in the following section called "FarmerBeatsDonkeyBehaviorClass").

The populated object class expression (within the populated object class statement) consists of several fields, followed by a two-part attribute cluster expression.

### 11.14.1.      Basic Form: Standalone Definition

This populated object class can be used to set values for a house cat object instance – it describes a brown house cat that is in the "sitting" state. The StructuralParentClass XML element is required for standalone populated object class definitions; for populated object class definitions within a behavior class it can be derived from a field called BridgeObjectFrameClass. (Several comments are included that help explain the sections).

```
PopulatedObjectClass "HouseCatBrownSitting"
(
    <StructuralParentClass ref = EverydayObjectStructuralParentClass />
    <ObjectFrameClass ref = HouseCatObjectFrameClass />

    // TwoPartAttributeCluster:
     // DimensionSetExpression:
     <DimensionSystem ref = PhysicalObjectMillimeterCoordinates />
     <Attribute ref = X-Coordinate var = x$ />
     <Attribute ref = Y-Coordinate var = y$ />
     <Attribute ref = Z-Coordinate var = z$ />
     // value attributes:
     <Attribute ref = ExteriorColor val = "Brown" />
     <Attribute ref = StandingState val = "Sitting" />
);
```

### 11.14.2.      Basic Form as Used in a Behavior Class

The following example is of a populated object class that represents an instance of a farmer class in the state of "not beating" (an animal). (Refer to the behavior class section for a description of the BinderSourceFlag field).

```
    PopulatedObjectClass "AntecedentActor"
    (
```

```
          <ObjectFrameClass ref = FarmerObjectFrameClass />
          <BinderSourceFlag val = "true" />
          <DimensionSystem ref = RelativePosition />
          <Attribute ref = RelativeLocation var = a$ />
          <Attribute ref = RelativeTime var = t1$ />
          <Attribute ref = BeatingState val = "NotBeating" />
      );
```

### 11.14.3.　　Name

The name of a populated object class (e.g. "AntecedentActor", above), is only descriptive and is not used by a behavior class that contains the populated object class.

### 11.14.4.　　ObjectFrameClass Field

This XML element field contains a reference to the object frame class for which the populated object class is defined.

### 11.14.5.　　Participant Designation Fields

There are two optional fields that can be used to designate that the populated object is in the passive role or the "extra" role. The default role is *actor*. An object in the passive role is illustrated here:

```
      PopulatedObjectClass "AntecedentActee"
      (
          <ObjectFrameClass ref = DonkeyObjectFrameClass />
          <PassiveParticipant val = "true" />
          <DimensionSystem ref = RelativePosition />
          <Attribute ref = RelativeLocation expr = (a$+1) />
          <Attribute ref = RelativeTime expr = t1$ />
          <Attribute ref = PassiveIsBeatenState val = "NotBeaten" />
      );
```

### 11.14.6.　　Use of UniqueIdentityAttributeType Fields

Within behavior classes, a populated object class may need to define a variable that is used by attributes within nested behaviors in order to identify a particular instance within the rule. This is needed when a rule part (e.g. the antecedent) contains more than one object of the same type – e.g. a person class object in the actor role and a person class object in the passive (actee) role. The populated object class employs the unique identity attribute type for such a purpose. (Note that the variable name, "q$", is just an arbitrarily-chosen name).

```
      PopulatedObjectClass "AntecedentActor"
      (
          <ObjectFrameClass ref = FarmerObjectFrameClass />
          <BinderSourceFlag val = "true" />
          <DimensionSystem ref = RelativePosition />
          <Attribute ref = RelativeLocation var = a$ />
          <Attribute ref = RelativeTime var = t1$ />
          <Attribute ref = BeatingState val = "NotBeating" />
```

```
<Attribute ref = UniqueIdentityAttributeType var = q$ />  // IDENTITY
);
```

### 11.14.7.    Probability Field

The optional probability field can be used by populated object classes that exist within behavior classes. For instance, when used in the consequent of a rule, it indicates the probability that the state is true, given the antecedent states. This example represents the probability that an animal (that has been beaten) is dead (the "IsKilled" state). (i.e. the probability is 50% that is dead).

```
PopulatedObjectClass "ConsequentActee"
(
    <ObjectFrameClass ref = AnimalObjectFrameClass />
    <Probability expr = 0.5 />
    <PassiveParticipant val = "true" />
    <DimensionSystem ref = RelativePosition />
    <Attribute ref = RelativeLocation expr = (a$+1) />
    <Attribute ref = RelativeTime expr = (t1$+1) />
    <Attribute ref = PassiveIsKilledState val = "IsKilled" />
);
```

### 11.14.8.    Collections (Multiple flag)

The following populated object class represents a collection. This definition is from a behavior class called "TalkerAdvocatesActionWithListenersWhoAnticipateSomething" that is shown in full in the appendix. The Multiple flag indicates that the populated object class represents a collection: in this case it represents a set of "listeners" – i.e. all persons who *hear* that some one or group of persons advocates violence.

```
PopulatedObjectClass "ConsequentExtra"  // Listener(s)
(
    <ObjectFrameClass ref = PersonObjectFrameClass />
    <Multiple val = "true" /> // Collection
    <ExtraParticipant val = "true" />
    <DimensionSystem ref = RelativePosition />
    <Attribute ref = RelativeLocation expr = (a$+1) />
    <Attribute ref = RelativeTime expr = t1$ />
    <Attribute ref = CommunicationReceivedState val = "CommunicationReceived" />
);
```

## 11.15.  Behavior Class Statement

### 11.15.1.    Overview

The behavior class is the basis for describing processes: at its simplest a behavior class represents a sequence that involves at least two subsequent states. (A behavior class can even represent only states that exist at the same time). Besides their use in representing simple processes, behavior classes can also represent events, actions, causal processes, and processes that involve a correlation between multiple states, which may or may not be causative. A behavior class associates a set of "prior" states with a set of "post" states. Examples of behavior classes for the PersonObjectFrameClass class include "PersonHitsPerson", "PersonWalks" and "PersonCommunicates". Behavior classes have the following structure:

- A *bridge structural parent class* – a reference to an object frame class that contains a dimension system that must be shared by all object frame classes in the behavior class, so that locational relationships can be specified within the binder construct that ties objects of the prior states section to objects of the post states section.
- A *PriorStates section*, consisting of a list of populated object classes. This is like the antecedent (the "if part") within a rule.
- A *PostStates section*, consisting of a list of populated object classes. This is like the consequent (the "then part") within a rule.

### 11.15.2.  Basic Form

The basic form of the behavior class is illustrated using a rudimentary behavior class for a person hitting another person.

```
//------------------------------------------------------------------------
//
// BehaviorClass: PersonHitsPerson
//
//       E.g.: "The man hit the woman."
//
//       Before:
//          Man-1 hits
//          Woman-1 not yet hit (by Man-1).
//       After:
//          Woman-1 has been hit.
//
//------------------------------------------------------------------------
//
BehaviorClass "PersonHitsPerson"
(
    <BridgeObjectFrameClass ref = BehavioralStructuralParentClass />

    Dictionary
    (
      English
      (
        { "hit",    // (infinitive/base)
          "hit",    // (simple past)
          "hit",    // (past participle)
          "hits",   // (simple present, 3rd p.s.)
          "hitting", // (present participle)

          "punch",
          "punched",
          "punched",
          "punches",
          "punching"
        }
```

```
                    );
                );

            PriorStates
            (
                PopulatedObjectClass "AntecedentActor"   // (name is descriptive only)
                (
                    <ObjectFrameClass ref = PersonObjectFrameClass />
                    <BinderSourceFlag val = "true" />
                    <DimensionSystem ref = RelativePosition />
                    <Attribute ref = RelativeLocation var = x$ />
                    <Attribute ref = RelativeTime var = t$ />
                    <Attribute ref = HittingState val = "Hitting" />
                );
                PopulatedObjectClass "AntecedentActee"
                (
                    <ObjectFrameClass ref = PersonObjectFrameClass />
                    <PassiveParticipant val = "true" />
                    <DimensionSystem ref = RelativePosition />
                    <Attribute ref = RelativeLocation expr = (x$+1) />
                    <Attribute ref = RelativeTime expr = t1$ />
                    <Attribute ref = PassiveHitState val = "NotHit" />
                    <Attribute ref = UniqueIdentityAttributeType var = q$ />  // Identity
                );
            );
            PostStates
            (
                PopulatedObjectClass "ConsequentActee"
                (
                    <ObjectFrameClass ref = PersonObjectFrameClass />
                    <PassiveParticipant val = "true" />
                    <DimensionSystem ref = RelativePosition />
                    <Attribute ref = RelativeLocation expr = (x$+1) />
                    <Attribute ref = RelativeTime expr = (t$+1) />
                    <Attribute ref = PassiveHitState val = "Hit" />
                    <Attribute ref = UniqueIdentityAttributeType expr = q$ />  // Identity
                );
            );
        ); // BehaviorClass "PersonHitsPerson"
```

This behavior class uses a mechanism that involves a qualitative attribute that is based on a UniqueIdentityAttributeType: this allows for differentiation between the person who hits and the person who is hit.

### 11.15.3.        Syntax

The syntax of the behavior class is as follows:

```
BehaviorClassStatement  ->  BehaviorClassKeyword BehaviorClassName BehaviorClassExpression;

BehaviorClassKeyword  -> 'BehaviorClass' ;

BehaviorClassName ->  '"' identifier '"' ;

BehaviorClassExpression ->

    '('
```

```
[ XMLElementSealedClassBooleanFlag ]
[ XMLElementCausalRuleBooleanFlag ]
[ XMLElementRuleDirectionEnumeratedTypeStringValue ]
 XMLElementBridgeObjectFrameClassReference
[ XMLElementNegationBooleanFlag ]
[ DictionaryPriorWordExpression ]
[ DictionaryExpression ]
[ ModificationSection ]
[ HigherClassesSection ]
PriorStatesRulePartSection
PostStatesRulePartSection
')' ';' ;

XMLElementSealedClassBooleanFlag  -> '<' "SealedClass" "val" '=' BooleanStringValue '/' '>' ;

XMLElementCausalRuleBooleanFlag  -> '<' "CausalRule" "val" '=' BooleanStringValue '/' '>' ;

BooleanStringValue  -> "true"
                 | "false" ;

XMLElementRuleDirectionEnumeratedTypeStringValue  -> "Unspecified"
                                          | "Forward"
                                          | "Backward" ;

XMLElementBridgeObjectFrameClassReference  -> '<' "BridgeObjectFrameClass" "ref" '='
                                          ObjectFrameClassName '/' '>' ;

ObjectFrameClassName  -> identifier ;

XMLElementNegationBooleanFlag  -> '<' "Negation" "val" '=' BooleanStringValue '/' '>' ;

DictionaryPriorWordExpression  -> // (see Dictionary Expression)

DictionaryExpression  -> // (see Dictionary Expression)
```

*(See following sections for description and examples of the remaining elements).*

### 11.15.4.      Behavior Class Name

The behavior class name is not required to be unique; lookup routines that use ROSS behavior classes will seek a match for one of the verbs defined in the Dictionary section and will attempt to match the object frame classes that are referred to within the populated object classes.

### 11.15.5.      Sealed Class Flag

The SealedClass flag is used by NLU systems that generate Star language code – it serves to indicate that the class is read-only and cannot be modified.

### 11.15.6.       Causal Rule Boolean Flag

The optional CausalRule flag, if true, indicates that this behavior should be treated as a causal or correlative rule. Where this flag is true, the states of the objects in the PriorStates section are treated as rule antecedents and those in the PostStates section are treated as rule consequents.

Optional causal feature qualitative attributes serve a unique role within behavior classes that are rules. During the application of a rule, as takes place during behavior resolution or during inference, a "major" structural parent instance is cloned to create a "minor" structural parent instance; likewise all component object instances are cloned. However the state attributes that are specified within populated objects in the rule are not applied to the cloned copies. Furthermore, optional causal feature attributes are not cloned: the purpose of an optional causal feature in a consequent section is the representation of a state that is caused by the conjunction of states in the antecedent; similarly, the purpose of an optional causal feature in an antecedent section is the representation of a state that has a causal relationship with the conjunction of states in the consequent.

### 11.15.7.       RuleDirection

The RuleDirection field designates whether a rule behavior class is intended for use as a forward rule or a backward rule. (A third value for RuleDirection is "unspecified", for non-directional rules). This field is used by inference routines that must distinguish the direction of causality (forward-directed or backward-directed).

### 11.15.8.       Probabilities

The behavior class does not have a probability field since probabilities can be specified per rule element within populated object classes and within nested rule references.

### 11.15.9.       BridgeObjectFrameClassReference and Binder Mechanism

The BridgeObjectFrameClass field refers to a structural parent object frame class that is used as the default structural parent for all participant objects. It provides a dimension system that allows all constituent objects to be related to one another. The objects are related to each other using the *binder* infrastructure. One of the constituent objects must define a BinderSourceFlag – this object then declares variables for each locational attribute. The other constituent objects of the behavior class contain locational attributes that contain expressions that specify spatial or temporal location in relation to the binder source object. (Note that a variable may also be declared within an attribute expression in a populated object class that is not the binder source, as is the case with attributes that use the UniqueIdentityAttributeType).

The binder mechanism is used during instance model creation and inference: it allows for the determination and setting of locational values for objects that get instantiated and/or positioned during application of the rule.

### 11.15.10.  Negation Boolean Flag

The Negation boolean flag is used in order to indicates whether or not the states that are specified in the consequent of the rule are to be negated.

### 11.15.11.  Dictionary Sections

A behavior class definition is required to have a Dictionary field. A DictionaryPriorWord field is optional.

### 11.15.12.  Modification Section

The modification section allows for the definition of additional modification words or phrases that are a necessary part of the lexical properties of the class. An example is as follows (from a PersonTriesToKillAnimalBehaviorClass).

```
Modification
(
  DictionaryModifyingVerbs
  (
    English
    (
      {
        "try"   // infinitive
      }
    );
  );
  DictionaryAdverbs
  (
    English
    (
      {
        "quickly"
      }
    );
  );
);
```

### 11.15.13.  HigherClasses Section

*(The HigherClasses section is not described in this version of the document).*

### 11.15.14.  RulePartMain Structure: PriorStates and PostStates

A behavior class contains two main parts, a PriorStates section, and a PostStates section. When the behavior class is used as a rule, the PriorStates section functions as an antecedent section and the PostStates section plays the role of a consequent section. The PriorStates section and the PostStates section are each an instance of a type called a *RulePartMain*.

### 11.15.14.1.    PriorStates and PostStates

By default and if not overridden by time-related attributes in the binder, the PriorStates section contains object frame classes that exist at an earlier time than those that are contained in the PostStates section. Since ROSS does not actually contain objects that exist for more than a single time instant, it is important to realize that an object, e.g. a person that is represented by a PersonObjectFrameClass in the PriorStates section is actually not the same object as the person represented by a PersonObjectFrameClass in the PostStates section. Nevertheless some occasions necessitate that a particular object that exists in the PriorStates section be designated such that an equivalent object in the PostStates section can be identified with it. This is accomplished using an attribute type called "UniqueIdentityAttributeType" (an upcoming example demonstrates this feature).

### 11.15.14.2.    Elements of a RulePartMain

A RulePartMain section contain elements that are either populated object classes or nested behavior references. A populated object class is an abstraction that allows for the specification of a propositional truth about an object. A populated object class is associated with a single object frame class, and it specifies a set of attributes about the object frame class. The set of attributes is a two-part attribute cluster. (The two-part attribute cluster fully describes the location and at least one aspect of the qualitative state of an object).

It should be noted that the object frame classes that are the basis for each populated object class within a rule part do not overlap with one another (in the sense that they each have a unique identity, as there may be spatial overlap). This is of particular relevance if one or more populated object classes represents a collection. (The inference routine that is described in this document has a dependency on this non-overlapping aspect).

Nested behavior references are specified using the "BehaviorClassReference" keyword. The following "farmer beats donkey" behavior class example demonstrates the use of a nested behavior reference.

### 11.15.14.3.    Nested Behavior Class Reference

A nested behavior class reference is a definitional construct that refers to another behavior class which is causally related to the behavior class that contains the reference. The BehaviorClassReference construct allows for the specification of parameters for each of "actor", "actee" (passive role) and "extra" participants. The classes referred to within these parameters must exist within the main behavior class in which the nested behavior reference appears. A nested behavior reference can contain a probability field; if one exists it serves the same purpose as does the probability field within a populated object class within a rule.

The following nested behavior reference is contained in a behavior class called "FarmerBeatsDonkeyBehaviorClass" (the full example is shown below).

BehaviorClassReference
    (

```
<BehaviorClass ref = ActiveOwnershipBehaviorClass />  // -->> DEFINED-BEHAVIOR-CLASS
<ParameterActor ref = FarmerObjectFrameClass expr = q$ />
<ParameterActee ref = DonkeyObjectObjectFrameClass />
);
```

## 11.15.15.          Example: FarmerBeatsDonkeyBehaviorClass

The following behavior class represents a behavior that is derived from a well-known logic
example: "If a farmer owns a donkey then he beats it." However, since the ROSS behavior class is
a representation of *capability,* the following behavior class is actually a representation of this
sentence:

"Every farmer who owns a donkey is capable of beating it."

The behavior class depends on several object frame classes: these are described first.

### 11.15.15.1.          Preliminary: Object Frame Classes

The following ROSS object frame classes are needed to support the main behavior class that
follows. (Note: in a typical ROSS ontology, the PersonObjectFrameClass and
AnimalObjectFrameClass, referred to here as higher classes, would already exist).

```
ObjectFrameClass "FarmerObjectFrameClass"
(
   <StructureTrait val = "Compound"/>

   Dictionary ( English
   (
      { "farmer",     // singular
        "farmers" }   // plural
   ););

   HigherClasses ( { "PersonObjectFrameClass" } );

   AttributeTypes
   (
      AttributeType "BeatingState"
      (
         <SuperType val = "Qualitative"/>

         <StateAttributeType val = "true" />

         "Values"
         (
            {
              "NotBeating",
              "Beating"
            }
         );
      );
   );
);
```

```
ObjectFrameClass "DonkeyObjectFrameClass"
(
    <StructureTrait val = "Compound"/>

    Dictionary ( English
    (
        { "donkey",
          "donkeys" }
    ););

    HigherClasses ( { "AnimalObjectFrameClass" } );

    AttributeTypes
    (
        AttributeType "PassiveIsBeatenState"
        (
            <SuperType val = "Qualitative"/>

            <StateAttributeType val = "true" />

            "Values"
            (
                {
                  "NotBeaten",
                  "Beaten"
                }
            );
        );
    );
);
```

### 11.15.15.2.        Behavior Class

The FarmerBeatsDonkey behavior class is shown below. The nested behavior reference in this example represents the fact that the farmer owns the donkey. It refers to another behavior class (not shown here) called "ActiveOwnershipBehaviorClass". The BehaviorClassReference construct is capable of associating its object references (actor and actee) to a populated object within the same rule: in this case the actor (the farmer class) refers to the populated object that involves a farmer class shown earlier in the rule. The association is accomplished using the UniqueIdentityAttributeType.

```
BehaviorClass "FarmerBeatsDonkeyBehaviorClass"
(
    <CausalRule val = "true" />

    <BridgeObjectFrameClass ref = BehavioralStructuralParentClass />

    Dictionary ( English
    (
        {
          "beat",
          "beat",
          "beaten",
          "beats",
          "beating"
```

```
        }
    ););

    PriorStates
    (
        PopulatedObjectClass "AntecedentActor"
        (
            <ObjectFrameClass ref = FarmerObjectFrameClass />
            <BinderSourceFlag val = "true" />
            <DimensionSystem ref = RelativePosition />
            <Attribute ref = RelativeLocation var = a$ />
            <Attribute ref = RelativeTime var = t1$ />
            <Attribute ref = BeatingState val = "NotBeating" />
            <Attribute ref = UniqueIdentityAttributeType var = q$ />
        );
        PopulatedObjectClass "AntecedentActee"
        (
            <ObjectFrameClass ref = DonkeyObjectFrameClass />
            <PassiveParticipant val = "true" />
            <DimensionSystem ref = RelativePosition />
            <Attribute ref = RelativeLocation expr = (a$+1) />
            <Attribute ref = RelativeTime expr = t1$ />
            <Attribute ref = PassiveIsBeatenState val = "NotBeaten" />
        );
        BehaviorClassReference
        (
            <BehaviorClass ref = ActiveOwnershipBehaviorClass />  // DEFINED-BEHAVIOR-CLASS -->>
            <ParameterActor ref = FarmerObjectFrameClass expr = q$ />
            <ParameterActee ref = DonkeyObjectObjectFrameClass />
        );
    );
    PostStates
    (
        PopulatedObjectClass "ConsequentActor"
        (
            <ObjectFrameClass ref = FarmerObjectFrameClass />
            <DimensionSystem ref = RelativePosition />
            <Attribute ref = RelativeLocation expr = (a$+1) />
            <Attribute ref = RelativeTime expr = (t1$+1) />
            <Attribute ref = BeatingState val = "Beating" />
        );
        PopulatedObjectClass "ConsequentActee"
        (
            <ObjectFrameClass ref = DonkeyObjectFrameClass />
            <PassiveParticipant val = "true" />
            <DimensionSystem ref = RelativePosition />
            <Attribute ref = RelativeLocation expr = (a$+1) />
            <Attribute ref = RelativeTime expr = (t1$+1) />
            <Attribute ref = PassiveIsBeatenState val = "Beaten" />
        );
    );
);  // FarmerBeatsDonkeyBehaviorClass
```

## 11.15.16.     Example: FarmerTriesToKillAnimalBehaviorClass

The next example is a somewhat similar behavior class (but without a nested behavior), showing the use of a Modification structure. A modification structure contains two dictionary

sections – the first (DictionaryModifyingVerbs) allows the behavior to be modified by a valid form of the verb or verbs (usually only one is used). The second section (DictionaryAdverbs) contains adverbs that further qualify the behavior. (The example shown – "quickly" is not used here and is commented out). This behavior class represents the action that would involve a person (possibly a farmer) *trying* to kill an animal, e.g. a donkey. This example also shows the use of higher classes for constituent populated object classes – the PersonObjectFrameClass is used instead of the FarmerObjectFrameClass and the AnimalObjectFrameClass is used rather than DonkeyObjectFrameClass. This allows all objects that inherit from person and animal to potentially participate in the behavior. (The attribute types "AttemptingToKillState", for the person class and "PassiveIsKilledState" for the animal class are defined within the person and animal classes respectively and are not shown here).

Since this behavior class represents "trying to kill", note that the result state for the animal (within the PostStates section) specifies an attribute that indicates that the animal is "NotIsKilled".

```
BehaviorClass "PersonTriesToKillAnimalBehaviorClass"
(
    <BridgeObjectFrameClass ref = BehavioralStructuralParentClass />

    Dictionary ( English
    (
        {
          "kill", "kill", "killed", "kills", "killing"
        }
    ););

    Modification
    (
        DictionaryModifyingVerbs
        (
          English
          (
              {
                "try"   // infinitive
              }
          );
        );
        //DictionaryAdverbs
        //(
        //English
        //   (
        //      {
        //         "quickly"
        //      }
        //   );
        //);
    );

    PriorStates
    (
        PopulatedObjectClass "AntecedentActor"
        (
            <ObjectFrameClass ref = FarmerObjectFrameClass />
            <BinderSourceFlag val = "true" />
```

```
                    <DimensionSystem ref = RelativePosition />
                    <Attribute ref = RelativeLocation var = a$ />
                    <Attribute ref = RelativeTime var = t1$ />
                    <Attribute ref = AttemptingToKillState val = "AttemptingToKill" />
                );
                PopulatedObjectClass "AntecedentActee"
                (
                    <ObjectFrameClass ref = AnimalObjectFrameClass />
                    <PassiveParticipant val = "true" />
                    <DimensionSystem ref = RelativePosition />
                    <Attribute ref = RelativeLocation expr = (a$+1) />
                    <Attribute ref = RelativeTime expr = t1$ />
                    <Attribute ref = PassiveIsKilledState val = "NotIsKilled" />
                );
            );

            PostStates
            (
                PopulatedObjectClass "ConsequentActee"
                (
                    <ObjectFrameClass ref = AnimalObjectFrameClass />
                    <PassiveParticipant val = "true" />
                    <DimensionSystem ref = RelativePosition />
                    <Attribute ref = RelativeLocation expr = (a$+1) />
                    <Attribute ref = RelativeTime expr = (t1$+1) />
                    <Attribute ref = PassiveIsKilledState val = "NotIsKilled" />
                );
            );
        ); // PersonTriesToKillAnimalBehaviorClass
```

### 11.15.17.       Use of Behavior Class to Represent a Nominal Process

The behavior class as thus described is associated with verbs; however a behavior class can also be associated with nouns that represent processes, actions or events. (Examples of verb-based behavior classes include "PersonWalksBehaviorClass" and "PersonCommunicatesBehaviorClass"). Examples of noun-based behavior classes include "StormBehaviorClass", "EarthquakeBehaviorClass", and "CheckingAccountWithdrawalBehaviorClass". This feature is useful for NLU systems that need the level of detail about objects and states that the behavior class provides. For instance, an NL sentence might state that "Last Tuesday's earthquake caused extensive damage throughout the city." The existence of an earthquake behavior class with a dictionary that contains "earthquake" and "earthquakes" allows the NLU system to generate a sequence of states within an instance model that are a representation of the event.

### 11.15.18.       How Behavior Classes are Associated With Object Frame Classes

Object frame classes and behavior classes are related to each other in either of two ways: in the first case, an object frame class may specify an associated behavior class using the BehaviorsPotential section. If this section is absent, the Star compiler will associate the object frame class with the behavior via a post-compilation processing step.

## 12. Infopedia: Ontology and Knowledge Base Concepts

This section deals with concepts that pertain to ROSS ontologies/knowledge bases.

### 12.1.  Overview

A ROSS Infopedia contains Star language definitions (and optionally, special behavior classes that are rule-like). An Infopedia includes a mixture of definitions that cross the spectrum from universal and generic to domain-specific. The upper ontology definitions of an Infopedia include generic object frame classes for high-level abstract objects, e.g. ObjectObjectFrameClass and EverydayObjectFrameClass. The upper ontology also includes a variety of supporting definitions for attribute value sets, attribute types, value set mappings, and dimension system types. Middle ontology classes include those from which lower-level classes can derive features: for instance, a *container* object frame class and an *enclosable object* class. The lower ontology has object frame classes such as PersonObjectFrameClass and VehicleObjectFrameClass, and behavior classes such as PersonHitsPerson.

An Infopedia is extensible: definitions can be hand-code using the Star language (they are stored in a set of user-editable text files) or they can be derived as part of a knowledge acquisition process that interprets user-entered input natural language text and generates Star language definitions.

### 12.2.  Higher-Level Generic Infopedia Definitions

The Infopedia that is used in the Comprehendor NLU system contains a collection of files that have definitions that primarily exist to support commonsense reasoning and NLU tasks such as anaphora resolution. These files include:

- BasicDefinitions.h: attribute value sets, attribute types, mappings, dimension systems
- EarthboundObjectDefinitions.h: supporting definitions and two object frame classes: the first represents the ground ("earth", or "planet surface") from the perspective of a human observer, the second represents an "earthbound" object. These classes are useful in providing features that are needed for situations that implicitly involve the ground as a frame of reference.
- EverydayObjectDefinitions.h: supporting definitions and two main classes: an EverydayObjectStructuralParent, which provides a structural parent class that is used by most objects in the commonsense representation and reasoning area, and EverydayObjectFrameClass, which provides a higher class from which many objects can inherit features: this is mainly the feature of having the EverydayObjectStructuralParent as a structural parent.
- InformationDefinitions.h and IntelligentAgentClasses.h: classes that relate to the task of the representation of information, cognition (memory, data, and processes), and communication on the part of an intelligent agent.

### 12.3.  Internal Representation of Infopedia in an NLU System

Within the Comprehendor NLU system, all definitions are contained in C++ maps, which use a balanced binary tree as the underlying storage to support efficient key-based searching. The Comprehendor in-memory Infopedia maps include the following:

- MapDeclarationIntegers
- MapDeclarationFloatingPoints
- MapDeclarationStrings
- MapValueSets
- MapMappings
- MapAttributeTypes
- MapRelationshipTypes
- MapDimensionSystems
- MapSpecificationSystems
- MapObjectFrameClasses
- MapTemplateClasses
- MapPopulatedObjectClasses
- MapBehaviorClasses

The definitions are used by a set of Infopedia query routines to support both simple lookups and more-complex queries, such as those that are needed during NLU processing by the Comprehendor semantic engine.

### 12.4.  Flexibility of the Infopedia Concept

ROSS Infopedias are interchangeable; this provides considerable flexibility for the modeling of classes and world knowledge. Domain-specific Infopedias can be created (alternately, domain-specific information can be added to a general-purpose Infopedia). For instance the NLU domain of information about consumer automobiles (e.g. including articles about cars in an auto enthusiast's magazine), would benefit from having a more detailed set of classes about cars and the structure of cars than would be needed for the general-use commonsense domain that has vehicle information as part of the transportation information category.

### 12.5.  Ontology Derivation/ Knowledge Acquisition

Automated learning of classes is an important area that uses the features of ROSS. The use of learning techniques is not an absolute necessity for ontology and knowledge acquisition, since both generic and domain-specific ROSS definitions can be created by a human knowledge engineer or ontology practitioner. (Upper ontology ROSS definitions are better suited to hand-crafted creation – these include general-use attribute value sets, attribute types and dimension systems). Nevertheless knowledge engineering has long been recognized as a bottleneck for AI; automated approaches can facilitate and greatly ease the arduous and time-consuming task of knowledge acquisition. The following are several broad categories of automated knowledge acquisition that involve learning of classes and class features from natural language text:

- Intermediate-depth approaches that learn features based on associations. E.g. (unsupervised) learning that cars can be blue based on sentences that associate "blue" with "car".
- Learning new sub-classes and their behaviors based on controlled natural language input: i.e. simple sentences of the form "an x is a y that does z". (E.g. "an electrician is a person who fixes electrical problems").
- Deeper approaches that learn structure, features and behaviors from NL descriptions that explicitly describe structure and features. (E.g. learning of classes and behaviors from a simple encyclopedia entry on the automobile).

The ontology derivation task is not limited to natural language-based approaches. Other possibilities include the use of interactive tools that such as those that would allow human users to draw objects. Another approach would involve the processing of engineered specifications to generate ROSS classes.

## 12.6.  Ontology and Knowledge Acquisition from Controlled English Input

The Comprehendor NLU system has capabilities for generating class definitions from natural language sentences that are input by the user. The following Star language definition was auto-generated by Comprehendor for the sentence: "A student is a person.". (Note that a PersonObjectFrameClass already existed when this class was generated).

```
ObjectFrameClass "StudentObjectFrameClass"
(
    Dictionary ( English
    (
        { "student",
          "students" }
    ););
    HigherClasses ( { "PersonObjectFrameClass" } );
);
```

The following classes are generated from the sentence "An electrician is a person who fixes electrical problems.". (Several minor post-generation edits have been applied). Note that these classes refer to several upper ontology classes that already existed in the Infopedia at the time of generation.

```
ObjectFrameClass "ElectricianObjectFrameClass"
(
    <StructureTrait val = "Compound"/>

    Dictionary ( English
    (
        { "electrician",
```

```
          "electricians" }
    ););

    HigherClasses ( { "PersonObjectFrameClass" } );

    AttributeTypes
    (
       AttributeType "FixingState"
       (
          <SuperType val = "Qualitative"/>
          <StateAttributeType val = "true" />

          "Values"
          (
             {
               "NotFixing",
               "Fixing"
             }
          );
       );
    );
); // ObjectFrameClass "ElectricianObjectFrameClass"

BehaviorClass "FixesBehaviorClass"
(
    <BridgeObjectFrameClass ref = BehavioralStructuralParentClass />

    Dictionary ( English
    (
       { "fix", "fixed", "fixed", "fixes", "fixing" }
    ););

    PriorStates
    (
       PopulatedObjectClass "AntecedentActor"
       (
          <ObjectFrameClass ref = ElectricianObjectFrameClass />
          <BinderSourceFlag val = "true" />
          <DimensionSystem ref = RelativePosition />
          <Attribute ref = RelativeLocation var = a$ />
          <Attribute ref = RelativeTime var = t1$ />
          <Attribute ref = FixingState val = "NotFixing" />
       );
       PopulatedObjectClass "AntecedentActee"
       (
          <ObjectFrameClass ref = ElectricalProblemObjectFrameClass />
          <PassiveParticipant val = "true" />
          <DimensionSystem ref = RelativePosition />
          <Attribute ref = RelativeLocation expr = (a$+1) />
          <Attribute ref = RelativeTime expr = t1$ />
          <Attribute ref = PassiveIsFixedState val = "NotFixed" />
       );
    );
    PostStates
    (
       PopulatedObjectClass "ConsequentActor"
       (
          <ObjectFrameClass ref = ElectricianObjectFrameClass />
```

```
            <DimensionSystem ref = RelativePosition />
            <Attribute ref = RelativeLocation expr = a$ />
            <Attribute ref = RelativeTime expr = (t1$+1) />
            <Attribute ref = FixingState val = "Fixing" />
        );

    PopulatedObjectClass "ConsequentActee"
    (
        <ObjectFrameClass ref = ElectricalProblemObjectFrameClass />
        <PassiveParticipant val = "true" />
        <DimensionSystem ref = RelativePosition />
        <Attribute ref = RelativeLocation expr = (a$+1) />
        <Attribute ref = RelativeTime expr = (t1$+1) />
        <Attribute ref = PassiveIsFixedState val = "Fixed" />
    );
); // BehaviorClass "FixesBehaviorClass"
```

## 12.6.1. Auto-Generated Classes for the "Trophy and Suitcase" Winograd Schema

The following shows a few of the several classes that were auto-generated in support of the method that was developed by the author in order to handle the Winograd schema challenge schema #2 ("trophy and suitcase"). The input text for each is shown first. (Not shown: "ContainerObjectObjectFrameClass", "SuitcaseObjectFrameClass", and a second "NOT_FitBehaviorClass" that contains the functional attribute type for "too small").

The first class shown here is the object frame class that was generated from this input text: "An enclosable object is an everyday object.". (The EverydayObjectFrameClass is part of the upper ontology and is not shown here).

```
ObjectFrameClass "EnclosableObjectObjectFrameClass"
(
    <StructureTrait val = "Compound"/>

    DictionaryPriorWord
    (
        English
        (
            { "enclosable",
              "enclosable" }
        );
    );
    Dictionary ( English
    (
        { "object",
          "objects" }
    ););
    HigherClasses ( { "EverydayObjectFrameClass" } );
);
```

The trophy class was then generated from the input text "A trophy is an enclosable object."

```
ObjectFrameClass "TrophyObjectFrameClass"
(
    <StructureTrait val = "Compound"/>
```

```
Dictionary ( English
(
    { "trophy",
      "trophies" }
););

HigherClasses ( { "EnclosableObjectObjectFrameClass" } );
);
```

The following section shows code that was generated from the input text: "If an enclosable object is too big then it does not fit in a container object.". This includes: 1) the additional class information that was added to the EnclosableObjectFrameClass, and 2) the generated behavior class. (Several comments were manually added afterwards).

```
ObjectFrameClass "EnclosableObjectObjectFrameClass"
(
    AttributeTypes
    (
        AttributeType "FittingState"
        (
            <SuperType val = "Qualitative"/>
            <StateAttributeType val = "true" />

            "Values"
            (
                {
                  "NotFitting",  // e.g. not starting motion to fit into something
                  "Fitting"      // e.g. in motion to fit into something
                }
            );
        );

        AttributeType "FunctionalAttributeType1"
        (
            <SuperType val = "Qualitative"/>
            <StateAttributeType val = "true" />
            <OptionalCausalFeature val = "true" />

            "Values"
            (
                {
                  "NotTooBig",
                  "TooBig" : Dictionary ( English ( { "big" } ); );
                }
            );
        );
    );
);  // ObjectFrameClass "EnclosableObjectObjectFrameClass"

BehaviorClass "NOT_FitBehaviorClass"
(
    <CausalRule val = "true" />
    <BridgeObjectFrameClass ref = BehavioralStructuralParentClass />
    <Negation val = "true" />

    Dictionary ( English
```

```
(
    {
      "fit",
      "fit",
      "fitted",
      "fits",
      "fitting"
    }
););

PriorStates
(
    PopulatedObjectClass "AntecedentActor"
    (
        <ObjectFrameClass ref = EnclosableObjectObjectFrameClass />
        <BinderSourceFlag val = "true" />
        <DimensionSystem ref = RelativePosition />
        <Attribute ref = RelativeLocation var = a$ />
        <Attribute ref = RelativeTime var = t1$ />
        <Attribute ref = FittingState val = "NotFitting" />
        <Attribute ref = FunctionalAttributeType1 val = "TooBig" />
    );
    PopulatedObjectClass "AntecedentActee"
    (
        <ObjectFrameClass ref = ContainerObjectObjectFrameClass />
        <PassiveParticipant val = "true" />
        <DimensionSystem ref = RelativePosition />
        <Attribute ref = RelativeLocation expr = (a$+1) />
        <Attribute ref = RelativeTime expr = (t1$+0) />
        <Attribute ref = PassiveIsFittedState val = "NotFitted" />
    );
);

PostStates
(
    PopulatedObjectClass "ConsequentActor"
    (
        <ObjectFrameClass ref = EnclosableObjectObjectFrameClass />
        <DimensionSystem ref = RelativePosition />
        <Attribute ref = RelativeLocation expr = (a$+1) />
        <Attribute ref = RelativeTime expr = (t1$+1) />
        <Attribute ref = FittingState val = "Fitting" />
    );
    PopulatedObjectClass "ConsequentActee"
    (
        <ObjectFrameClass ref = ContainerObjectObjectFrameClass />
        <PassiveParticipant val = "true" />
        <DimensionSystem ref = RelativePosition />
        <Attribute ref = RelativeLocation expr = (a$+1) />
        <Attribute ref = RelativeTime expr = (t1$+1) />
        <Attribute ref = PassiveIsFittedState val = "Fitted" />
    );
);
); // BehaviorClass "NOT_FitBehaviorClass"
```

## 13. Fact Repository Concepts

This section deals with concepts that pertain to ROSS fact repositories and with the processes such as instantiation that generate the information that exists in fact repositories.

### 13.1.  Fact Repositories: Transcripts and Instance Models

There are a variety of representational artifacts that use the ROSS approach with representational constructs that are fact-like. The term "fact-like" includes representations that are facts about past situations, and it includes other assertions such as plan goals and predictions. "Fact repository" is defined to include any representational artifact containing such constructs. A fact repository has a top-level structure: the repository may represent multiple situations, e.g. situations that are a mixture of ones from the past (from various places and times), others that are present-tense, and some that are hypothetical.

#### 13.1.1. Transcripts

A transcript is a document that contains fact-like representational constructs for use in AI automated reasoning applications. There are a number of transcript types that use ROSS. These include the following:

- Past fact transcripts that are useful for automated reasoning about past fact situations (e.g. fault diagnosis).
- Specification transcripts for automated inference for design or planning problems; these transcripts contain fact-like constructs that include predicted states and goals.

#### 13.1.2. Instance Models

An *instance model* is a type of fact repository for NLU systems: it is a meaning representation instance that represents factual information about past and/or present situations and events. It is an artifact that is a structured representation of the subject matter of an input natural language text fragment such as a story.

### 13.2.  Situations and Object Instances

This section describes the things that are represented by fact repositories.

#### 13.2.1. The Situation

A fact repository may contain one or multiple *situations*. A *situation* is a collection of related facts, each of which involves entities that all share a common structural parent instance or set of consecutive structural parent instances, e.g. along a time-line.

### 13.2.2. Structural Parent Instance

A structural parent instance is a special type of object frame instance that has a unique role in a ROSS fact repository. A structural parent instance is a top-level object frame instance. A ROSS situation contains a *set* of structural parent instances, each of which has been instantiated from a common structural parent class, and each of which serves a special function as a "structural parent" for object instances that can be attached to it. A structural parent instance exists at a point along a timeline – it is like a snapshot or a single frame from a movie. The structural parent object frame instance has an *InstanceStructure* section that specifies all object frame instances that are immediate children that are within the spatial and temporal range of the structural parent object frame instance.

The analogy of the diorama is a useful one for describing a structural parent instance. The object instances that get attached to a structural parent instance (e.g. a person instance, e.g. a car instance) are like figures in a diorama. (There is one important difference – object instances are, strictly speaking, empty rectangular or cuboid regions which *can hold* the figures in question).

### 13.2.3. Object Frame Instance

A ROSS object frame instance, or "object instance" is a concept that may be implemented as an in-memory data object, or as an information record (e.g. in an external XML instance model). An object instance is an instantiated instance of an object frame class.

The structure of an object instance is shown here.

```
ObjectInstance ->

    ObjectFrameClassName
    ObjectInstanceUniqueIdentifier
    CausalityRole
    RelationshipToParent structure
    Attributes list
    Relationships list
    InstanceStructure (structure containing list of object instances)
```

The first field of an object instance is the object frame class from which it was instantiated. The next field is a unique identifier that refers to the instance as it exists or existed in the space-time frame of reference of the structural parent of the context.

The next field, CausalityRole, designates whether the object instance is part of a cause or part of an effect. If the object instance is the structural child of a parent object instance, the RelationshipToParent structure can specify the specific attributes that relate the child to the parent. The Attributes and Relationships lists contain attribute and relationship attribute information about the object instance. Finally, the InstanceStructure is a collection of references to all child instances. For instance, the representation of a "car" instance would typically contain object instances in this section for "engine", "transmission", "body frame", etc.

### 13.3.    The Instantiation Process

Instantiation is the process of creating an object frame instance within a fact repository from an object frame class; it involves the sub-tasks of *attachment* and of *infusion* or *population*. (For purposes of illustration, each of these concepts is described here in terms how it is performed by a NLU semantic engine, e.g. when the engine generates an object instance within an instance model).

### 13.3.1. Attachment

*Attachment* is the process of creating an object frame instance. When a structural parent object frame instance is created within a situation, it is simply given a unique identifier or name. However when an object frame instance that is a child of a structural parent, or of other object frame instances is created, attachment involves creating the instance, giving it an identifier or name, and possibly setting its RelationshipToParent attributes.  It also involves specifying a reference to the child instance within the InstanceStructure section of the parent instance.

The effect of performing a group of attachments can be visualized as analogous to a process of creating a diorama frame and then inserting various empty smaller cuboid-shaped wire-frame structures (some nested within others) into it.

### 13.3.2. Overlapping Object Frame Instances

Object frame instances within a structural parent instance may overlap with one another. Since an object frame instance is an empty container-like abstraction, this does not create problems; however, instantiation and inference processes must perform calculations to determine if a candidate infusion attempt  is possible or if it would cause a collision with an existing infused value at a shared location. (E.g. two successive NL sentences illustrate this: "There is a large piano on the green mat.", and  "There is a cat on the same mat.". An NLU semantic engine must address the question of whether or not the two objects collide with each other).

### 13.3.3. Anchor Points

Anchor points are needed so that the "At locations" and "Spatial orientation" attributes of a new component can be properly set when it is attached within an object frame instance that is its parent. *(Details about anchor points are not described in this version)*.

### 13.3.4. Infusion and Population

The process of *infusion* operates on empty object frame instances: it sets actual values for them. Infusion as applied to a unit-sized object frame instance just involves setting its value. Infusion of a value into an object frame range instance makes use of a template class. *Population* is similar to infusion and involves using a populated object class to set the values of an aggregate object frame instance.

### 13.3.5. Global Assumptions

Practical considerations involving the creation and maintenance of representations in a fact repository artifact necessitate the use of several convenience assumptions. These assumptions can be specified as being "on" or "off" within the global scope of a fact repository, e.g. a transcript. A set of global assumptions in an instance model would look like this:

```
<GlobalAssumptions>
    <!-- Any location that has not been infused has a value that
        inherits from the "SpaceValue" value category -->
    <EmptySpaceAssumption value ="true" />
    <!-- Attached objects are permanent through time -->
    <PermanentAttachmentsAssumption value = "true" />
    <!-- Stationary values at t = n perpetuate forward in time -->
    <PerpetuationAssumption value ="true" />
</GlobalAssumptions>
```

### 13.3.5.1.   EmptySpace Assumption

The empty space assumption is as follows: within a structural parent, at the first time point and for all subsequent time points, any unit-sized location that has not been overtly infused or populated is assumed to have a value that inherits from the SpaceValue value category.

### 13.3.5.2.   Permanence Assumption

The permanence assumption involves attachments: it allows for the attachment (i.e. a declaration) of an object frame instance at time point *t*, and it includes the assumption that subsequent time points contain the same object frame instance at the same location. This avoids the need for the specification of the *detachment* of each and every object frame instance within a time point and the subsequent attachment of the object frame instances within subsequent time points.

### 13.3.5.3.   Perpetuation Assumption

The perpetuation assumption involves perpetuation of values along a time line; it can be used in similar fashion as the empty space assumption: the assumption is that for any unit-sized location that has been infused with a value at time t=n, it can be assumed that the subsequent unit-sized location at the same spatial location (at time t=(n+1)) will have the same value unless it is overtly specified to have a different value. This assumption is useful for stationary objects but does not address the representation of objects in motion.

## 14. Fact Repository for NLU: The ROSS Instance Model

The ROSS instance model has been developed specifically for natural language understanding and thus contains features that facilitate a variety of NLU processing tasks.

At the top level, an NLU instance model contains a list of *contexts*. The order of contexts in the list usually corresponds to the order of occurrence of sentences in the input text, however this is not a requirement.

### 14.1.   Epistemology

A ROSS instance model may use either of two methods in order to handle the epistemological aspects of natural language text. Although it adds a layer of informational and computational overhead, there are several advantages that are gained by the second approach.

- Without using a *meta* representation of the communicative agent or of the communicated information. With this approach, all natural language text contained in the input document is treated as factual (if it is declarative), or handled using a direct approach (e.g. questions). The author (referred to as the "communicative agent") is treated as autonomous, i.e. declarative information is deemed to be true.
- Using a *meta* representation of the communicative agent and the communicated information. *(This approach is not described in detail in this version)*.

### 14.2.   How an Instance Model Implements Behaviors

Object instances implement states of behaviors at points along a timeline. Within an instance model, each single-time-point object frame instance *participates in* behaviors via attributes that specify its state. An object frame instance (at a single time point) can thus participate in multiple behaviors simultaneously due to its having multiple attributes, each of which represent some aspect of its state.

### 14.3.   Definition of "Context"

An instance model *context* corresponds to a single *situation*: it is a representational construct that pertains to a particular space and time frame of reference. A context contains multiple time-sequential states of a situation. Each situation state is represented by a single structural parent instance.

An instance of natural language discourse may have many such contexts. For instance, a story may contain the following two sentences in sequence: "A Seattle home was burglarized late yesterday. John Smith owns the home". The first sentence is in the past tense and is the basis for a context. The second sentence is in the present tense and thus provides the basis for a second and separate context. An instance model contains at least one context.

(When used in order to represent meta-information about communicative agent(s) and communicated information, the context concept may also be used in order to represent the context of spoken or written communication. In this case it designates a separate frame of reference that represents the information that was communicated by a human agent and the information itself).

## 14.4.    Context and Context List

The following listing contains the Context C++ data structure and the ContextList structure. The important map that contains all top-level structural parent instances is in bold:

```
struct Context
{
    char szUniqueIdentifier [MAX_SIZE_UNIQUEID_STRING];
    DiscourseContext discourseContext;
    char szLeadingObjectInstanceClassName [MAXLEN_CONTENTSTRING_STAR];
    char szTemporalAttributeValueLastUsed [ATTRIBUTE_VALUE_MAX_SIZE];

    //-----------------------------------------------------------------------------------------------
    // Map that contains all structural parent instances, indexed by time attributes:
    //
    MapObjectInstances *pMapObjectInstances;

    // Methods not shown:
};

struct ContextListNode
{
    Context *pContext;

    struct ContextListNode *prev;
    struct ContextListNode *next;

    // Methods not shown:
};

class ContextList
{
private:
    ContextListNode *m_head;
    ContextListNode *m_tail;

public:
    // Public methods not shown:
};
```

The MapObjectInstances structure stores structural parent object instances, each of which is indexed by a temporal attribute value.

```
// MapObjectInstances:
//
// - the wrapper class is not shown; the map of object instances contains ObjectInstance pointers:
//
typedef map <string, ObjectInstance*> MapTypeObjectInstances;
typedef pair<MapTypeObjectInstances::iterator,bool> retvalMapTypeObjectInstances;
```

## 14.5. Object Instance

The ObjectInstance C++ class is shown here:

```
class ObjectInstance
{
private:
    ObjectFrameClass *m_pReferenceObjectFrameClass;  // (ptr to class from which it was instantiated)

public:
    bool fInstanceIsPartOfClassStructure;

    // (Special) Parser Information:
    char szContentString [MAXLEN_CONTENTSTRING_STAR];

    //-------------------------------------------------------

    char szUniqueIdentifier [MAXLEN_UNIQUEID_STRING];

    bool fMultiple;

    //-------------------------------------------------------
    // Upon instantiation, each of the following derives any
    // available detail from the object frame class:
    //
    RelationshipToParent relationshipToParent;

    // (from ObjectFrameClass::Structure structure)
    InstanceStructure structure;

    // (from ObjectFrameClass::Attributes attributes)
    AttributeBaseExpression *rgpAttributeExpressions [MAX_OBJECTFRAMEINSTANCE_ATTRIBUTES];

    // (note that the object instance can only have one
    //  applied template at any given time point)
    Composition composition;

    RelationshipExpression *rgpRelationshipExpressions [MAX_OBJECTFRAMEINSTANCE_RELATIONSHIPS];

    //-------------------------------------------------------
    // List of associated behaviors: (this uses a list of class pointers)
    //
    BehaviorClassListNode *pBehaviorClassListNodeHead;

    // Methods not shown
};
```

The InstanceStructure member contains the embedded objects: this is important insofar as the structural parent object instance is only a "holder". For instance, a structural parent instance based on the EverydayObjectStructuralParentClass may contain an object instance for a HouseClass and a DrivewayClass.

## 14.6. Collections of Objects

An object instance has a "Multiple" flag (fMultiple): if this is true it indicates that the object instance region holds a collection of objects of the type.

### 14.7.   Attachment and Infusion

An object instance must be attached to a structural parent; examples include an instance of a person head that is attached to a person instance, and an instance of a person instance that is attached to an instance of the EverydayObjectStructuralParent class.

An object instance is not required to be infused with a value or set of values (as is the case where a template is used to fill in the object). This is referred to as "transparent mode" and is useful for applications that only need to query or set qualitative state attributes. (The object instances in the example of the following diagram are in transparent mode).

### 14.8.   An Example Actual Situation for Which an Instance Model Can be Created

**Figure 3** represents a process, or situation that occurs in the past that involved a person hitting another person. This shows action along a timeline: person A hit person B.

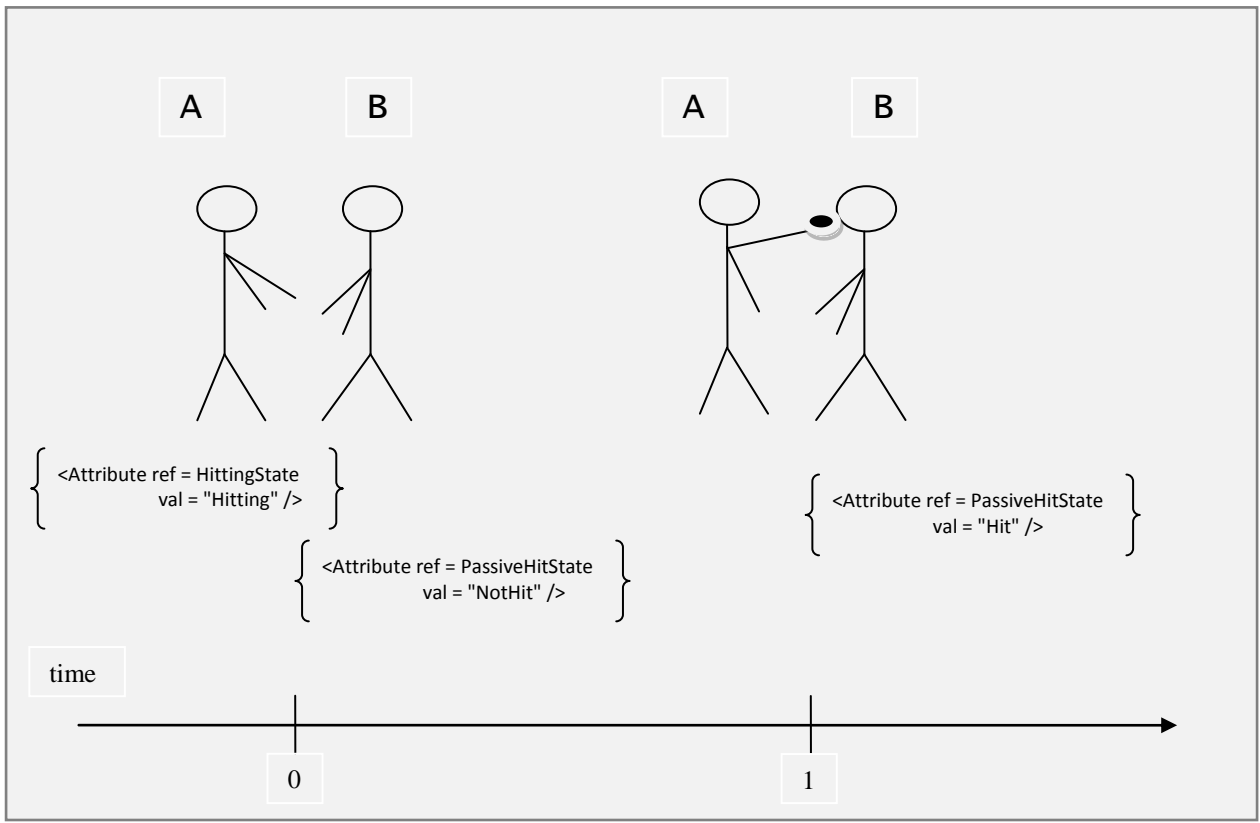

*Figure 3: Visualization of Instance Model for "A person hit another person"*

The process is fully described by the attributes that are shown. Note that no motion is represented as having occurred between time point 0 and time point 1.

## 14.10.  External XML-Based Instance Models

**Figure 4** shows the basic structure an external XML-based instance model.

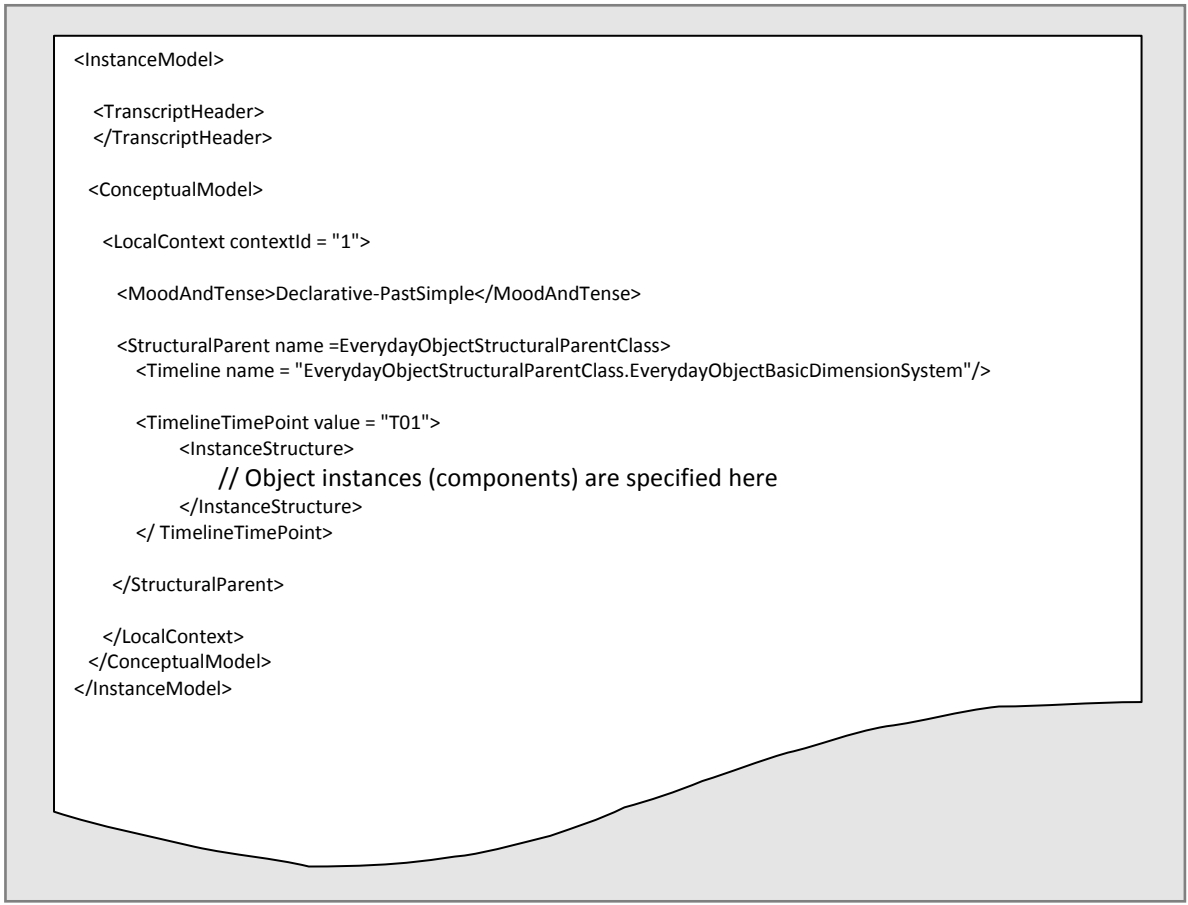

```
<InstanceModel>

  <TranscriptHeader>
  </TranscriptHeader>

  <ConceptualModel>

    <LocalContext contextId = "1">

      <MoodAndTense>Declarative-PastSimple</MoodAndTense>

      <StructuralParent name =EverydayObjectStructuralParentClass>
        <Timeline name = "EverydayObjectStructuralParentClass.EverydayObjectBasicDimensionSystem"/>

        <TimelineTimePoint value = "T01">
            <InstanceStructure>
                // Object instances (components) are specified here
            </InstanceStructure>
        </ TimelineTimePoint>

      </StructuralParent>

    </LocalContext>
  </ConceptualModel>
</InstanceModel>
```

*Figure 4: External NLU Instance Model*

This diagram demonstrates the following:

- At the top level, an instance model contains a TranscriptHeader section and a ConceptualModel section.
- The ConceptualModel section contains a list of contexts.
- A context has a single structural parent class (in this example it is "EverydayObjectStructuralParentClass").
- Each point on the timeline is associated with an instance of the structural parent class.

The structural parent is an object frame instance and thus it has an "InstanceStructure" section; this contains all components, such as person instances

### 14.10.1. XML Instance Model Example: "A Person Hit another Person"

Here is the full external XML-based instance model for the sentence "A young man hit the boy." The instance model also shows the use of the global assumptions.

```
<?xml version="1.0" encoding="US-ASCII" standalone="yes"?>

<InstanceModel>

  <TranscriptHeader>
    <TextSource value="SubmittedFromWebClient">
    </TextSource>
  </TranscriptHeader>

  <ConceptualModel>

    <GlobalAssumptions>
      <EmptySpaceAssumption value ="true" />
      <PermanentAttachmentsAssumption value = "true" />
      <PerpetuationAssumption value ="true" />
    </GlobalAssumptions>

    <LocalContext contextId = "1">

      <MoodAndTense>
        Declarative-PastSimple
      </MoodAndTense>

      <StructuralParent name = EverydayObjectStructuralParentClass >
        <Timeline name = "EverydayObjectStructuralParentClass.EverydayObjectDimensionSystem"/>

      <TimelineTimePoint value = 0 >
        <InstanceStructure>
          <Component>
            ManObjectFrameClass.ManObjectFrameInstance-1 (man) (young)
            <Attributes>
              <Attribute>
                PersonAge = YoungPerson
              </Attribute>
              <Attribute>
                PersonObjectFrameClass.HittingState = Hitting
              </Attribute>
            </Attributes>
          </Component>
          <Component>
            BoyObjectFrameClass.BoyObjectFrameInstance-1 (boy)
            <Attributes>
              <Attribute>
                PersonObjectFrameClass.PassiveHitState = NotHit
              </Attribute>
            </Attributes>
          </Component>
        </InstanceStructure>
      </TimelineTimePoint>

      <TimelineTimePoint value = 1 >
        <InstanceStructure>
          <Component>
            ManObjectFrameClass.ManObjectFrameInstance-1 (man) (young)
```

```
              <Attributes>
                <Attribute>
                  PersonAge = YoungChild
                </Attribute>
              </Attributes>
            </Component>
            <Component>
              BoyObjectFrameClass.BoyObjectFrameInstance-1 (boy)
              <Attributes>
                <Attribute>
                  PersonObjectFrameClass.PassiveHitState = Hit
                </Attribute>
              </Attributes>
            </Component>
          </InstanceStructure>
        </TimelineTimePoint>

      </StructuralParent>

    </LocalContext>

  </ConceptualModel>

</InstanceModel>
```

This example shows one of the features that implements structure: the *InstanceStructure* element. This element contains a list of *Components*, which are representational constructs that represent object frame instances. However it should be noted that in this example, the InstanceStructure mechanism is used solely to provide a structural parent (EverydayObjectStructuralParent) for the components (ManObjectFrameInstance-1 and BoyObjectFrameInstance-1).

### 14.10.2.        XML Instance Model Example: "The farmer beat the donkey"

The following is an external instance model for the sentence: "The farmer beat the donkey." It is a representation of a simple fact that occurred as part of some past situation. Note that ROSS classes have a considerable degree of flexibility and that the attribute states shown here (which are based on the above ROSS ontology classes) are only one way of representing what happens during an event such as this one. For instance, this example shows that at t = 2 ("T02") the farmer is (actively) beating the donkey – alternately the situation could be modeled to show that the beating took place, occurred during a time interval, and then stopped.

```
<?xml version="1.0" encoding="US-ASCII" standalone="yes"?>

<InstanceModel>

  <TranscriptHeader>
    <TextSource value="DocumentFile">
    </TextSource>
    <DocumentFile name="Samples\SimpleSentence.txt">
    </DocumentFile>
  </TranscriptHeader>
```

```
<ConceptualModel>

  <GlobalAssumptions>
    <EmptySpaceAssumption value ="true" />
    <PermanentAttachmentsAssumption value = "true" />
    <PerpetuationAssumption value ="true" />
  </GlobalAssumptions>

  <LocalContext contextId = "1">

    <MoodAndTense>
      Declarative-PastSimple
    </MoodAndTense>

    <StructuralParent name = EverydayObjectStructuralParentClass >
      <Timeline name = "EverydayObjectStructuralParentClass.EverydayObjectDimensionSystem"/>

      <TimelineTimePoint value = "T01">
        <InstanceStructure>
          <Component>
            FarmerObjectFrameClass.FarmerObjectFrameClass-1 (farmer)
            <Attributes>
              <Attribute>
                FarmerObjectFrameClass.BeatingState = NotBeating
              </Attribute>
            </Attributes>
          </Component>
          <Component>
            DonkeyObjectFrameClass.DonkeyObjectFrameClass-1 (donkey)
            <Attributes>
              <Attribute>
                DonkeyObjectFrameClass.PassiveIsBeatenState = NotBeaten
              </Attribute>
            </Attributes>
          </Component>
        </InstanceStructure>
      </TimelineTimePoint>

      <TimelineTimePoint value = "T02">
        <InstanceStructure>
          <Component>
            FarmerObjectFrameClass.FarmerObjectFrameClass-1 (farmer)
            <Attributes>
              <Attribute>
                FarmerObjectFrameClass.BeatingState = Beating
              </Attribute>
            </Attributes>
          </Component>
          <Component>
            DonkeyObjectFrameClass.DonkeyObjectFrameClass-1 (donkey)
            <Attributes>
              <Attribute>
                DonkeyObjectFrameClass.PassiveIsBeatenState = Beaten
              </Attribute>
            </Attributes>
          </Component>
        </InstanceStructure>
      </TimelineTimePoint>
```

```
            </StructuralParent>
          </LocalContext>
        </ConceptualModel>
    </InstanceModel>
```

### 14.11. Use of Instance Models

The following are just a few of the possible uses of the information contained in an instance model:

- Query and question answering against instance model fact data
- In-depth summarization
- Topic analysis/modeling
- Entity extraction
- Relationship extraction

## 15. Introduction to Inference Using ROSS

### 15.1. Overview

This section deals with concepts that pertain to inference using the ROSS behavior class. Behavior classes are used in order to implement correlative rules. A correlative rule is a representational construct that represents correlation in a problem domain. Correlation may or may not involve causality (i.e. the laws for some domain). There is not a limit on the types of correlative rules that can be constructed using the ROSS KR scheme as a foundation - this is due to the view that inference (reasoning) is a multifaceted set of tasks that should not be overly constrained by predefined approaches. Rules are not a part of ROSS fact repository artifacts: since rules are handled separately from facts and other fact-like constructs, a variety of rule base approaches are possible.

### 15.2. Types of Inference Not Covered Here

The following topics are not addressed in this document.

- Inference based on definitional axioms (including set-theory axioms) (e.g. transitivity of PartOf: "PartOf($\alpha,\beta$) $\Lambda$ PartOf($\beta,\gamma$) $\rightarrow$ PartOf($\alpha,\gamma$)")
- Inference based on dimension system axioms/postulates (cf. geometrical axioms, e.g. the Pythagorean theorem)

### 15.3.    ROSS Inference Versus Logic Approaches

Many logic-based formal methods within the AI knowledge representation reasoning field involve a model of entailment that involves a single knowledge base ("KB") that contains a mixture of facts, or fact-like representational constructs and rules. (A first order logic "rule" is defined as an expression that involves implication and the "for all x" quantifier). Inference with ROSS is handled using a different approach: ROSS inference relies on two main inputs: behavior classes from the ontology/knowledge base and known facts (or "seed facts") for a situation. The main output of ROSS inference is one or more new derived/inferred facts.

### 15.4.    The Binder - Relating Antecedent to Consequent in a Correlative Rule

In the field of logic, connexive logic (e.g. relevance logic) addresses the need for correlating information in the antecedent of a rule with information in the consequent. ROSS formalizes the concept of associating antecedent with consequent using a representational construct that is a part of the ROSS behavior class, referred to as the binder. A binder is an abstraction that is implemented in such a way that the locational attributes of entities in the antecedent of a rule are related to the locational attributes of entities in the consequent of the same rule.

### 15.5.    Applications of Inference

Automated reasoning with ROSS is open-ended and unrestricted due to the loose coupling between representation and reasoning. The following are a few examples of broad categories of reasoning tasks that can be accomplished using the ROSS method:

- Fact Determination: reasoning about situations and events that occurred in the past to perform past fact derivation. This category includes diagnosis of faults/failures.
- Prediction: reasoning about a future or hypothetical situation to derive result facts from various possible conditions.
- Design: reasoning from requirements specifications to generate design artifacts.
- Planning: reasoning from plan goals to generate plans.

### 15.6.    Forward-Directed Inference Involving Sandbox Context and Rule Application

An example of inference from the category of fact determination is described here. In particular, a forward-directed inference process is described. This example is taken from the anaphora resolution routine's embedded commonsense reasoning functional area; the example is focused on Winograd schema #1 ("councilmen and demonstrators").

This is a partial description of the embedded inference process that exists within a generate-and-test process that determines the validity of a candidate referent for an anaphor (pronoun). *(Note: the term "forward-directed" as used here has a unique meaning that is an indication of the* *semantics of the time sequence involved; this usage shares similarities with but is not identical to the "forward chaining" of logic-based inference).*

### 15.6.1. Calling Routine

First, an example that shows part of a calling routine is shown below ("GenerateAndTest_ProcessOneForwardRule ()"). The inputs to this function include:

- A pointer to an object instance candidate (e.g. an object instance representing the "councilmen", or an object instance representing the "demonstrators"). The object instance data structure contains a pointer to the object frame class from which it is derived so that object frame class information, such as structural parent class can be obtained.
- A behavior class that has been retrieved by a prior search process that provided one or more object frame classes and a verb-based expression. An example such behavior class is called "TalkerAdvocatesActionWithListenersWhoAnticipateSomething" – this behavior class was retrieved based on the verb "advocates" along with other criteria.
- A pronoun feature set data structure; this includes information about the other syntactic and semantic entities of the clause or phase wherein the pronoun is contained. E.g. for "because they advocated violence", it includes "violence" as a syntactic direct object and as an object that fills the actee semantic role within that clause.

This routine creates a temporary working memory context called the "sandbox context". The output of this routine as shown below is the sandbox context as it has been added to by the insertion of a major structural parent instance, a minor structural parent instance, and object instances within the structural parent instances. The object instances have had their state attributes set with values that will later get matched against attribute values of other object instances from another sandbox context in order to determine if the candidate is the correct antecedent for the unresolved pronoun.

Note that the example rule shown here contains an object for the "Talker" – this is handled as a single talker even though it needs to be matched against a possible group of talkers (e.g. councilmen or demonstrators) because the singular/plural aspect is not relevant for the inference process (either "councilman" or "councilmen" will work). In contrast, the "listeners" are represented as a collection since it is necessary to represent the fact that there is a set of possible listeners; there is logic that determines that that set can include the councilmen, for the cases where the councilmen are not the talker.

```
GenerateAndTest_ProcessOneForwardRule ()    // (partial listing)

// Original NL sentence example:
//   "The city councilmen refused the demonstrators a permit because they advocated violence."
//
//    INPUT: Main forward behavior class: TalkerAdvocatesActionWithListenersWhoAnticipateSomething
//
//     ANTECEDENT: (not shown)
```

```
//          ...
//      CONSEQUENT:
//
//      PopulatedObjectClass "ConsequentActor" ( // Talker
//          <ObjectFrameClass ref = PersonObjectFrameClass />
//          <Attribute ref = CommunicatingState val = "CommunicatingCompleted" />
//      );
//      PopulatedObjectClass "ConsequentActee" ( // Representation-of-Action
//          <ObjectFrameClass ref = CommunicationUnitProposedActionObjectFrameClass /> // e.g. violence
//          <PassiveParticipant val = "true" />
//          <Attribute ref = PassiveIsCommunicatedState val = "Communicated" />
//      );
//      PopulatedObjectClass "ConsequentExtra" (    // Listener(s)
//          <ObjectFrameClass ref = PersonObjectFrameClass />
//          <Multiple val = "true" /> // Collection
//          <ExtraParticipant val = "true" />
//          <Attribute ref = CommunicationReceivedState val = "CommunicationReceived" />
//          <Attribute ref = UniqueIdentityAttributeType var = extra$ />
//      );
//      // reference to a nested rule: this represents that whoever is the listener will fear violence:
//      BehaviorClassReference (
//          <BehaviorClass ref = AnticipateHarmfulEventBehaviorClass />
//          <ParameterActor ref = PersonObjectFrameClass expr = extra$ /> // (reference to the listener(s))
//          <ParameterActee ref = CognitiveRepresentationOfHarmfulEvent />
//      );

//-----------------------------------------------------------------------------------------
// Create a new temporary context along with a structural parent instance ("major"),
//   - sets context fields, and inserts the structural parent instance into the context.
//   - (by default use the first ordinal temporal attribute value of the structural parent class)
//
CreateSandboxContext()

//-----------------------------------------------------------------------------------------
// Create object instances and set values for semantic roles:
//   - create clone of the candidate object instance (pObjInstCandidate) // e.g. "councilmen"
//   - use the pronoun feature set to determine other object instances, e.g. "violence"
//
EstablishObjectInstances()

//-----------------------------------------------------------------------------------------
// Attach all object instances to the structural parent ("major") within the sandbox context:
//
AttachObjectInstancesToStructuralParentMajorAndInstantiate()

//-----------------------------------------------------------------------------------------
// Invoke the Main Inference Routine:
//
PerformForwardDirectedInferenceWithNestedBehavior()

Return from  GenerateAndTest_ProcessOneForwardRule ()
```

## 15.6.2. Main Inference Routine: Application of Two Rules

Now that the sandbox context has been prepared and all temporary object instances exist, the inference process is invoked in order to apply the rules. The inference process involves the application of two rules: the main (forward) rule, and then the nested rule. This routine is called

"PerformForwardDirectedInferenceWithNestedBehavior()". Note that the nested behavior within the main rule's consequent serves the same purpose as an optional causal feature attribute; in this case it represents at the class level what ultimately gets inferred at the instance level.

```
PerformForwardDirectedInferenceWithNestedBehavior()

    //-------------------------------------------------------------------------------------------------
    // Apply the main rule, generating a new structural parent ("minor"):
    //  - also get a pointer to the nested behavior class so it can be subsequently applied
    // Generate new object instances and states from the main ForwardRule:
    // State: CommunicationUnitProposedActionObjectFrameClass:Instance (not used)
    // State: PersonObjectFrameClass (extra$) :Instance: CommunicationRecvState = "CommunicationReceived"
    //
    ApplyBehaviorClass()

    //-------------------------------------------------------------------------------------------------
    // Insert new structural parent minor into the sandbox context:
    //  - also determine the object instance that is to be used when applying the nested behavior
    //
    InsertStructuralParentMinorAndMarkObjectInstanceCorrespondingToUniqueIdentity()

    //-------------------------------------------------------------------------------------------------
    // Apply the nested rule:
    //
    // Generate new object instances and states from the nested rule:
    // State: PersonObjectFrameClass (extra$):Instance: AnticipatingHarmfulEventState = "Anticipating"
    // State: CognitiveRepresentationOfHarmfulEvent:Instance: PassiveIsAnticipatedState = "Anticipated"
    //
    // (this will create another StructuralParentMinor that will be used later during the matching process)
    //
    ApplyBehaviorClass()

Return from PerformForwardDirectedInferenceWithNestedBehavior ()
```

This routine uses the ApplyBehaviorClass() function that is shown below.

### 15.6.3. Lower Inference Routine: ApplyBehaviorClass Routine

The main lower-level worker function that is part of the inference process is called ApplyBehaviorClass(). (This description is limited to the version of this function that processes a single input structural parent instance ("major") in order to generate a single output structural parent instance ("minor"), however the inference process is not limited to two structural parent instances). *(Note: this does not describe the additional functionality required when using the binder mechanism).*

The logic that is implemented here, when the candidate is "demonstrators", is as follows:

**If:**
- the input candidate object instance is "demonstrators" (the actor in the "advocate" clause) AND
- the verb is "advocate" AND
- the current candidate behavior class has been identified as TalkerAdvocatesActionWithListenersWhoAnticipateSomething

**Then:**
- there will exist a collection of listeners (with the "extra" role) that cannot include the actor (demonstrators) (cf. reference for the behavior class, above, that explains this constraint)
- the nested behavior will occur, as a causal (i.e. caused) feature (it must take place given the conjunction of conditions specified in the antecedent of the main rule);  therefore apply the nested rule

The inputs to ApplyBehaviorClass() are:

- A structural parent instance referred to as StructuralParentMajor
- One or more object instances, situated within the StructuralParentMajor. In this particular case, there is at least one "seed" object instance – the candidate object instance. (E.g. "councilmen", e.g. "demonstrators").
- The behavior class, containing:
  - o Bridge object frame class
  - o Specification of which populated object class is the binder source
  - o Antecedent states: a list of rule nodes
  - o Consequent states: a list of rule nodes

Side-effect modifications to input data structures that are performed by this function include:

- Possible attachment of additional newly-created object instances to StructuralParentMajor
- Settings for values of qualitative (state) attributes of object instances that are child objects of StructuralParentMajor. Since this is processing a forward-directed causal rule, only non-optional causal feature state qualitative attributes can be applied to the object instances of the StructuralParentMajor.

The outputs of this function are:

- A structural parent instance referred to as StructuralParentMinor
- A time-related attribute value for StructuralParentMinor
- Attachment of object instances to StructuralParentMinor
- Settings for values of qualitative (state) attributes of object instances that are child objects of StructuralParentMinor

```
ApplyBehaviorClass()

// INPUTS:
//  - Seed object instance
//  - (optional) other object instances
//  - BehaviorClass,  e.g. TalkerAdvocatesActionWithListenersWhoAnticipateSomething

  //---------------------------------------------------------------------------
  // Calculate the timeline time point that will be used for the minor structural parent:
  //
  DetermineNextTimelineTimePoint ()
```

```
//-------------------------------------------------------------------------
// Create a new structural parent (minor):
//
CreateStructuralParentMinor()

//-------------------------------------------------------------------------

Loop:  process Antecedent Elements/Rule Nodes  // (E.g. operate against StructuralParentMajor @T1)

    If Rule node is a populated object class:
    {
        iResult = ApplyPopulatedObjectClassToObjectInstance
                            (// pointer to populated object class,
                             // pointer to list of object instances
                             // flag to indicate this is called from the antecedent
                             // address of pointer for new object instance);

        If a new object was created:
        {
            // Insert it into structural parent major: InsertComponentIntoInstanceStructure ()
        }
    }

    Else If Rule node is a nested behavior class reference:
    {
        // Save the behavior class reference and return it as a parameter
    }

End Loop // Process Antecedent Elements

//-------------------------------------------------------------------------

Loop:  process Consequent Elements/Rule Nodes  // (E.g. operate against StructuralParentMinor @T2)

    If Rule node is a populated object class:
    {
        iResult = ApplyPopulatedObjectClassToObjectInstance
                            (// pointer to populated object class,
                             // pointer to list of object instances
                             // flag to indicate this is called from the antecedent
                             // address of pointer for new object instance);

        If a new object was created:
        {
            // Insert it into structural parent minor: InsertComponentIntoInstanceStructure ()
        }
    }

End Loop // Process Consequent Elements

Return  from  ApplyBehaviorClass ()
```

The function "ApplyPopulatedObjectClassToObjectInstance()" searches the input list of object instances for a match of these criteria: the populated object class's object frame class and the semantic role (e.g. passive role). If an object instance matches it is used, otherwise a new object instance gets created. Subsequently, all qualitative attributes for the object instance are set, based on the qualitative attributes of the populated object class.

### 15.7.    Conclusion

After GenerateAndTest_ProcessOneForwardRule() has completed forward-directed execution for the forward rule and the nested rule, the sandbox context will contain a new structural parent instance (the second "minor" structural parent), which itself will contain object instances (with states). These states are then used by the caller and higher-level callers to support the task of determining if there is a match against another set of states that are either known or inferred (using backward-directed inference) from the main situation context.

# Appendix: Star Classes for the Solution for Winograd Schema #1

## 1. Overview

The Winograd Schema Challenge (Levesque et al: 2012, Davis: 2011) is a set of benchmark tests for assessing whether or not an automated natural language understanding system performs comprehension. This challenge includes a variety of schemas: a schema consists of a pair of descriptive sentences and an associated pair of questions that tests whether or not the system has understood the sentence and its alternate. The NLP task involves anaphora or coreference resolution for an ambiguous, or difficult pronoun that exists in the original sentence. The purpose of the challenge is not to test for simple disambiguation; rather it is to use this task as a test of underlying intelligent capabilities.

The sections below show the classes that were used to create a working system that solves Winograd schema #1: the "councilmen and demonstrators" schema. This schema is as follows:

> "The city councilmen refused the demonstrators a permit because they [feared|advocated] violence."

The system resolves the difficult pronoun for the "feared violence" variant of this schema using a general method that does not use inference against the internal instance mode. However it resolves the pronoun for the "advocated violence" variant of the schema using an embedded commonsense reasoning method; the use of this embedded inference-based method is necessary because the conceptual connection between advocating violence and refusing to grant a permit (on the part of government officials) is not a direct one.

Note that there are a few aspects of the solution that are not shown; they include classes from the "intelligent agent cognition" area (e.g. CognitiveRepresentationOfHarmfulEvent, CommunicationUnitProposedActionObjectFrameClass). *(During the development of the method it was determined that these cognition classes could be used as placeholders as they are not active parts of the inference processes).*

## 2. Supporting Upper Ontology Definitions

### 2.1. Dimension Systems

A file called BasicDefinitions.h contains value set statements, mapping statements, and dimension system statements. The definitions that are used by the object frame classes and behavior classes that follow are included here.

```
//===============================================================
// { Definitions from BasicDefinitions.h }
//
//   Description:
//
```

```
//    General-purpose supporting definitions that are used by other
//    definitions: including value sets, mapping, dimension systems.
//
//================================================================

//--------------------------------------------------------------
// Dimension Systems
//--------------------------------------------------------------
//
DimensionSystem "RelativePosition"   // Dimension system that is used
                                     // only by behavior classes:
(
    LocationAttributeTypes
    (
        SpatialAttributeTypes
        (
            "RelativeLocation"
            (
                <SuperType val = "Locational"/>

                "ValueSet"
                (
                    <SuperTypeUsage val = "LocationalValues" />

                    { "Identical",
                      // (the following enum values allow for relative placement of actor/actee/extra object instances)
                      "Adjacent1",
                      "Adjacent2",
                      "Adjacent3",
                      "NotAdjacent" }
                );
            );
        );
        TemporalAttributeTypes
        (
            "RelativeTime"
            (
                <SuperType val = "Locational"/>

                "ValueSet"
                (
                    <SuperTypeUsage val = "LocationalValues" />

                    { "Identical", // e.g. expr = (t$+0) // (identical)
                      "After",     // e.g. expr = (t$+1) // (after)
                      "Before" }   // e.g. expr = (t$+2) // (before)
                );
            );
        );
    );
); // DimensionSystem "RelativePosition"

Integer lenMaxPhysicalDimension = 1000000000;
// (1 million meters is large enough for the intended uses
//  of this dimension system)

DimensionSystem "PhysicalObjectMillimeterCoordinates"
(
    LocationAttributeTypes
```

```
(
    SpatialAttributeTypes
    (
        "X-Coordinate"
        (
            <SuperType val = "Locational"/>
            "ValueSet"
            (
                <BaseValueSet ref = Millimeter />
                <SuperTypeUsage val = "LocationalValues" />
                { 1, .. lenMaxPhysicalDimension }
            );
        );
        "Y-Coordinate"
        (
            <SuperType val = "Locational"/>
            "ValueSet"
            (
                <BaseValueSet ref = Millimeter />
                <SuperTypeUsage val = "LocationalValues" />
                { 1, .. lenMaxPhysicalDimension }
            );
        );
        "Z-Coordinate"
        (
            <SuperType val = "Locational"/>
            "ValueSet"
            (
                <BaseValueSet ref = Millimeter />
                <SuperTypeUsage val = "LocationalValues" />
                { 1, .. lenMaxPhysicalDimension }
            );
        );
    );
); // DimensionSystem "PhysicalObjectMillimeterCoordinates"

Integer lenMaxTimelineValue = 1000000;
// (1 million seconds is large enough for the intended uses
//  of this dimension system)

DimensionSystem "PhysicalObjectSecondBasedTimeline"
(
    LocationAttributeTypes
    (
        TemporalAttributeTypes
        (
            "AttributeTypeTime"
            (
                <SuperType val = "Locational"/>
                "ValueSet"
                (
                    <BaseValueSet ref = Second />
                    <SuperTypeUsage val = "LocationalValues" />
                    { 1, .. lenMaxTimelineValue }
                );
            );
        );
    );
```

```
);  // DimensionSystem "PhysicalObjectSecondBasedTimeline"
```

## 2.2. EverydayObjectStructuralParent

The high-level object frame class that is used as a structural parent class by many lower ontology object frame classes is called "EverydayObjectStructuralParentClass " and is shown here. It is preceded by some definitions that it needs. (Note that it is usually not directly used by lower ontology classes, but is used indirectly via the inheritance mechanism, as lower ontology classes derive from an intermediary EverydayObjectFrameClass, shown in the next section).

```
//-----------------------------------------------------------------------
//
//  File: EverydayObjectDefinitions.h
//
//  Description:
//
//    Classes and definitions for "everyday objects", defined as
//    objects that are on a scale that allows them to be perceived
//    by and interacted with by human beings or animals.
//
//-----------------------------------------------------------------------

#include <BasicDefinitions.h>

//---------------------------------------------------------------
//
//  EverydayObjectTimelineDimensionSystem
//
//---------------------------------------------------------------
//
DimensionSystem "EverydayObjectTimelineDimensionSystem"
(
   LocationAttributeTypes
   (
     TemporalAttributeTypes
     (
        "Time"  // Values for a timeline that is useful for
                // simple situations involving everyday objects
        (
           <SuperType val = "Locational"/>

           "ValueSet"
           (
              <SuperTypeUsage val = "LocationalValues" />

              // Note: if OrderedCollection is true, the processing system
              //     must implement the equality operators: ==, !=,
              //     and the relational operators: <, >, <=, >=.
              //
              //     The default value is "false" for value sets
              //     that are enumerated lists of values.
              //
              <OrderedCollection val = "true" />

              { "T01", "T02", "T03", "T04", "T05", "T06", "T07", "T08",
```

```
                    "T09", "T10", "T11", "T12", "T13", "T14", "T15", "T16",
                    "T17", "T18", "T19", "T20", "T21", "T22", "T23", "T24",
                    "T25", "T26", "T27", "T28", "T29", "T30" }
              );
           );
        );
     );
  ); // DimensionSystem "EverydayObjectTimelineDimensionSystem"

     //-------------------------------------------------------------
     //
     // EverydayObjectStructuralParentClass
     //
     //-------------------------------------------------------------
     //
     ObjectFrameClass "EverydayObjectStructuralParentClass"
     (
         <StructureTrait val = "Compound"/>
         <StructuralParentClass val = "true"/>

         Dictionary ( English ({ "nil" } ););

         HigherClasses ();         // (not used)
         RelationshipToParent (); // (not used)
         AttributeTypes ();        // (not used)

         DimensionSystems
         (
            DimensionSystem "EverydayObjectDimensionSystem"
            (
               Merge (PhysicalObjectMillimeterCoordinates, EverydayObjectTimelineDimensionSystem);
            );
         );

     ); // EverydayObjectStructuralParentClass
```

## 2.3. Higher-level Class: EverydayObjectFrameClass

The EverydayObjectFrameClass is used extensively by the lower ontology classes. Lower
ontology classes that derive from this class inherit its use of the
EverydayObjectStructuralParentClass as a base structural parent class.

(This class defines a sample MainColor attribute type which may not actually be used by most
lower ontology classes, since they may choose to define their own "color" attribute type (if one is
needed)).

```
     //-------------------------------------------------------------
     //
     // EverydayObjectFrameClass
     //
     //-------------------------------------------------------------
     //
     ObjectFrameClass "EverydayObjectFrameClass"
     (
         <StructureTrait val = "Compound"/>
```

```
// (note: no dictionary entries needed here since this is used
//        solely as an abstract class)
Dictionary ( English ({ "nil" } ););

HigherClasses ();

StructuralParentClassesBase
(
   { "EverydayObjectStructuralParentClass" }
);

RelationshipToParent
(
   AtLocations ();
   OrientationSpecifiers ();
   OuterDimensionSystemExtents ();
);

AttributeTypes
(
   AttributeType "MainColor"
   (
      <SuperType val = "Qualitative"/>

      "Values"
      (
      // (some sample colors)
         { "White" :  Dictionary ( English ( { "white" } ); ); ,
           "Red" :  Dictionary ( English ( { "red" } ); ); ,
           "Yellow" :  Dictionary ( English ( { "yellow" } ); ); ,
           "Blue" :  Dictionary ( English ( { "blue" } ); ); ,
           "Green" :  Dictionary ( English ( { "green" } ); ); ,
           "Orange" :  Dictionary ( English ( { "orange" } ); ); ,
           "Purple" :  Dictionary ( English ( { "purple" } ); ); ,
           "Brown" :  Dictionary ( English ( { "brown" } ); ); ,
           "Black" :  Dictionary ( English ( { "black" } ); );
         }
      );
   );

   Structure ();

); // EverydayObjectFrameClass
```

## 2.4. BehavioralStructuralParent

This is a generic structural parent object frame class that is used by the behavior classes. The important section is DimensionSystems, which contains a declaration of the "RelativePosition" dimension system that was listed earlier.

```
//--------------------------------------------------------------------
//
// "BehavioralStructuralParentClass"
//
//--------------------------------------------------------------------
//
ObjectFrameClass "BehavioralStructuralParentClass"
```

```
(
    <SealedClass val = "true" />
    //
    <StructureTrait val = "Compound"/>
    <StructuralParentClass val = "true"/>

    Dictionary ( English ({ "nil" } ););

    HigherClasses ();        // (not used)
    RelationshipToParent (); // (not used)
    AttributeTypes ();       // (not used)

    DimensionSystems
    (
        DimensionSystem "RelativePosition" (RelativePosition);
    );

    Structure ();

); // "BehavioralStructuralParentClass"
```

## 3.  Transform (Dimension System to Dimension System)

Since the upcoming behavior classes use the BehavioralStructuralParent class, and the object frame classes that are referred to within these behavior classes make routine use of the EverydayObjectStructuralParent class, for some NLU tasks there is a need to convert coordinates from one dimension system to the other. The Transform statement is used for this purposes: a sample is listed here:

```
// Transform for:  (source) BehavioralStructuralParentClass.RelativePosition to
   (dest) EverydayObjectStructuralParentClass.EverydayObjectSpatialCoordinates, which is based upon
   PhysicalObjectMillimeterCoordinates:

        Transform "RelativePositionSpatialToMillimeterBasedCoords-01"
        (
            <Source ref = RelativePosition.SpatialAttributeTypes />
            <Dest ref = PhysicalObjectMillimeterCoordinates.SpatialAttributeTypes />

            bool Routine
            {
                Parameters
                (
                    string Source;   // one of: "IdenticalLocation", "Adjacent", "NotAdjacent"
                    int Dest[3];
                );

                Locals
                (
                    int x = 0;
                    int y = 0;
                    int z = 0;
                );

                if (Source == "IdenticalLocation")
                {
                    Dest [x] = 0;
```

```
            Dest [y] = 0;
            Dest [z] = 0;
        }
        else if (Source == "Adjacent")
        {
            Dest [x] = 2;  // arbitrary distance of 2 millimeters
            Dest [y] = 0;
            Dest [z] = 0;
        }
        else if (Source == "NotAdjacent")
        {
            Dest [x] = 1000;  // arbitrary distance of 1000 millimeters
            Dest [y] = 0;
            Dest [z] = 0;
        }

        Return true;
    }
```

## 4.  Object Frame Classes

### 4.1. IntelligentAgent: Attribute Types

This shows a portion of an IntelligentAgentObjectFrameClass; this is to show the attribute types that are used in the behavior classes that follow.

```
ObjectFrameClass "IntelligentAgentObjectFrameClass"
(
    <StructureTrait val = "Compound"/>
    Dictionary ( English ({ "nil" } ););
    HigherClasses ();

    StructuralParentClassesBase
    (
        {  "EverydayObjectStructuralParentClass" }
    );

    AttributeTypes
    (
        AttributeType "CommunicatingState"
        (
            <SuperType val = "Qualitative"/>
            "Values"
            (
                { "Communicating",
                  "CommunicatingCompleted" }
            );
        );
        AttributeType "AnticipatingHarmfulEventState"
        (
            <SuperType val = "Qualitative"/>
            <StateAttributeType val = "true" />
            <OptionalCausalFeature val = "true" />

            "Values"
            (
                {
```

```
                "NotAnticipating",
                "Anticipating"
            }
        );
    );
    AttributeType "AnticipatingScheduleConflictState"
    (
        <SuperType val = "Qualitative"/>
        <StateAttributeType val = "true" />
        <OptionalCausalFeature val = "true" />

        "Values"
        (
            {
                "NotAnticipating",
                "Anticipating"
            }
        );
    );
); // IntelligentAgentObjectFrameClass
```

## 4.2. Selected Parts of the Person Class

A portion of the Person class is shown here, showing two items that are needed: the reference within the HigherClasses section to the IntelligentAgent class, and several attribute types.

```
ObjectFrameClass "PersonObjectFrameClass"
(
    <StructureTrait val = "Compound"/>

    Dictionary
    (
        English
        (
            {
                "person",
                "persons",
                "human",
                "humans"
            }
        );
    );

    HigherClasses
    (
        { "EverydayObjectFrameClass",
          "IntelligentAgentObjectFrameClass",
          "EarthBoundObjectFrameClass" } // provides orientation specifiers, e.g. "above", "below"
    );

    AttributeTypes
    (

        AttributeType "RequestingState"
        (
            <SuperType val = "Qualitative"/>
```

```
                <StateAttributeType val = "true"/>

                "Values"
                (
                  { "NotRequesting",
                    "Requesting" }
                );
            );

            AttributeType "PassiveIsRequestedFromState"
            (
                <SuperType val = "Qualitative"/>
                <StateAttributeType val = "true"/>

                "Values"
                (
                  { "NotRequestedFrom",
                    "RequestedFrom" }
                );
            );

            AttributeType "RefusingState"
            (
                <SuperType val = "Qualitative"/>
                <StateAttributeType val = "true"/>

                "Values"
                (
                  { "NotRefusing",
                    "Refusing" }
                );
            );

            AttributeType "PassiveIsRefusedState"
            (
                <SuperType val = "Qualitative"/>
                <StateAttributeType val = "true"/>

                "Values"
                (
                  { "NotRefused",
                    "Refused" }
                );
            );

        ); // AttributeTypes

    ); // ObjectFrameClass "PersonObjectFrameClass"
```

## 4.3. Other Lower-ontology Classes

The following are several bottom-level classes that are used by the behavior classes that follow.

```
        ObjectFrameClass "GovernmentSubjectObjectFrameClass"
        (
            // (generated from) "A government subject is a person."
```

```
<StructureTrait val = "Compound"/>

DictionaryPriorWord
(
    <DictionaryWordIsNoun val = "true" />

    English
    (
        { "government",
          "government" }
    );
);

Dictionary ( English
(
    { "subject",
      "subjects" }
););

HigherClasses ( { "PersonObjectFrameClass" } );

);

ObjectFrameClass "GovernmentOfficialObjectFrameClass"
(
    // (generated from) "A government official is a person."
    // (generated from) "A government official can grant a request."

    <StructureTrait val = "Compound"/>

    DictionaryPriorWord
    (
        <DictionaryWordIsNoun val = "true" />

        English
        (
            { "government",
              "government" }
        );
    );

    Dictionary ( English
    (
        { "official",
          "officials" }
    ););

    HigherClasses ( { "PersonObjectFrameClass" } );

    AttributeTypes
    (
        AttributeType "GrantingState"
        (
            <SuperType val = "Qualitative"/>

            <StateAttributeType val = "true" />

            "Values"
```

```
            (
                {
                  "NotGranting",
                  "Granting"
                  }
                );
            );
        );

ObjectFrameClass "CityCouncilmanObjectFrameClass"
(
    // (generated from) "A city councilman is a government official."

    <StructureTrait val = "Compound"/>

    DictionaryPriorWord
    (
        <DictionaryWordIsNoun val = "true" />

        English
        (
            { "city",
              "city" }
        );
    );

    Dictionary ( English
    (
        { "councilman",
          "councilmen" }
    ););

    HigherClasses ( { "GovernmentOfficialObjectFrameClass" } );
);

ObjectFrameClass "DemonstratorObjectFrameClass"
(
    // (generated from) "A demonstrator is a person."

    <StructureTrait val = "Compound"/>

    Dictionary ( English
    (
        { "demonstrator",
          "demonstrators" }
    ););

    HigherClasses ( { "GovernmentSubjectObjectFrameClass" } );
);
```

## 5.  Behavior Classes

### 5.1. Classes that are Referenced as Nested Classes

The following two classes involve anticipation of a harmful event on the part of a person: they are referred to as nested behavior classes by the behavior classes shown in subsequent sections.

```
//---------------------------------------------------------------------
//
//  BehaviorClass: AnticipateHarmfulEventBehaviorClass
//
//     "A person can anticipate a harmful event."
//
//---------------------------------------------------------------------
//
BehaviorClass "AnticipateHarmfulEventBehaviorClass"
(
    <BridgeObjectFrameClass ref = BehavioralStructuralParentClass />

    Dictionary ( English
    (
        {
          "anticipate",
          "anticipated",
          "anticipated",
          "anticipates",
          "anticipating",
          "fear",
          "feared",
          "feared",
          "fears",
          "fearing"
        }
    ));

    PriorStates
    (
        PopulatedObjectClass "AntecedentActor"
        (
            <ObjectFrameClass ref = PersonObjectFrameClass />
            <BinderSourceFlag val = "true" />
            <DimensionSystem ref = RelativePosition />
            <Attribute ref = RelativeLocation var = a$ />
            <Attribute ref = RelativeTime var = t1$ />
            <Attribute ref = AnticipatingHarmfulEventState val = "NotAnticipating" />
        );
        PopulatedObjectClass "AntecedentActee"
        (
            <ObjectFrameClass ref = CognitiveRepresentationOfHarmfulEvent />
            <PassiveParticipant val = "true" />
            <DimensionSystem ref = RelativePosition />
            <Attribute ref = RelativeLocation expr = (a$+1) />
            <Attribute ref = RelativeTime expr = t1$ />
            <Attribute ref = PassiveIsAnticipatedState val = "NotAnticipated" />
        );
    );

    PostStates
    (
        PopulatedObjectClass "ConsequentActor"
        (
            <ObjectFrameClass ref = PersonObjectFrameClass />
            <DimensionSystem ref = RelativePosition />
            <Attribute ref = RelativeLocation expr = (a$+1) />
            <Attribute ref = RelativeTime expr = (t1$+1) />
```

```
                    <Attribute ref = AnticipatingHarmfulEventState val = "Anticipating" />
                );
            PopulatedObjectClass "ConsequentActee"
            (
                    <ObjectFrameClass ref = CognitiveRepresentationOfHarmfulEvent />
                    <PassiveParticipant val = "true" />
                    <DimensionSystem ref = RelativePosition />
                    <Attribute ref = RelativeLocation expr = (a$+1) />
                    <Attribute ref = RelativeTime expr = (t1$+1) />
                    <Attribute ref = PassiveIsAnticipatedState val = "Anticipated" />
                );
        );
); // BehaviorClass "AnticipateHarmfulEventBehaviorClass"

//-------------------------------------------------------------------------
//
//  BehaviorClass: AnticipateScheduleConflictBehaviorClass
//
//    "A person can anticipate a schedule conflict."
//
//-------------------------------------------------------------------------
//
BehaviorClass "AnticipateScheduleConflictBehaviorClass"
(
        <BridgeObjectFrameClass ref = BehavioralStructuralParentClass />

    Dictionary ( English
        (
            {
              "anticipate",
              "anticipated",
              "anticipated",
              "anticipates",
              "anticipating"
            }
        ););

    PriorStates
        (
        PopulatedObjectClass "AntecedentActor"
            (
                    <ObjectFrameClass ref = PersonObjectFrameClass />
                    <BinderSourceFlag val = "true" />
                    <DimensionSystem ref = RelativePosition />
                    <Attribute ref = RelativeLocation var = a$ />
                    <Attribute ref = RelativeTime var = t1$ />
                    <Attribute ref = AnticipatingScheduleConflictState val = "NotAnticipating" />
                );
        PopulatedObjectClass "AntecedentActee"
            (
                    <ObjectFrameClass ref = CognitiveRepresentationOfScheduleConflict />
                    <PassiveParticipant val = "true" />
                    <DimensionSystem ref = RelativePosition />
                    <Attribute ref = RelativeLocation expr = (a$+1) />
                    <Attribute ref = RelativeTime expr = t1$ />
                    <Attribute ref = PassiveIsAnticipatedState val = "NotAnticipated" />
                );
        );
```

```
    PostStates
    (
        PopulatedObjectClass "ConsequentActor"
        (
            <ObjectFrameClass ref = PersonObjectFrameClass />
            <DimensionSystem ref = RelativePosition />
            <Attribute ref = RelativeLocation expr = (a$+1) />
            <Attribute ref = RelativeTime expr = (t1$+1) />
            <Attribute ref = AnticipatingScheduleConflictState val = "Anticipating" />
        );
        PopulatedObjectClass "ConsequentActee"
        (
            <ObjectFrameClass ref = CognitiveRepresentationOfScheduleConflict />
            <PassiveParticipant val = "true" />
            <DimensionSystem ref = RelativePosition />
            <Attribute ref = RelativeLocation expr = (a$+1) />
            <Attribute ref = RelativeTime expr = (t1$+1) />
            <Attribute ref = PassiveIsAnticipatedState val = "Anticipated" />
        );
    );
); // BehaviorClass "AnticipateScheduleConflictBehaviorClass"
```

## 5.2. Behavior Class Used by Both General and Embedded Inference Routines

This behavior class is actually used by both forms of the schema (the "fear violence" variant and the "advocate violence" variant). (Note that the probability field within the behavior class reference is optional: see the following section for explanation of how this field is used). The nested behavior (in the BehaviorClassReference element), as it exists within the antecedent of the rule, has a causal relationship with the elements of the consequent part of the rule.

```
//-----------------------------------------------------------------------
//
// BehaviorClass: "RefusingSomethingDueToFearBehaviorClass"
//
//    "If a person(s) anticipates a harmful event
//      then he/she/they will not grant a thing that was requested (e.g. a permit request)."
//
//-----------------------------------------------------------------------
//
BehaviorClass "RefusingSomethingDueToFearBehaviorClass"
(
    <CausalRule val = "true" />

    <BridgeObjectFrameClass ref = BehavioralStructuralParentClass />

    DictionaryPriorWord ( English
    (
        { "", "", "", "", "",
          "not", "not", "not", "not", "not" }
    ););

    Dictionary ( English
    (
        { "refuse", "refused", "refused", "refuses", "refusing",
          "grant", "granted", "granted", "grants", "granting" }
    ););
```

```
PriorStates  // (antecedent)
(
    PopulatedObjectClass "AntecedentActor"
    (
        <ObjectFrameClass ref = PersonObjectFrameClass />  // e.g. government official(s)
        <BinderSourceFlag val = "true" />
        <DimensionSystem ref = RelativePosition />
        <Attribute ref = RelativeLocation var = a$ />
        <Attribute ref = RelativeTime var = t1$ />
        <Attribute ref = RefusingState val = "NotRefusing" />
        <Attribute ref = UniqueIdentityAttributeType var = q$ />
    );
    BehaviorClassReference  // ("if a person anticipates a harmful event")
    (
        <Probability expr = 0.9 />
        <BehaviorClass ref = AnticipateHarmfulEventBehaviorClass />
        <ParameterActor ref = PersonObjectFrameClass expr = q$ /> // (refers to the actor)
        <ParameterActee ref = CognitiveRepresentationOfHarmfulEvent />
    );
    PopulatedObjectClass "AntecedentActee" // e.g. demonstrators
    (
        <ObjectFrameClass ref = PersonObjectFrameClass />
        <PassiveParticipant val = "true" />
        <DimensionSystem ref = RelativePosition />
        <Attribute ref = RelativeLocation expr = (a$+1) />
        <Attribute ref = RelativeTime expr = t1$ />
        <Attribute ref = PassiveIsRefusedState val = "NotRefused" />
    );
    PopulatedObjectClass "AntecedentExtra"
    (
        <ObjectFrameClass ref = CommunicationUnitRequestObjectFrameClass />  // e.g. the permit
        <ExtraParticipant val = "true" />
        <DimensionSystem ref = RelativePosition />
        <Attribute ref = RelativeLocation expr = (a$+2) />
        <Attribute ref = RelativeTime expr = t1$ />
        <Attribute ref = PassiveRepresentedItemIsRefusedState val = "NotRefused" />
    );
);
PostStates  // (consequent)
(
    PopulatedObjectClass "ConsequentActor"
    (
        <ObjectFrameClass ref = PersonObjectFrameClass />  // e.g. government official(s)
        <DimensionSystem ref = RelativePosition />
        <Attribute ref = RelativeLocation expr = a$ />
        <Attribute ref = RelativeTime expr = (t1$+1) />
        <Attribute ref = RefusingState val = "Refusing" />
    );
    PopulatedObjectClass "ConsequentActee" // e.g. demonstrators
    (
        <ObjectFrameClass ref = PersonObjectFrameClass />
        <PassiveParticipant val = "true" />
        <DimensionSystem ref = RelativePosition />
        <Attribute ref = RelativeLocation expr = (a$+1) />
        <Attribute ref = RelativeTime expr = (t1$+1) />
        <Attribute ref = PassiveIsRefusedState val = "Refused" />
    );
    PopulatedObjectClass "ConsequentExtra"
```

```
    (
        <ObjectFrameClass ref = CommunicationUnitRequestObjectFrameClass /> // e.g. the permit
        <ExtraParticipant val = "true" />
        <DimensionSystem ref = RelativePosition />
        <Attribute ref = RelativeLocation expr = (a$+2) />
        <Attribute ref = RelativeTime expr = (t1$+1) />
        <Attribute ref = PassiveRepresentedItemIsRefusedState val = "Refused" />
    );
  );
); // BehaviorClass "RefusingSomethingDueToFearBehaviorClass"
```

### 5.3. Additional Behavior Classes Included for Testing Purposes

The following behavior classes are included in order that the ontology/knowledge base might more accurately model a real-world ontology and knowledge base.

### 5.3.1. RefusingSomethingDueToFearOnPartOfRequestorBehaviorClass

This class is similar to the behavior class above and is included for purposes of testing the probability field within the nested behavior. The anaphora reference system finds both behavior classes and determines that each is a match (RefusingSomethingDueToFearBehaviorClass matches the actor pronoun candidate object instance, and the behavior class here matches the actee candidate). The resolution method then compares probability values and chooses the candidate for which there is a greater probability (e.g., it is more probable that the person refusing the request (the actor, e.g. the councilmen) anticipated a harmful event than that this was the requesting person(s) (e.g. the demonstrators). Note that there is a very low probability of this behavior occurring (designated here at 2%), but it is possible and could be a reasonable explanation for the refusal of a request for something. The probability values here and above have been chosen somewhat arbitrarily; a system that performs ontology derivation from data would determine the actual probability values based on the data.

```
    //---------------------------------------------------------------------
    //
    // BehaviorClass: "RefusingSomethingDueToFearOnPartOfRequestorBehaviorClass"
    //
    //   "If a first person is requested something from
    //    someone who anticipates a harmful event,
    //     then the first person does not grant the thing that was requested."
    //
    // (For testing: included to test probability)
    //
    //---------------------------------------------------------------------
    //
    BehaviorClass "RefusingSomethingDueToFearOnPartOfRequestorBehaviorClass"
    (
        <CausalRule val = "true" />

        <BridgeObjectFrameClass ref = BehavioralStructuralParentClass />

        //** <Negation val = "true" />

        DictionaryPriorWord ( English
```

```
(
    { "", "", "", "", "",
      "not", "not", "not", "not", "not" }
););

Dictionary ( English
(
    { "refuse", "refused", "refused", "refuses", "refusing",
      "grant", "granted", "granted", "grants", "granting" }
););

PriorStates
(
    PopulatedObjectClass "AntecedentActor"
    (
        <ObjectFrameClass ref = PersonObjectFrameClass />  // e.g. government official(s)
        <BinderSourceFlag val = "true" />
        <DimensionSystem ref = RelativePosition />
        <Attribute ref = RelativeLocation var = a$ />
        <Attribute ref = RelativeTime var = t1$ />
        <Attribute ref = RefusingState val = "NotRefusing" />
    );
    BehaviorClassReference
    (
        <BehaviorClass ref = AnticipateHarmfulEventBehaviorClass />
        <Probability expr = 0.02 />
        <ParameterActor ref = PersonObjectFrameClass expr = q$ />  // (refers to the actee)
        <ParameterActee ref = CognitiveRepresentationOfHarmfulEvent />
    );
    PopulatedObjectClass "AntecedentActee"  // e.g. demonstrators
    (
        <ObjectFrameClass ref = PersonObjectFrameClass />
        <PassiveParticipant val = "true" />
        <DimensionSystem ref = RelativePosition />
        <Attribute ref = RelativeLocation expr = (a$+1) />
        <Attribute ref = RelativeTime expr = t1$ />
        <Attribute ref = PassiveIsRefusedState val = "NotRefused" />
        <Attribute ref = UniqueIdentityAttributeType var = q$ />
    );
    PopulatedObjectClass "AntecedentExtra"
    (
        <ObjectFrameClass ref = CommunicationUnitRequestObjectFrameClass />  // e.g. the permit
        <ExtraParticipant val = "true" />
        <DimensionSystem ref = RelativePosition />
        <Attribute ref = RelativeLocation expr = (a$+2) />
        <Attribute ref = RelativeTime expr = t1$ />
        <Attribute ref = PassiveRepresentedItemIsRefusedState val = "NotRefused" />
    );
);
PostStates
(
    PopulatedObjectClass "ConsequentActor"
    (
        <ObjectFrameClass ref = PersonObjectFrameClass />  // e.g. government official(s)
        <DimensionSystem ref = RelativePosition />
        <Attribute ref = RelativeLocation expr = a$ />
        <Attribute ref = RelativeTime expr = (t1$+1) />
        <Attribute ref = RefusingState val = "Refusing" />
    );
```

```
PopulatedObjectClass "ConsequentActee" // e.g. demonstrators
(
    <ObjectFrameClass ref = PersonObjectFrameClass />
    <PassiveParticipant val = "true" />
    <DimensionSystem ref = RelativePosition />
    <Attribute ref = RelativeLocation expr = (a$+1) />
    <Attribute ref = RelativeTime expr = (t1$+1) />
    <Attribute ref = PassiveIsRefusedState val = "Refused" />
);
PopulatedObjectClass "ConsequentExtra"
(
    <ObjectFrameClass ref = CommunicationUnitRequestObjectFrameClass />  // e.g. the permit
    <ExtraParticipant val = "true" />
    <DimensionSystem ref = RelativePosition />
    <Attribute ref = RelativeLocation expr = (a$+2) />
    <Attribute ref = RelativeTime expr = (t1$+1) />
    <Attribute ref = PassiveRepresentedItemIsRefusedState val = "Refused" />
);
);
); // BehaviorClass "RefusingSomethingDueToFearOnPartOfRequestorBehaviorClass"
```

### 5.3.2. RefusingSomethingDueToScheduleConflictBehaviorClass

This class is similar to the behavior classes above; it is part of the Infopedia for purposes of functionality and scalability testing of the anaphora resolution and inference methods. This class and the other "refusing" behavior classes above are each retrieved by the behavior class query routine based on the query criteria (involving object frame class types and the "refused" verb); this class is used by the inference routine but rejected when the matching process fails to match it against an instance of the nested AnticipateScheduleConflict behavior class (since the councilmen did not anticipate a schedule conflict).

```
//-----------------------------------------------------------------------
//
// BehaviorClass: "RefusingSomethingDueToScheduleConflictBehaviorClass"
//
//   This behavior corresponds to:
//
//   "If a person(s) anticipates a schedule conflict
//      then he/she/they will not grant a thing that was requested (e.g. a permit request)."
//
//-----------------------------------------------------------------------
//
BehaviorClass "RefusingSomethingDueToScheduleConflictBehaviorClass"
(
    <CausalRule val = "true" />

    <BridgeObjectFrameClass ref = BehavioralStructuralParentClass />

    DictionaryPriorWord ( English
    (
      { "", "", "", "", "",
        "not", "not", "not", "not", "not" }
    ););

    Dictionary ( English
    (
```

```
    { "refuse", "refused", "refused", "refuses", "refusing",
      "grant", "granted", "granted", "grants", "granting" }
););

PriorStates
(
    PopulatedObjectClass "AntecedentActor"
    (
        <ObjectFrameClass ref = PersonObjectFrameClass />  // e.g. government official(s)
        <BinderSourceFlag val = "true" />
        <DimensionSystem ref = RelativePosition />
        <Attribute ref = RelativeLocation var = a$ />
        <Attribute ref = RelativeTime var = t1$ />
        <Attribute ref = RefusingState val = "NotRefusing" />
        <Attribute ref = UniqueIdentityAttributeType var = q$ />
    );
    BehaviorClassReference
    (
        <BehaviorClass ref = AnticipateScheduleConflictBehaviorClass />  // NESTED-BEHAVIOR-->>
        <ParameterActor ref = PersonObjectFrameClass expr = q$ />
        <ParameterActee ref = CognitiveRepresentationOfScheduleConflict />
    );
    PopulatedObjectClass "AntecedentActee"  // e.g. demonstrators
    (
        <ObjectFrameClass ref = PersonObjectFrameClass />
        <PassiveParticipant val = "true" />
        <DimensionSystem ref = RelativePosition />
        <Attribute ref = RelativeLocation expr = (a$+1) />
        <Attribute ref = RelativeTime expr = t1$ />
        <Attribute ref = PassiveIsRefusedState val = "NotRefused" />
    );
    PopulatedObjectClass "AntecedentExtra"
    (
        <ObjectFrameClass ref = CommunicationUnitRequestObjectFrameClass />  // e.g. a permit
        <ExtraParticipant val = "true" />
        <DimensionSystem ref = RelativePosition />
        <Attribute ref = RelativeLocation expr = (a$+2) />
        <Attribute ref = RelativeTime expr = t1$ />
        <Attribute ref = PassiveRepresentedItemIsRefusedState val = "NotRefused" />
    );
);
PostStates
(
    PopulatedObjectClass "ConsequentActor"
    (
        <ObjectFrameClass ref = PersonObjectFrameClass />  // e.g. government official(s)
        <DimensionSystem ref = RelativePosition />
        <Attribute ref = RelativeLocation expr = a$ />
        <Attribute ref = RelativeTime expr = (t1$+1) />
        <Attribute ref = RefusingState val = "Refusing" />
    );
    PopulatedObjectClass "ConsequentActee"  // e.g. demonstrators
    (
        <ObjectFrameClass ref = PersonObjectFrameClass />
        <PassiveParticipant val = "true" />
        <DimensionSystem ref = RelativePosition />
        <Attribute ref = RelativeLocation expr = (a$+1) />
        <Attribute ref = RelativeTime expr = (t1$+1) />
        <Attribute ref = PassiveIsRefusedState val = "Refused" />
```

```
        );
        PopulatedObjectClass "ConsequentExtra"
        (
            <ObjectFrameClass ref = CommunicationUnitRequestObjectFrameClass /> // e.g. a permit
            <ExtraParticipant val = "true" />
            <DimensionSystem ref = RelativePosition />
            <Attribute ref = RelativeLocation expr = (a$+2) />
            <Attribute ref = RelativeTime expr = (t1$+1) />
            <Attribute ref = PassiveRepresentedItemIsRefusedState val = "Refused" />
        );
    );
); // BehaviorClass "RefusingSomethingDueToScheduleConflictBehaviorClass"
```

## 5.4. Behavior Class that is Used by the Embedded Inference Routines

The following class describes the behavior of a "talker" – a person or persons who communicates "something" – this behavior involves advocating a proposed action or set of actions (e.g. "violence"). The behavior class involves three main roles: an actor, an actee (the passive role), and an "extra" role. The actor role is filled by the talker (e.g. "demonstrators"), and the extra role is filled by a collection, or set of listeners (e.g. "councilmen"). The populated object classes for the extra role have a field shown as "<Multiple val = "true" />. This field indicates that there is a collection of the object (person) referred to.

The nested behavior reference in the consequent part refers to the AnticipateHarmfulEventClass that was previously described. The inference process that uses this rule must not only process this rule against a temporary ("sandbox") internal instance model, but it must also extract and then apply, or fire the nested behavior rule.

```
//------------------------------------------------------------------------
//
// BehaviorClass: TalkerAdvocatesActionWithListenersWhoAnticipateSomething
//
//     Actor: Talker
//     Actee: Repr-Action
//     Extra: Listener(s)
//
//------------------------------------------------------------------------
//
BehaviorClass "TalkerAdvocatesActionWithListenersWhoAnticipateSomething"
(
    <CausalRule val = "true" />
    <RuleDirection type = "Forward" />
    <BridgeObjectFrameClass ref = BehavioralStructuralParentClass />

    Dictionary ( English
    (
        {
          "advocate",
          "advocated",
          "advocated",
          "advocates",
          "advocating"
        }
    ););
```

```
PriorStates  // (Antecedent)
(
    PopulatedObjectClass "AntecedentActor"  // Talker
    (
        <ObjectFrameClass ref = PersonObjectFrameClass />
        <BinderSourceFlag val = "true" />
        <DimensionSystem ref = RelativePosition />
        <Attribute ref = RelativeLocation var = a$ />
        <Attribute ref = RelativeTime var = t1$ />
        <Attribute ref = CommunicatingState val = "Communicating" />
    );
    PopulatedObjectClass "AntecedentActee"  // Repr-Action
    (
        <ObjectFrameClass ref = CommunicationUnitProposedActionObjectFrameClass />  // e.g. to do violence
        <PassiveParticipant val = "true" />
        <DimensionSystem ref = RelativePosition />
        <Attribute ref = RelativeLocation expr = (a$+2) />
        <Attribute ref = RelativeTime expr = t1$ />
        <Attribute ref = PassiveIsCommunicatedState val = "NotCommunicated" />
    );
    PopulatedObjectClass "AntecedentExtra"  // Listener(s)
    (
        <ObjectFrameClass ref = PersonObjectFrameClass />
        <Multiple val = "true" />  // Collection
        <ExtraParticipant val = "true" />
        <DimensionSystem ref = RelativePosition />
        <Attribute ref = RelativeLocation expr = (a$+1) />
        <Attribute ref = RelativeTime expr = t1$ />
        <Attribute ref = CommunicationReceivedState val = "NotCommunicationReceived" />
    );
);
PostStates  // (Consequent)
(
    PopulatedObjectClass "ConsequentActor"  // Talker
    (
        <ObjectFrameClass ref = PersonObjectFrameClass />
        <DimensionSystem ref = RelativePosition />
        <Attribute ref = RelativeLocation expr = a$ />
        <Attribute ref = RelativeTime expr = (t1$+1) />
        <Attribute ref = CommunicatingState val = "CommunicatingCompleted" />
    );
    PopulatedObjectClass "ConsequentActee"  // Repr-Action
    (
        <ObjectFrameClass ref = CommunicationUnitProposedActionObjectFrameClass />  // e.g. to do violence
        <PassiveParticipant val = "true" />
        <DimensionSystem ref = RelativePosition />
        <Attribute ref = RelativeLocation expr = (a$+2) />
        <Attribute ref = RelativeTime expr = (t1$+1) />
        <Attribute ref = PassiveIsCommunicatedState val = "Communicated" />
    );
    PopulatedObjectClass "ConsequentExtra"  // Listener(s)
    (
        <ObjectFrameClass ref = PersonObjectFrameClass />
        <Multiple val = "true" />  // Collection
        <ExtraParticipant val = "true" />
        <DimensionSystem ref = RelativePosition />
        <Attribute ref = RelativeLocation expr = (a$+1) />
        <Attribute ref = RelativeTime expr = (t1$+1) />
        <Attribute ref = CommunicationReceivedState val = "CommunicationReceived" />
```

```
                <Attribute ref = UniqueIdentityAttributeType var = extra$ />
        );
        BehaviorClassReference
        (
            <BehaviorClass ref = AnticipateHarmfulEventBehaviorClass />
            <ParameterActor ref = PersonObjectFrameClass expr = extra$ />
            <ParameterActee ref = CognitiveRepresentationOfHarmfulEvent />
        );
    );
); // BehaviorClass "TalkerAdvocatesActionWithListenersWhoAnticipateSomething"
```

## 6.  Summary Conclusion

The anaphora resolution method that uses the above classes handles the "fear violence" variant of the schema differently from the "advocate violence" variant. (Details of this method are beyond the scope of this document – please refer to section 16 "Introduction to Inference Using ROSS" for an overview of the embedded inference process).